\documentclass[conference,compsoc]{IEEEtran} %

\usepackage{booktabs}
\usepackage[T1]{fontenc} 
\usepackage{amsmath}
\usepackage{rotating}
\usepackage{epsfig,endnotes}
\usepackage[colorinlistoftodos,prependcaption,textsize=tiny]{todonotes}
\usepackage{url}

\usepackage{mathtools}
\usepackage{algorithm}
\usepackage[noend]{algpseudocode}
\usepackage{tabularx,colortbl}
\usepackage{multirow}
\usepackage[Symbolsmallscale]{upgreek}
\usepackage{mathtools}
\usepackage{tablefootnote}
\usepackage{siunitx}
\usepackage{listings}
\usepackage{enumitem}
\usepackage{color} 
\usepackage{footmisc}
\usepackage{amsmath}
\usepackage{hyperref}
\usepackage{booktabs}
\usepackage{subcaption}
\usepackage[numbers,sort&compress]{natbib}

\usepackage{makecell}
\usepackage{gensymb}
\usepackage{listings}
\usepackage{multirow}
\usepackage{caption}

\usepackage{mathtools, nccmath}
\usepackage{titlesec}

\usepackage{times}
\usepackage{epsfig}
\usepackage{graphicx}
\usepackage{amsmath}
\usepackage{amssymb}
\usepackage{subcaption}
\usepackage{listings}
\usepackage{multirow}
\usepackage{enumitem}
\usepackage{xcolor}

\definecolor{codegreen}{rgb}{0,0.6,0}
\definecolor{codegray}{rgb}{0.5,0.5,0.5}
\definecolor{codepurple}{rgb}{0.58,0,0.82}
\definecolor{backcolour}{rgb}{0.95,0.95,0.92}

\lstdefinestyle{mystyle}{
    backgroundcolor=\color{backcolour},   
    commentstyle=\color{codegreen},
    keywordstyle=\color{magenta},
    numberstyle=\tiny\color{codegray},
    stringstyle=\color{codepurple},
    basicstyle=\ttfamily\footnotesize,
    breakatwhitespace=false,         
    breaklines=true,                 
    captionpos=b,                    
    keepspaces=true,                 
    numbers=left,                    
    numbersep=5pt,                  
    showspaces=false,                
    showstringspaces=false,
    showtabs=false,                  
    tabsize=2
}

\usepackage[T1]{fontenc}
\usepackage[utf8]{inputenc}
\usepackage{babel}
\usepackage[font=small,labelfont=bf]{caption}

\usepackage[framemethod=tikz]{mdframed}
\mdfdefinestyle{remarkstyle}{
backgroundcolor=lightgray,
roundcorner=3pt,
innerleftmargin=3pt,
innerrightmargin=3pt,
innertopmargin=3pt,
innerbottommargin=3pt,
linewidth=0.5pt,
}
\newcounter{insight}[section]
\newenvironment{insight}
    {\refstepcounter{insight}
    \begin{mdframed}[style=remarkstyle]
    \noindent
    \textbf{Insight~\theinsight}: \em
    }
    {
    \end{mdframed}
    \vspace{-1mm}
    }

\usepackage{tikz}
\usetikzlibrary{tikzmark}

\title{A Privacy Enhancing Technique to Evade Detection by Street Video Cameras Without Using Adversarial Accessories}

\newcommand\Mark[1]{\textsuperscript#1}

\begingroup
\centering
\author{
Jacob Shams\Mark{1}, Ben Nassi\Mark{1}, Satoru Koda\Mark{2}, Asaf Shabtai\Mark{1}, Yuval Elovici\Mark{1}\\
\Mark{1}Ben-Gurion University of the Negev,
\Mark{2}Fujitsu Limited\\
\{jacobsh, nassib\}@post.bgu.ac.il\\
koda.satoru@fujitsu.com\\
\{shabtaia, elovici\}@bgu.ac.il
}

\endgroup

\makeatletter
\newcommand{\linebreakand}{%
  \end{@IEEEauthorhalign}
  \hfill\mbox{}\par
  \mbox{}\hfill\begin{@IEEEauthorhalign}
}
\makeatother

\begin{document}
\maketitle

\begin{abstract} 
In this paper, we propose a privacy-enhancing technique leveraging an inherent property of automatic pedestrian detection algorithms, namely, that the training of deep neural network (DNN) based methods is generally performed using curated datasets and laboratory settings, while the operational areas of these methods are dynamic real-world environments. In particular, we leverage a novel side effect of this gap between the laboratory and the real world: location-based weakness in pedestrian detection. We demonstrate that the position (distance, angle, height) of a person, and ambient light level, directly impact the confidence of a pedestrian detector when detecting the person. We then demonstrate that this phenomenon is present in pedestrian detectors observing a stationary scene of pedestrian traffic, with blind spot areas of weak detection of pedestrians with low confidence. We show how privacy-concerned pedestrians can leverage these blind spots to evade detection by constructing a minimum confidence path between two points in a scene, reducing the maximum confidence and average confidence of the path by up to 0.09 and 0.13, respectively, over direct and random paths through the scene. To counter this phenomenon, and force the use of more costly and sophisticated methods to leverage this vulnerability, we propose a novel countermeasure to improve the confidence of pedestrian detectors in blind spots, raising the max/average confidence of paths generated by our technique by 0.09 and 0.05, respectively. In addition, we demonstrate that our countermeasure improves a Faster R-CNN-based pedestrian detector’s TPR and average true positive confidence by 0.03 and 0.15, respectively.

\end{abstract}
\section {Introduction}

In recent years, street video cameras have been increasingly deployed in public spaces, for purposes including pedestrian traffic observation \cite{zhou2022prediction} and automatic identification of criminals/terrorists \cite{ahmed2021hawkeye}. This deployment is widespread throughout much of the world. For example, there are nearly 22,000 street video cameras in Tokyo, Japan, an average of 1.67 cameras per 1000 residents \cite{sheng2021surveilling}.

This widespread deployment of street video-based automatic pedestrian detection systems raises potential privacy concerns.

One concern is that some street video cameras stream footage of pedestrians over the Internet in real-time. This is a privacy concern for pedestrians since this footage is often uploaded online without the pedestrians' consent.
An additional concern is that pedestrian footage can be stored on unsecured servers. This exposes the footage to potential data breaches and jeopardizes pedestrian privacy.

In response to this privacy risk, various privacy-enhancing technologies/methods have been introduced to help pedestrians evade automatic detection by algorithms applied using video footage captured by street video cameras. Such methods showed how the use of adversarial accessories, including masks \cite{zolfi2022adversarial}, glasses \cite{sharif2016accessorize}, hats \cite{Komkov_2021}, makeup \cite{guetta2021dodgingattackusingcarefully}, and shirts \cite{xu2020adversarial}, degrades the performance of automatic pedestrian detection algorithms in the physical domain, allowing a person to walk undetected in the presence of such algorithms. However, such methods require expertise in adversarial machine learning, white-box access to the target model, and the necessity to create and carry adversarial accessories (e.g., the ability to print the adversarial perturbation on a sticker, hat, or shirt). Therefore, we consider the following question: Can a pedestrian evade automatic detection without the need to utilize adversarial accessories?

In this paper, we demonstrate that the performance of automatic pedestrian detection algorithms in the physical domain is affected by the height, angle, and distance of the object detector's view, as well as the ambient light of the scene. 
This yields different detection confidences of objects in the same image, and in some cases can lead to "blind spots" of areas that are captured in the image but recognized with lower confidence scores. 
This has significant implications for applications such as pedestrian detection in street video footage or surveillance cameras, where the angle, distance, and height a person stands, relative to an AI-based pedestrian detection system, as well as the ambient light, results in varying average confidence of the AI system's pedestrian detections.
This demonstrates that a person's position, as well as ambient light, can have a direct impact on the confidence of a pedestrian detection system. 
This can be leveraged by a pedestrian to protect their privacy by either avoiding detection or be detected with reduced confidence, without requiring any adversarial accessories \cite{Komkov_2021, zolfi2022adversarial, thys2019fooling, wu2020making} or prior setup in the scene \cite{274691}.

\begin{figure}[]
    \centering
    \includegraphics[width=0.99\columnwidth]{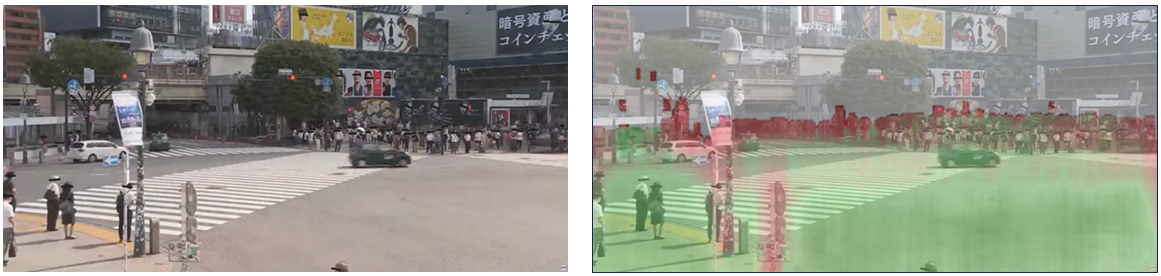}
    \caption{Left: A stationary scene observed by an automatic pedestrian detector. Right: A heatmap indicating varying average confidence in the automatic pedestrian detector depending on the pedestrian's location in the frame.}
    \label{fig:scene-and-heatmap}
\end{figure}

First, we analyze the factors affecting the performance of state-of-the-art automatic pedestrian detection systems using video footage from Shibuya Crossing, Broadway, and Castro Street. 
We show that the performance of state-of-the-art neural network-based pedestrian detection systems (Faster R-CNN \cite{ren2016fasterrcnnrealtimeobject}, YOLOv3 \cite{redmon2018yolov3incrementalimprovement}, SSD \cite{liu2016ssd}, DiffusionDet \cite{chen2023diffusiondet}, RTMDet \cite{lyu2022rtmdet}) is dependent on an observed pedestrian's angle, distance, and height, relative to the object detector camera, as well as the ambient light. 

We then demonstrate that in a still scene of pedestrian traffic observed by an automatic pedestrian detection system, the average confidence of the system's pedestrian detections is dependent on the location of the pedestrian when they are detected. 
This demonstrates the existence of "blind spots" of reduced confidence, which could be leveraged by pedestrians to protect their privacy, avoiding detection by walking a crafted path through a pedestrian detection system's weak detection areas (see Fig. \ref{fig:scene-and-heatmap}).

Based on these insights, we propose the Location-based Privacy Enhancing Technique (L-PET), which enables a pedestrian to traverse a scene observed by an automatic pedestrian detector, with minimal/no detection. 
We demonstrate that L-PET generates a path through the scene with lower max/average confidence than direct and random paths through the scene.
We tested L-PET in three scenes (Shibuya Crossing, Broadway, Castro Street), using five object detectors (Faster R-CNN, YOLOv3, SSD, DiffusionDet, RTMDet), and we found that L-PET lowers the max confidence and average confidence by 0.09 and 0.13, 0.08 and 0.07, 0.06 and 0.04, 0.06 and 0.05, and 0.08 and 0.09, on average for Faster R-CNN, YOLOv3, SSD, DiffusionDet, and RTMDet, respectively.

To counteract this, we propose the Location-Based Adaptive Threshold (L-BAT), a dynamic location-dependent detection threshold (compared to a universal detection threshold set by the object detector).
The threshold is determined according to a "heatmap" of areas with stronger and weaker detection confidence.
This produces an adaptive threshold which: (1) increases the average confidence of a pedestrian detection system, and (2) compensates for the degradation in performance stemming from angle-based and location-based vulnerabilities, improving the detection rate and making the confidence of pedestrian detection systems more stable over varying angles and locations of pedestrians.

\textbf{Contributions.} 
Our contributions are as follows: (1) We demonstrate an easier way to evade detection in automatic pedestrian detection systems, utilizing street video camera footage, by leveraging "blind spot" areas with lower detection confidence, rather than requiring adversarial patches, and present and evaluate a privacy enhancing technique that leverages these blind spots. (2) We create a framework intended to secure automatic pedestrian detection systems against these vulnerabilities, wrapping an object detector with a location-dependent threshold, rather than a universal threshold mechanism that is currently used. 
The countermeasure shrinks the presented technique's max and average confidence reduction from 0.09 to 0.0, and from 0.13 to 0.08, respectively, on Faster R-CNN.

\textbf{Ethics.}
In this paper, we utilized publicly available video footage of pedestrians in public areas. As part of the necessary procedures for ensuring ethical research with human subjects, we requested and obtained Institutional Review Board (IRB) approval regarding the usage, storage, and disposal of the data utilized in this paper. To protect the privacy of subjects in the video footage, the footage and derived experimental data are stored on an access-controlled database limited to only the researchers. 

\textbf{Structure.}
The structure of this paper is as follows; in Section \ref{sec:threat-model} we review the threat model. In Section \ref{sec:analysis} we review our analysis. In Sections \ref{sec:evaluation-lpea} and \ref{sec:evaluation-lbat} we present and evaluate L-PET and L-BAT's performance. We review related work in Section \ref{sec:related-works}, discuss limitations in Section \ref{sec:limitations}, and conclude the paper and propose future work in Section \ref{sec:conclusion}.

\section{Threat Model}
\label{sec:threat-model}

In this section, we present our threat model. 
The main actor in our threat model is a privacy-concerned \textit{pedestrian}, who wishes to leverage the location-based vulnerabilities of an automatic pedestrian detector to find a path across the scene with either minimal confidence or no detection. The pedestrian would then follow the path through the scene to achieve the lack of detection. The pedestrian's motivation is to proactively protect their privacy from potential misuse of surveillance systems. For instance, the pedestrian is motivated to evade detection in areas covered by video cameras that broadcast their footage to the internet unrestricted; many global cities have extensive use of surveillance cameras for pedestrian detection \cite{surveilledcities}.

\textbf{Pedestrian Assumptions.} 
We assume the pedestrian is able to obtain footage of the observed scene, and knows the object detector used in the surveillance system. These assumptions are usually considered in adversarial machine learning research \cite{zhou2018invisible,Komkov_2021,zolfi2022adversarial,sitawarin2018darts,Zhao2019SeeingIB,274691,athalye2018synthesizing,Chen_2019,eykholt2018robust,thys2019fooling}.
We also assume that the pedestrian has access to footage recorded by the pedestrian detection system (e.g., access to a live stream on the Internet).
In addition, we assume the pedestrian can create a heatmap of the surveillance system's confidence scores for different areas in the scene, given the needed footage and information about the surveillance system's object detector.
Finally, we assume the pedestrian has access to the physical scene, and can move through the scene in different positions.

\textbf{Steps of the Technique.} The steps of the privacy enhancing technique employed by the pedestrian are as follows: (1) The pedestrian creates a heatmap, using an object detector and video footage of the scene. We discuss the heatmap creation process in detail in Section \ref{sec:lpea-methodology}. (2) The pedestrian then uses the heatmap to determine a path between two areas captured in the video footage. (3) The pedestrian can finally walk through the determined path and avoid detection.

\textbf{Significance:} The significance of our threat model compared to related works is as follows:
(1) The pedestrian does not actively target the object detector camera with external tools, like projectors, lasers, or infrared light (compared to \cite{zhou2018invisible, lovisotto2021slap, sato2024invisible, shen2019vla, nassi2020phantom}), which would raise the suspicion of observers of the footage and/or individuals located in the scene.
(2) The technique preparation is simpler than those presented in related works. 
The pedestrian does not require sophisticated perturbation methods to produce an adversarial patch (compared to \cite{kurakin2016adversarial,sharif2016accessorize,Komkov_2021,ijcai2021p173,zolfi2022adversarial,sitawarin2018darts,Zhao2019SeeingIB,274691,athalye2018synthesizing,Chen_2019,eykholt2018robust,thys2019fooling,wu2020making,xu2020adversarial,Hu2022Adversarial,wang2020avdpattern,eykholt2018physicaladversarialexamplesobject,lee2019physicaladversarialpatchesobject,Yang_Tsai_Yu_Ho_Jin_2020,zolfi2020translucentpatchphysicaluniversal,hoory2020dynamicadversarialpatchevading}). 
Instead, the pedestrian calculates a confidence heatmap to generate a path for traversing the scene. 
In addition, the technique does not require white-box access to the target surveillance system.
\begin{table*}[th!]
\centering
\caption{Object detector confidence for each recorded position (height, distance, angle). Each row represents a configuration of camera height and distance, and each column represents an angle the person stood at relative to the camera. The cells representing positions which recorded average confidences of \textgreater{}0.8, 0.6-0.8, and \textless{}0.6 are presented in green, yellow, and red, respectively.}
\label{tab:standing-results-combined}
\resizebox{2.0\columnwidth}{!}{%
\begin{minipage}{2.4\columnwidth}
\subcaption{Faster R-CNN}
\label{tab:standing-results-faster}
\begin{subtable}{2.4\columnwidth}
\begin{tabular}{|cccccccccc|}
\multicolumn{10}{c}{Lab Conditions}                                                                                                                                                                                                                                                                                                                                                                                                                                                                       \\ \hline
\multicolumn{1}{|c|}{}                         & \multicolumn{1}{c|}{}                           & \multicolumn{8}{c|}{Angle}                                                                                                                                                                                                                                                                                                                                                                               \\ \cline{3-10} 
\multicolumn{1}{|c|}{\multirow{-2}{*}{Height}} & \multicolumn{1}{c|}{\multirow{-2}{*}{Distance}} & \multicolumn{1}{c|}{0\degree}                            & \multicolumn{1}{c|}{45\degree}                           & \multicolumn{1}{c|}{90\degree}                           & \multicolumn{1}{c|}{135\degree}                          & \multicolumn{1}{c|}{180\degree}                          & \multicolumn{1}{c|}{225\degree}                          & \multicolumn{1}{c|}{270\degree}                          & 315\degree                          \\
\hline
\end{tabular}%
\hspace{.1pt}
\begin{tabular}{|cccccccccc|}
\multicolumn{10}{c}{Morning}                                                                                                                                                                                                                                                                                                                                                                                                                                                                              \\ \hline
\multicolumn{1}{|c|}{}                         & \multicolumn{1}{c|}{}                           & \multicolumn{8}{c|}{Angle}                                                                                                                                                                                                                                                                                                                                                                               \\ \cline{3-10} 
\multicolumn{1}{|c|}{\multirow{-2}{*}{Height}} & \multicolumn{1}{c|}{\multirow{-2}{*}{Distance}} & \multicolumn{1}{c|}{0\degree}                            & \multicolumn{1}{c|}{45\degree}                           & \multicolumn{1}{c|}{90\degree}                           & \multicolumn{1}{c|}{135\degree}                          & \multicolumn{1}{c|}{180\degree}                          & \multicolumn{1}{c|}{225\degree}                          & \multicolumn{1}{c|}{270\degree}                          & 315\degree                          \\
\hline
\end{tabular}%
\newline
\vspace*{0.03 cm}
\newline
\noindent
\begin{tabular}{cccccccccc}
\multicolumn{10}{c}{Afternoon}                                                                                                                                                                                                                                                                                                                                                                                                                                                                                                   \\ \hline
\multicolumn{1}{|c|}{}                         & \multicolumn{1}{c|}{}                           & \multicolumn{8}{c|}{Angle}                                                                                                                                                                                                                                                                                                                                                                                                    \\ \cline{3-10} 
\multicolumn{1}{|c|}{\multirow{-2}{*}{Height}} & \multicolumn{1}{c|}{\multirow{-2}{*}{Distance}} & \multicolumn{1}{c|}{0\degree}                            & \multicolumn{1}{c|}{45\degree}                           & \multicolumn{1}{c|}{90\degree}                           & \multicolumn{1}{c|}{135\degree}                          & \multicolumn{1}{c|}{180\degree}                          & \multicolumn{1}{c|}{225\degree}                          & \multicolumn{1}{c|}{270\degree}                          & \multicolumn{1}{c|}{315\degree}                          \\
\hline
\end{tabular}%
\hspace{0.1pt}
\begin{tabular}{cccccccccc}
\multicolumn{10}{c}{Night}                                                                                                                                                                                                                                                                                                                                                                                                                                                                                                       \\ \hline
\multicolumn{1}{|c|}{}                         & \multicolumn{1}{c|}{}                           & \multicolumn{8}{c|}{Angle}                                                                                                                                                                                                                                                                                                                                                                                                    \\ \cline{3-10} 
\multicolumn{1}{|c|}{\multirow{-2}{*}{Height}} & \multicolumn{1}{c|}{\multirow{-2}{*}{Distance}} & \multicolumn{1}{c|}{0\degree}                            & \multicolumn{1}{c|}{45\degree}                           & \multicolumn{1}{c|}{90\degree}                           & \multicolumn{1}{c|}{135\degree}                          & \multicolumn{1}{c|}{180\degree}                          & \multicolumn{1}{c|}{225\degree}                          & \multicolumn{1}{c|}{270\degree}                          & \multicolumn{1}{c|}{315\degree}                          \\
\hline
\end{tabular}%
\end{subtable}
\end{minipage}
}
\newline
\vspace*{0.3 cm}
\newline
\resizebox{2.0\columnwidth}{!}{%
\begin{minipage}{2.4\columnwidth}
\subcaption{YOLOv3}
\label{tab:standing-results-yolo}
\begin{subtable}{2.4\columnwidth}
\begin{tabular}{cccccccccc}
\multicolumn{10}{c}{Lab Conditions}                                                                                                                                                                                                                                                                                                                                                                                                                                                                                              \\ \hline
\multicolumn{1}{|c|}{}                         & \multicolumn{1}{c|}{}                           & \multicolumn{8}{c|}{Angle}                                                                                                                                                                                                                                                                                                                                                                                                    \\ \cline{3-10} 
\multicolumn{1}{|c|}{\multirow{-2}{*}{Height}} & \multicolumn{1}{c|}{\multirow{-2}{*}{Distance}} & \multicolumn{1}{c|}{0\degree}                            & \multicolumn{1}{c|}{45\degree}                           & \multicolumn{1}{c|}{90\degree}                           & \multicolumn{1}{c|}{135\degree}                          & \multicolumn{1}{c|}{180\degree}                          & \multicolumn{1}{c|}{225\degree}                          & \multicolumn{1}{c|}{270\degree}                          & \multicolumn{1}{c|}{315\degree}                          \\
\hline
\end{tabular}%
\hspace{.1pt}
\begin{tabular}{cccccccccc}
\multicolumn{10}{c}{Morning}                                                                                                                                                                                                                                                                                                                                                                                                                                                                                                     \\ \hline
\multicolumn{1}{|c|}{}                         & \multicolumn{1}{c|}{}                           & \multicolumn{8}{c|}{Angle}                                                                                                                                                                                                                                                                                                                                                                                                    \\ \cline{3-10} 
\multicolumn{1}{|c|}{\multirow{-2}{*}{Height}} & \multicolumn{1}{c|}{\multirow{-2}{*}{Distance}} & \multicolumn{1}{c|}{0\degree}                     & \multicolumn{1}{c|}{45\degree}                    & \multicolumn{1}{c|}{90\degree}                    & \multicolumn{1}{c|}{135\degree}                   & \multicolumn{1}{c|}{180\degree}                   & \multicolumn{1}{c|}{225\degree}                   & \multicolumn{1}{c|}{270\degree}                   & \multicolumn{1}{c|}{315\degree}                   \\
\hline
\end{tabular}%
\newline
\vspace*{0.03 cm}
\newline
\noindent
\begin{tabular}{cccccccccc}
\multicolumn{10}{c}{Afternoon}                                                                                                                                                                                                                                                                                                                                                                                                                                                                                                   \\ \hline
\multicolumn{1}{|c|}{}                         & \multicolumn{1}{c|}{}                           & \multicolumn{8}{c|}{Angle}                                                                                                                                                                                                                                                                                                                                                                                                    \\ \cline{3-10} 
\multicolumn{1}{|c|}{\multirow{-2}{*}{Height}} & \multicolumn{1}{c|}{\multirow{-2}{*}{Distance}} & \multicolumn{1}{c|}{0\degree}                            & \multicolumn{1}{c|}{45\degree}                           & \multicolumn{1}{c|}{90\degree}                           & \multicolumn{1}{c|}{135\degree}                          & \multicolumn{1}{c|}{180\degree}                          & \multicolumn{1}{c|}{225\degree}                          & \multicolumn{1}{c|}{270\degree}                          & \multicolumn{1}{c|}{315\degree}                          \\
\hline
\end{tabular}%
\hspace{0.1pt}
\begin{tabular}{cccccccccc}
\multicolumn{10}{c}{Night}                                                                                                                                                                                                                                                                                                                                                                                                                                                                                                       \\ \hline
\multicolumn{1}{|c|}{}                         & \multicolumn{1}{c|}{}                           & \multicolumn{8}{c|}{Angle}                                                                                                                                                                                                                                                                                                                                                                                                    \\ \cline{3-10} 
\multicolumn{1}{|c|}{\multirow{-2}{*}{Height}} & \multicolumn{1}{c|}{\multirow{-2}{*}{Distance}} & \multicolumn{1}{c|}{0\degree}                            & \multicolumn{1}{c|}{45\degree}                           & \multicolumn{1}{c|}{90\degree}                           & \multicolumn{1}{c|}{135\degree}                          & \multicolumn{1}{c|}{180\degree}                          & \multicolumn{1}{c|}{225\degree}                          & \multicolumn{1}{c|}{270\degree}                          & \multicolumn{1}{c|}{315\degree}                          \\
\hline
\end{tabular}%
\end{subtable}
\end{minipage}
}
\end{table*}

\section {Analysis}
\label{sec:analysis}

In this section, we analyze two phenomena related to the performance of AI-based pedestrian detectors: (1) the behavior (detection confidence) of AI-based pedestrian detectors when detecting a pedestrian standing at different \textit{angles, distances, and heights} from the viewpoint of the pedestrian detector, and (2) the behavior of AI-based pedestrian detectors when detecting pedestrians in different \textit{locations} in the pedestrian detector's view.

\subsection{Effect of Pedestrian Position on Pedestrian Detection Systems}
\label{sec:position-analysis}

\begin{table*}[th!]
\centering
\caption{Object detector confidence for each recorded position (height, distance, angle). Each row represents a configuration of camera height and distance, and each column represents an angle the person stood at relative to the camera. The cells representing positions which recorded average confidences of \textgreater{}0.8, 0.6-0.8, and \textless{}0.6 are presented in green, yellow, and red, respectively.}
\label{tab:standing-results-combined-2}
\resizebox{2.0\columnwidth}{!}{%
\begin{minipage}{2.4\columnwidth}
\subcaption{SSD}
\label{tab:standing-results-ssd}
\begin{subtable}{2.4\columnwidth}
\begin{tabular}{cccccccccc}
\multicolumn{10}{c}{Lab Conditions}                                                                                                                                                                                                                                                                                                                                                                                                                                                                                              \\ \hline
\multicolumn{1}{|c|}{}                         & \multicolumn{1}{c|}{}                           & \multicolumn{8}{c|}{Angle}                                                                                                                                                                                                                                                                                                                                                                                                    \\ \cline{3-10} 
\multicolumn{1}{|c|}{\multirow{-2}{*}{Height}} & \multicolumn{1}{c|}{\multirow{-2}{*}{Distance}} & \multicolumn{1}{c|}{0\degree}                            & \multicolumn{1}{c|}{45\degree}                           & \multicolumn{1}{c|}{90\degree}                           & \multicolumn{1}{c|}{135\degree}                          & \multicolumn{1}{c|}{180\degree}                          & \multicolumn{1}{c|}{225\degree}                          & \multicolumn{1}{c|}{270\degree}                          & \multicolumn{1}{c|}{315\degree}                          \\
\hline
\end{tabular}%
\hspace{.1pt}
\begin{tabular}{cccccccccc}
\multicolumn{10}{c}{Morning}                                                                                                                                                                                                                                                                                                                                                                                                                                                                                                     \\ \hline
\multicolumn{1}{|c|}{}                         & \multicolumn{1}{c|}{}                           & \multicolumn{8}{c|}{Angle}                                                                                                                                                                                                                                                                                                                                                                                                    \\ \cline{3-10} 
\multicolumn{1}{|c|}{\multirow{-2}{*}{Height}} & \multicolumn{1}{c|}{\multirow{-2}{*}{Distance}} & \multicolumn{1}{c|}{0\degree}                            & \multicolumn{1}{c|}{45\degree}                           & \multicolumn{1}{c|}{90\degree}                           & \multicolumn{1}{c|}{135\degree}                          & \multicolumn{1}{c|}{180\degree}                          & \multicolumn{1}{c|}{225\degree}                          & \multicolumn{1}{c|}{270\degree}                          & \multicolumn{1}{c|}{315\degree}                          \\
\hline
\end{tabular}%
\newline
\vspace*{0.03 cm}
\newline
\noindent
\begin{tabular}{cccccccccc}
\multicolumn{10}{c}{Afternoon}                                                                                                                                                                                                                                                                                                                                                                                                                                                                                                   \\ \hline
\multicolumn{1}{|c|}{}                         & \multicolumn{1}{c|}{}                           & \multicolumn{8}{c|}{Angle}                                                                                                                                                                                                                                                                                                                                                                                                    \\ \cline{3-10} 
\multicolumn{1}{|c|}{\multirow{-2}{*}{Height}} & \multicolumn{1}{c|}{\multirow{-2}{*}{Distance}} & \multicolumn{1}{c|}{0\degree}                            & \multicolumn{1}{c|}{45\degree}                           & \multicolumn{1}{c|}{90\degree}                           & \multicolumn{1}{c|}{135\degree}                          & \multicolumn{1}{c|}{180\degree}                          & \multicolumn{1}{c|}{225\degree}                          & \multicolumn{1}{c|}{270\degree}                          & \multicolumn{1}{c|}{315\degree}                          \\
\hline
\end{tabular}%
\hspace{0.1pt}
\begin{tabular}{cccccccccc}
\multicolumn{10}{c}{Night}                                                                                                                                                                                                                                                                                                                                                                                                                                                                                                       \\ \hline
\multicolumn{1}{|c|}{}                         & \multicolumn{1}{c|}{}                           & \multicolumn{8}{c|}{Angle}                                                                                                                                                                                                                                                                                                                                                                                                    \\ \cline{3-10} 
\multicolumn{1}{|c|}{\multirow{-2}{*}{Height}} & \multicolumn{1}{c|}{\multirow{-2}{*}{Distance}} & \multicolumn{1}{c|}{0\degree}                            & \multicolumn{1}{c|}{45\degree}                           & \multicolumn{1}{c|}{90\degree}                           & \multicolumn{1}{c|}{135\degree}                          & \multicolumn{1}{c|}{180\degree}                          & \multicolumn{1}{c|}{225\degree}                          & \multicolumn{1}{c|}{270\degree}                          & \multicolumn{1}{c|}{315\degree}                          \\
\hline
\end{tabular}%
\end{subtable}
\end{minipage}
}
\newline
\vspace*{0.3 cm}
\newline
\resizebox{2.0\columnwidth}{!}{%
\begin{minipage}{2.4\columnwidth}
\subcaption{RTMDet}
\label{tab:standing-results-rtm}
\begin{subtable}{2.4\columnwidth}
\begin{tabular}{cccccccccc}
\multicolumn{10}{c}{Lab Conditions}                                                                                                                                                                                                                                                                                                                                                                                                                                                                                              \\ \hline
\multicolumn{1}{|c|}{}                         & \multicolumn{1}{c|}{}                           & \multicolumn{8}{c|}{Angle}                                                                                                                                                                                                                                                                                                                                                                                                    \\ \cline{3-10} 
\multicolumn{1}{|c|}{\multirow{-2}{*}{Height}} & \multicolumn{1}{c|}{\multirow{-2}{*}{Distance}} & \multicolumn{1}{c|}{0\degree}                            & \multicolumn{1}{c|}{45\degree}                           & \multicolumn{1}{c|}{90\degree}                           & \multicolumn{1}{c|}{135\degree}                          & \multicolumn{1}{c|}{180\degree}                          & \multicolumn{1}{c|}{225\degree}                          & \multicolumn{1}{c|}{270\degree}                          & \multicolumn{1}{c|}{315\degree}                          \\
\hline
\end{tabular}%
\hspace{.1pt}
\begin{tabular}{cccccccccc}
\multicolumn{10}{c}{Morning}                                                                                                                                                                                                                                                                                                                                                                                                                                                                                                     \\ \hline
\multicolumn{1}{|c|}{}                         & \multicolumn{1}{c|}{}                           & \multicolumn{8}{c|}{Angle}                                                                                                                                                                                                                                                                                                                                                                                                    \\ \cline{3-10} 
\multicolumn{1}{|c|}{\multirow{-2}{*}{Height}} & \multicolumn{1}{c|}{\multirow{-2}{*}{Distance}} & \multicolumn{1}{c|}{0\degree}                            & \multicolumn{1}{c|}{45\degree}                           & \multicolumn{1}{c|}{90\degree}                           & \multicolumn{1}{c|}{135\degree}                          & \multicolumn{1}{c|}{180\degree}                          & \multicolumn{1}{c|}{225\degree}                          & \multicolumn{1}{c|}{270\degree}                          & \multicolumn{1}{c|}{315\degree}                          \\
\hline
\end{tabular}%
\newline
\vspace*{0.03 cm}
\newline
\noindent
\begin{tabular}{cccccccccc}
\multicolumn{10}{c}{Afternoon}                                                                                                                                                                                                                                                                                                                                                                                                                                                                                                   \\ \hline
\multicolumn{1}{|c|}{}                         & \multicolumn{1}{c|}{}                           & \multicolumn{8}{c|}{Angle}                                                                                                                                                                                                                                                                                                                                                                                                    \\ \cline{3-10} 
\multicolumn{1}{|c|}{\multirow{-2}{*}{Height}} & \multicolumn{1}{c|}{\multirow{-2}{*}{Distance}} & \multicolumn{1}{c|}{0\degree}                            & \multicolumn{1}{c|}{45\degree}                           & \multicolumn{1}{c|}{90\degree}                           & \multicolumn{1}{c|}{135\degree}                          & \multicolumn{1}{c|}{180\degree}                          & \multicolumn{1}{c|}{225\degree}                          & \multicolumn{1}{c|}{270\degree}                          & \multicolumn{1}{c|}{315\degree}                          \\
\hline
\end{tabular}%
\hspace{0.1pt}
\begin{tabular}{cccccccccc}
\multicolumn{10}{c}{Night}                                                                                                                                                                                                                                                                                                                                                                                                                                                                                                       \\ \hline
\multicolumn{1}{|c|}{}                         & \multicolumn{1}{c|}{}                           & \multicolumn{8}{c|}{Angle}                                                                                                                                                                                                                                                                                                                                                                                                    \\ \cline{3-10} 
\multicolumn{1}{|c|}{\multirow{-2}{*}{Height}} & \multicolumn{1}{c|}{\multirow{-2}{*}{Distance}} & \multicolumn{1}{c|}{0\degree}                            & \multicolumn{1}{c|}{45\degree}                           & \multicolumn{1}{c|}{90\degree}                           & \multicolumn{1}{c|}{135\degree}                          & \multicolumn{1}{c|}{180\degree}                          & \multicolumn{1}{c|}{225\degree}                          & \multicolumn{1}{c|}{270\degree}                          & \multicolumn{1}{c|}{315\degree}                          \\
\hline
\end{tabular}%
\end{subtable}
\end{minipage}
}
\end{table*}

In this analysis, we investigate the effect of a pedestrian’s position (distance, height, and angle) on an AI-based pedestrian detector’s confidence.

\textbf{Overall Experimental Setup:} 
We analyzed a person standing at three different distances (2.5m, 5m, 10m), and eight different angles (0°, 45°, 90°, 135°, 180°, 225°, 270°, 315°), from a camera positioned at three different heights (0.6m, 1.8m, 2.4m).
The positioning of the camera and person can be seen in Fig. \ref{fig:exp-setup-visualization} in the appendix.

For each position (height, angle, distance) we recorded the person standing still for one minute, and fed the footage to a Faster R-CNN object detector.
For each one minute video, we recorded the confidence values of the bounding boxes labeled “person“ by the object detector.
For each video, we then calculated the mean confidence of the object detector’s “person” bounding boxes.
In other words, given a person’s position (height, angle, distance), how confident is the object detector when detecting the person?

\subsubsection{Effect of Pedestrian Distance}
\label{sec:distance-analysis}

Here, we analyze the effect of a pedestrian's distance from a pedestrian detector's camera on the pedestrian detector's confidence.

\textbf{Experimental Setup:} We analyzed a person standing at three different distances (2.5m, 5m, 10m) from a camera whose footage was fed to a Faster R-CNN object detector. These experiments were repeated with the person standing at eight different angles (0°, 45°, 90°, 135°, 180°, 225°, 270°, 315°), and the camera positioned at three different heights (0.6m, 1.8m, 2.4m).

\textbf{Results:} The results of this analysis are presented in Table \ref{tab:standing-results-faster}. We found that when keeping other variables (angle, height) constant, the distance of the person resulted in varying confidence in the object detector, ranging from 0.56 to 1.0 in the most extreme case.

\begin{insight} \label{insight:distance}
The confidence of an object detector is dependent on the \textit{distance} of the person from the object detector.
\end{insight}

\begin{figure*}[th!]
    \centering
     \includegraphics[width=0.355\textwidth]{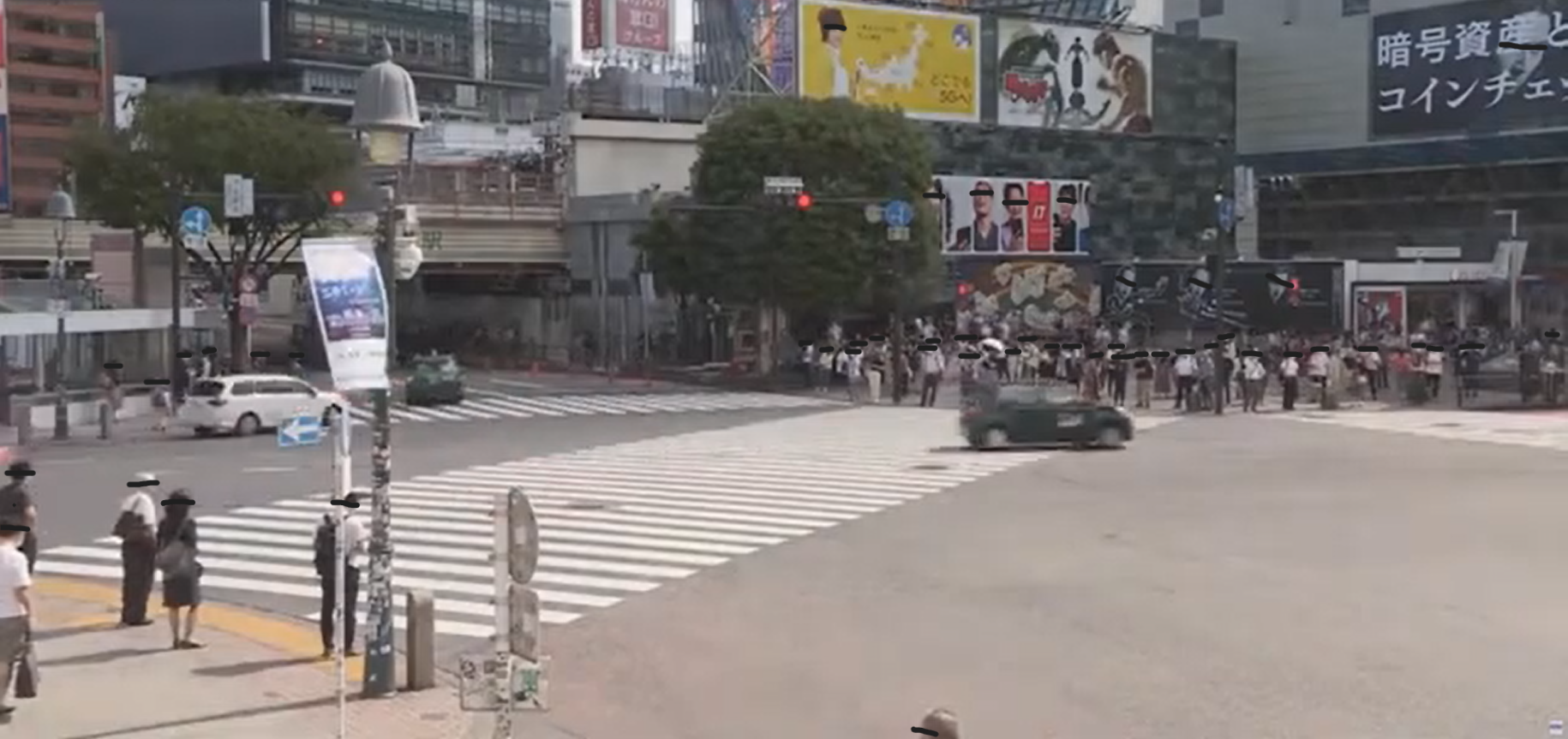}    
    \includegraphics[width=0.3\textwidth]{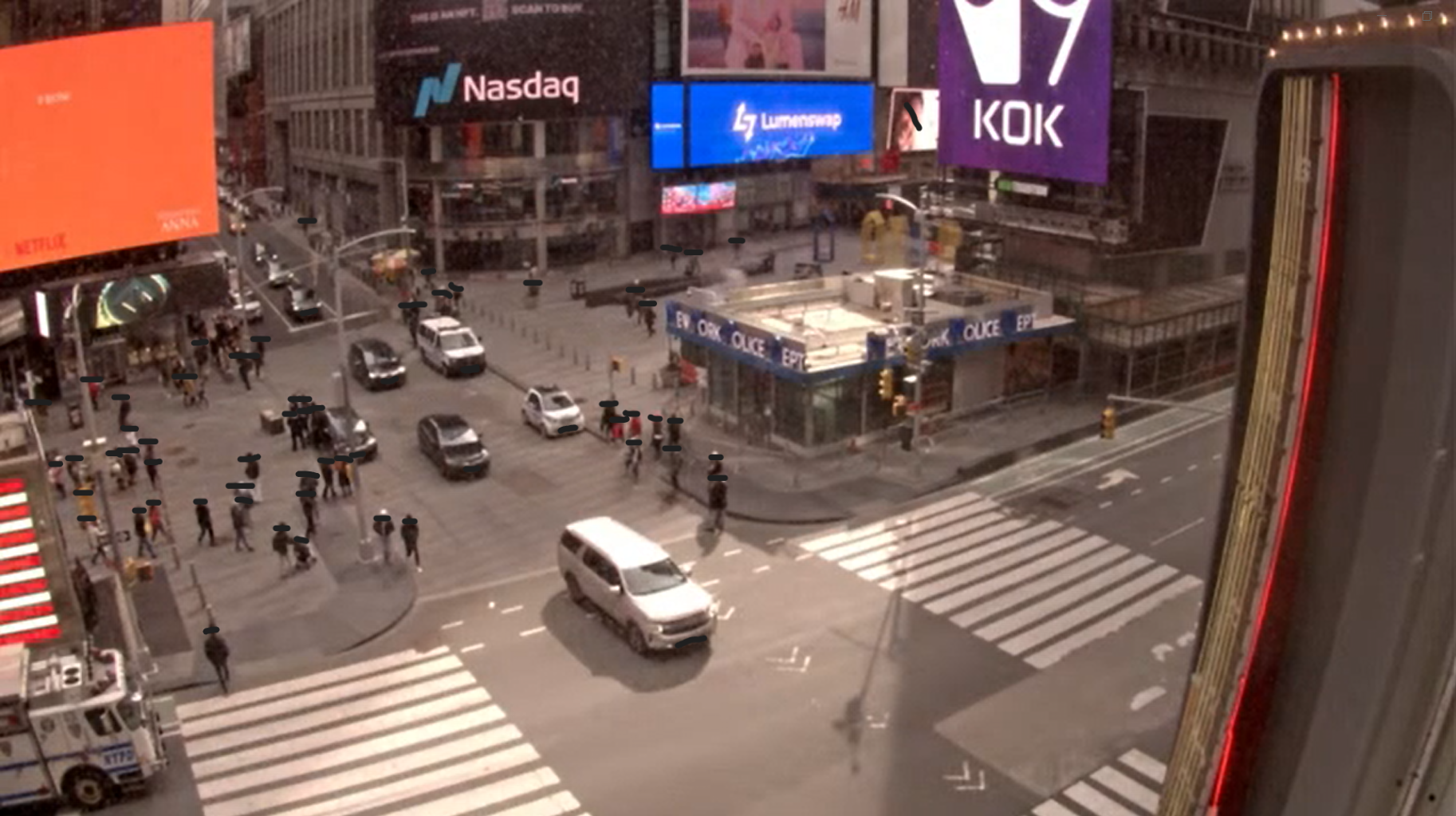}      
    \includegraphics[width=0.3\textwidth]{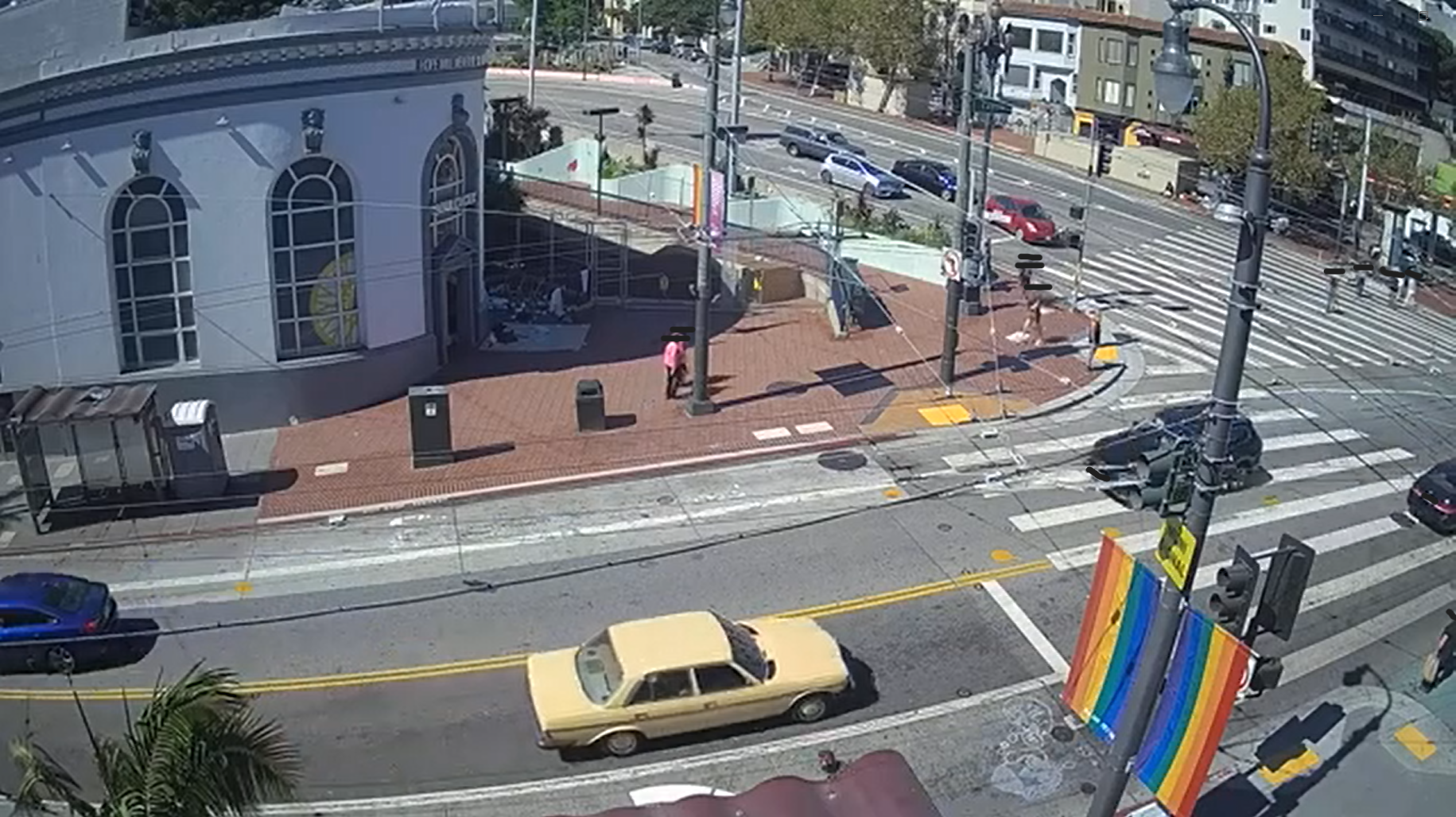}
    \caption{Observed scenes from (left to right): Shibuya Crossing, Broadway, and Castro Street.} 
    \label{fig:observed-scenes}     
\end{figure*}

\begin{figure*}[h]
\centering
       \begin{minipage}{0.99\textwidth}
       \centering  
          \includegraphics[width=0.16\textwidth]{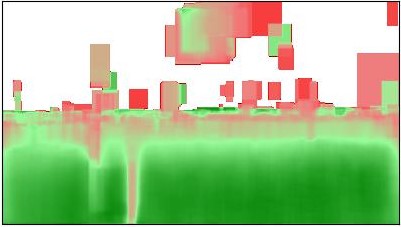} 
     \includegraphics[width=0.16\textwidth]{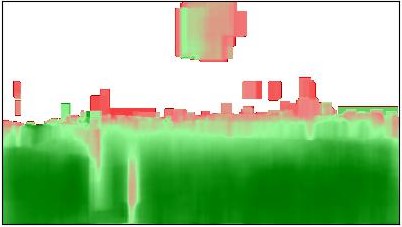}
      \includegraphics[width=0.16\textwidth]{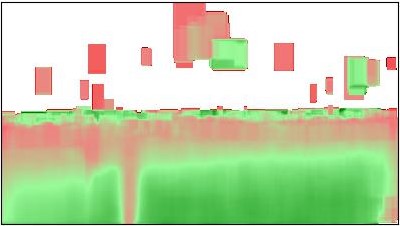} 
     \includegraphics[width=0.16\textwidth]{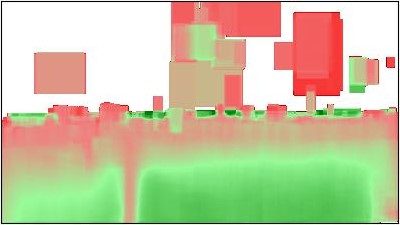}
      \includegraphics[width=0.16\textwidth]{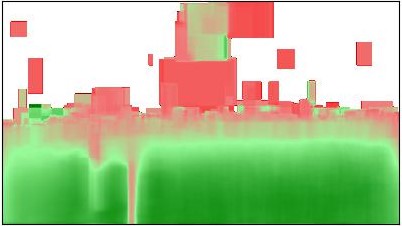}
      \includegraphics[width=0.16\textwidth]{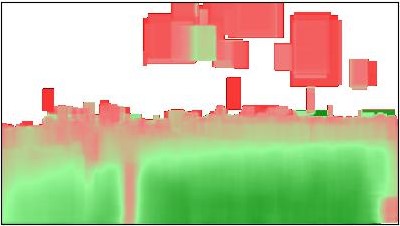}
      \subcaption{Faster R-CNN.} 
      \label{fig:shibuya-results-faster}
    \end{minipage} 
    \newline
       \begin{minipage}{0.99\textwidth}
       \centering  
          \includegraphics[width=0.16\textwidth]{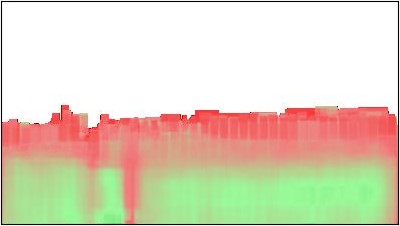} 
     \includegraphics[width=0.16\textwidth]{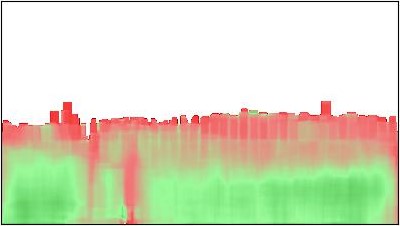}
      \includegraphics[width=0.16\textwidth]{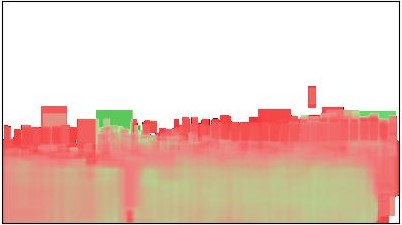} 
     \includegraphics[width=0.16\textwidth]{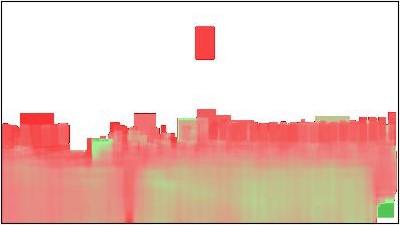}
      \includegraphics[width=0.16\textwidth]{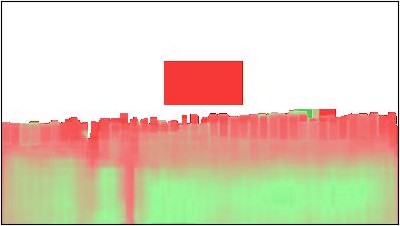}
      \includegraphics[width=0.16\textwidth]{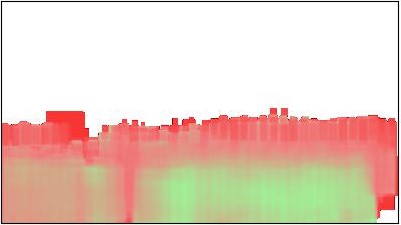}
      \subcaption{YOLOv3.} 
      \label{fig:shibuya-results-yolo}
    \end{minipage} 
    \newline
       \begin{minipage}{0.99\textwidth}
       \centering  
          \includegraphics[width=0.16\textwidth]{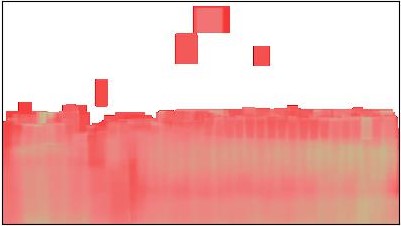} 
     \includegraphics[width=0.16\textwidth]{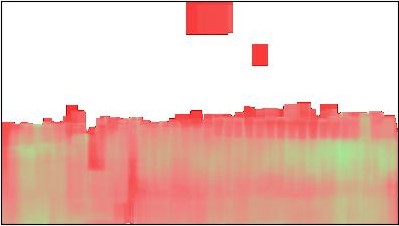}
      \includegraphics[width=0.16\textwidth]{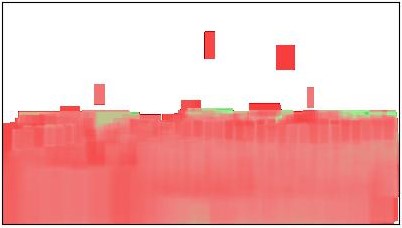} 
     \includegraphics[width=0.16\textwidth]{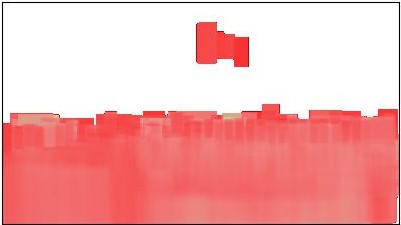}
      \includegraphics[width=0.16\textwidth]{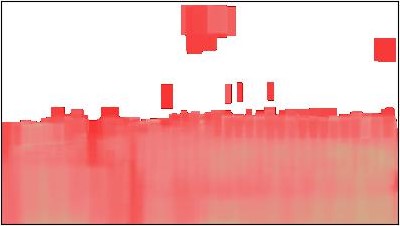}
      \includegraphics[width=0.16\textwidth]{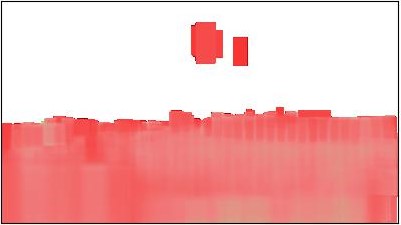}
      \subcaption{SSD.} 
      \label{fig:shibuya-results-ssd}
    \end{minipage}  
    \newline
       \begin{minipage}{0.99\textwidth}
       \centering  
          \includegraphics[width=0.16\textwidth]{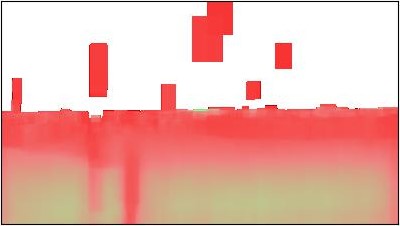} 
     \includegraphics[width=0.16\textwidth]{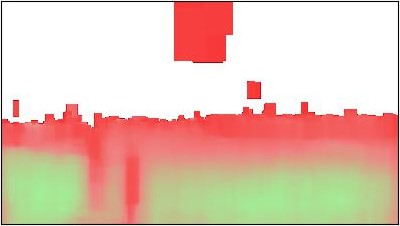}
      \includegraphics[width=0.16\textwidth]{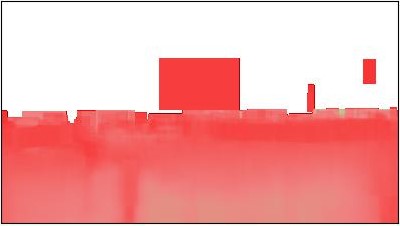} 
     \includegraphics[width=0.16\textwidth]{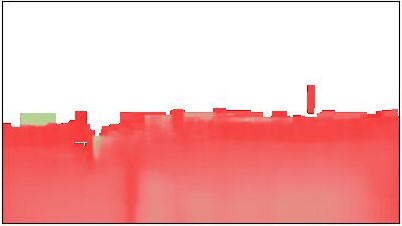}
      \includegraphics[width=0.16\textwidth]{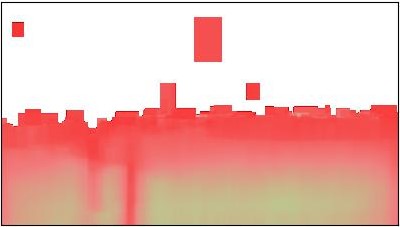}
      \includegraphics[width=0.16\textwidth]{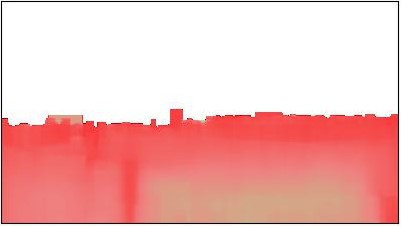}
      \subcaption{RTMDet.} 
      \label{fig:shibuya-results-rtm}
    \end{minipage} 
    \caption{The average confidence of pedestrian detection for each pixel in the observed area of Shibuya Crossing. The six videos, recorded in different lighting conditions (from left to right: daytime, daytime, afternoon, evening, night, and night), are presented separately.}
\label{fig:shibuya-results-combined}
\end{figure*}
\begin{figure*}[h]
\centering
       \begin{minipage}{0.99\textwidth}
       \centering  
          \includegraphics[width=0.16\textwidth]{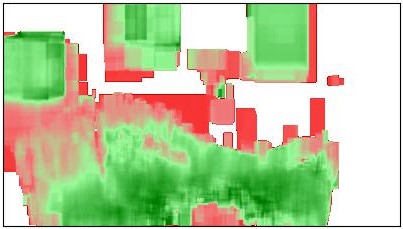} 
     \includegraphics[width=0.16\textwidth]{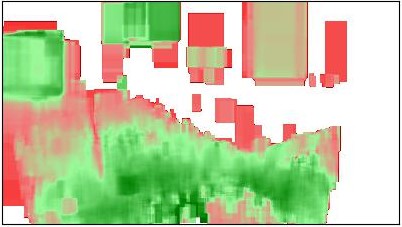}
      \includegraphics[width=0.16\textwidth]{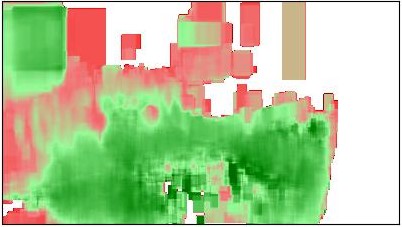} 
     \includegraphics[width=0.16\textwidth]{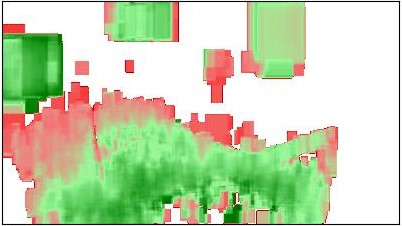}
      \includegraphics[width=0.16\textwidth]{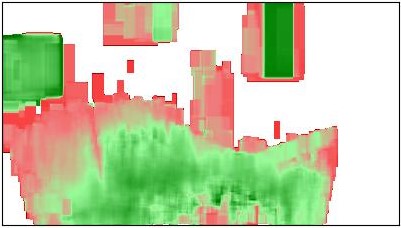}
      \includegraphics[width=0.16\textwidth]{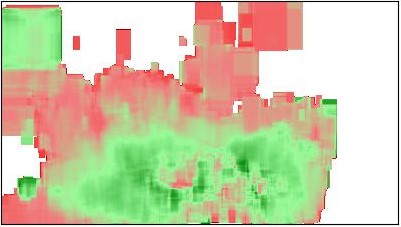}
      \subcaption{Faster R-CNN.} 
      \label{fig:new-york-results-faster}
    \end{minipage} 
    \newline
       \begin{minipage}{0.99\textwidth}
       \centering  
          \includegraphics[width=0.16\textwidth]{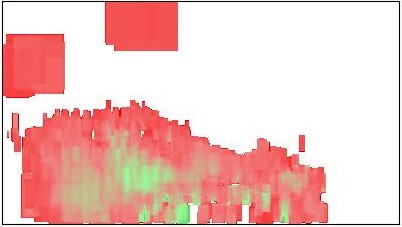} 
     \includegraphics[width=0.16\textwidth]{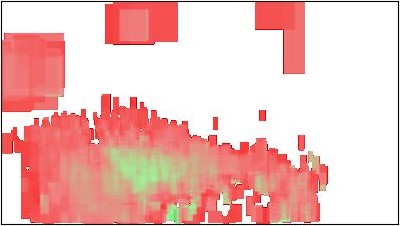}
      \includegraphics[width=0.16\textwidth]{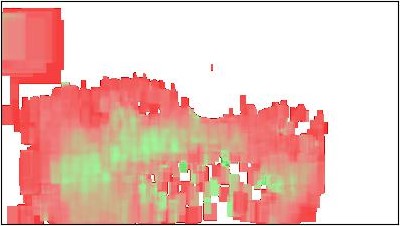} 
     \includegraphics[width=0.16\textwidth]{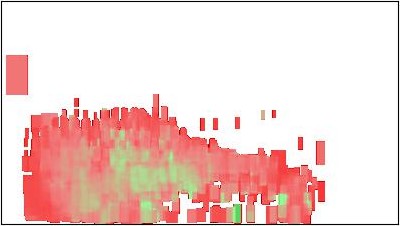}
      \includegraphics[width=0.16\textwidth]{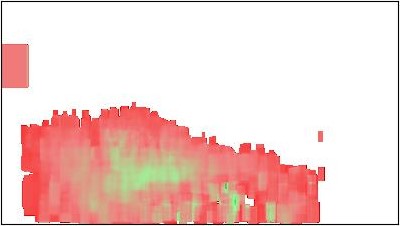}
      \includegraphics[width=0.16\textwidth]{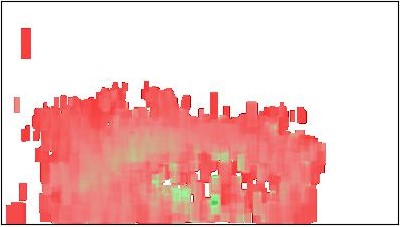}
      \subcaption{YOLOv3.} 
      \label{fig:new-york-results-yolo}
    \end{minipage} 
    \newline
       \begin{minipage}{0.99\textwidth}
       \centering  
          \includegraphics[width=0.16\textwidth]{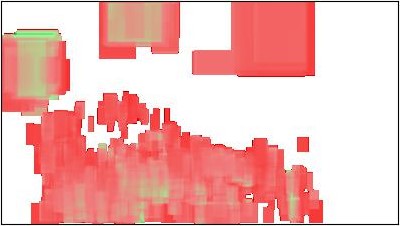} 
     \includegraphics[width=0.16\textwidth]{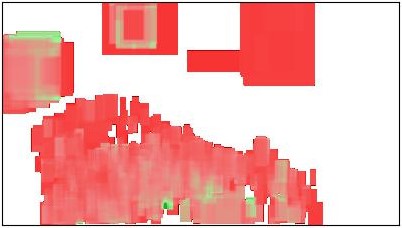}
      \includegraphics[width=0.16\textwidth]{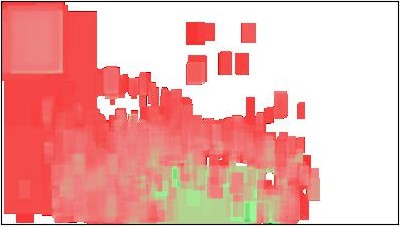} 
     \includegraphics[width=0.16\textwidth]{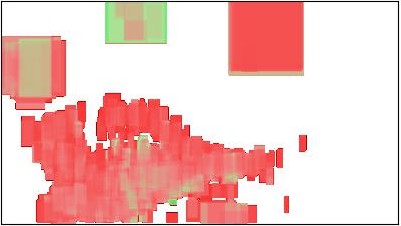}
      \includegraphics[width=0.16\textwidth]{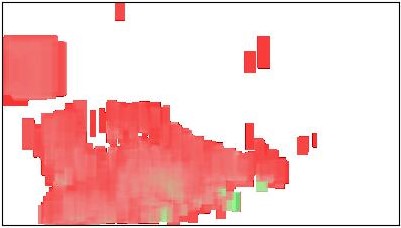}
      \includegraphics[width=0.16\textwidth]{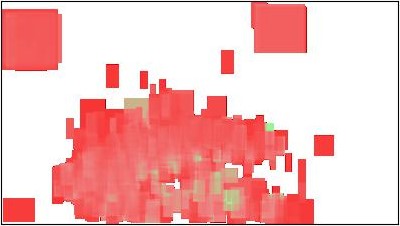}
      \subcaption{SSD.} 
      \label{fig:new-york-results-ssd}
    \end{minipage}  
    \newline
       \begin{minipage}{0.99\textwidth}
       \centering  
          \includegraphics[width=0.16\textwidth]{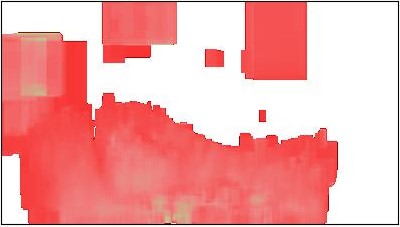} 
     \includegraphics[width=0.16\textwidth]{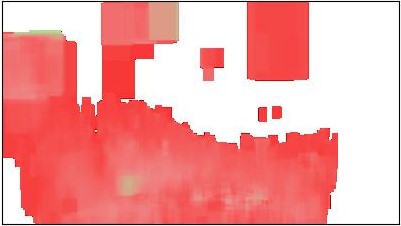}
      \includegraphics[width=0.16\textwidth]{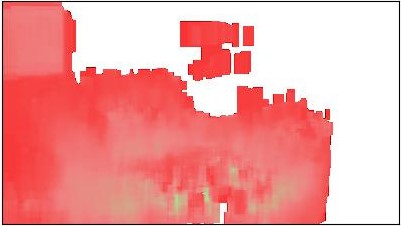} 
     \includegraphics[width=0.16\textwidth]{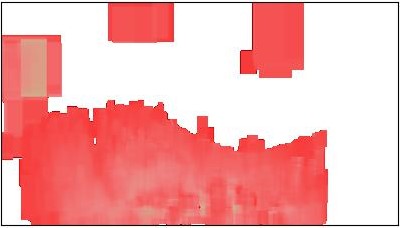}
      \includegraphics[width=0.16\textwidth]{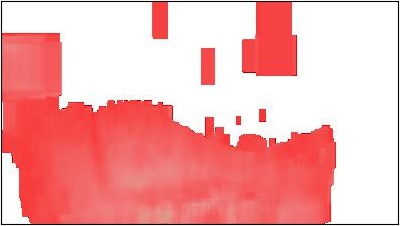}
      \includegraphics[width=0.16\textwidth]{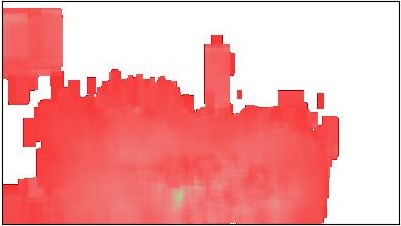}
      \subcaption{RTMDet.} 
      \label{fig:new-york-results-rtm}
    \end{minipage} 
    \caption{The average confidence of pedestrian detection for each pixel in the observed area of Broadway. The six videos, recorded in different lighting conditions (from left to right: daytime, daytime, daytime, night, night, and night), are presented separately.}
\label{fig:new-york-results-combined}
\end{figure*}
\begin{figure*}[h]
\centering
       \begin{minipage}{0.99\textwidth}
       \centering  
          \includegraphics[width=0.16\textwidth]{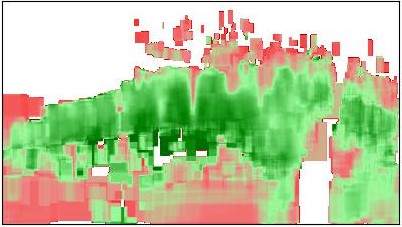} 
     \includegraphics[width=0.16\textwidth]{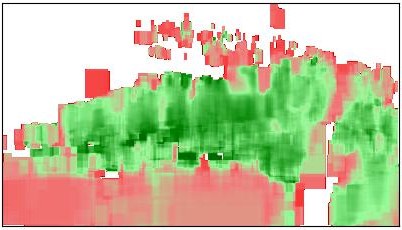}
      \includegraphics[width=0.16\textwidth]{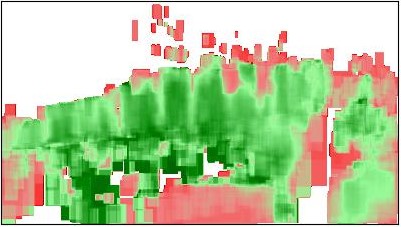} 
     \includegraphics[width=0.16\textwidth]{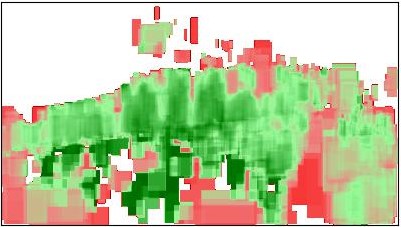}
      \includegraphics[width=0.16\textwidth]{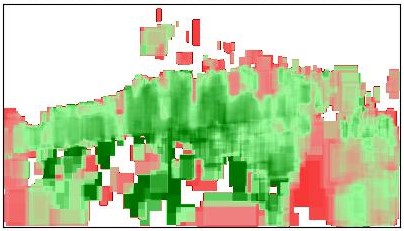}
      \subcaption{Faster R-CNN.} 
      \label{fig:castro-results-faster}
    \end{minipage} 
    \newline
       \begin{minipage}{0.99\textwidth}
       \centering  
          \includegraphics[width=0.16\textwidth]{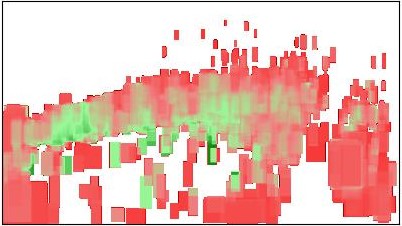} 
     \includegraphics[width=0.16\textwidth]{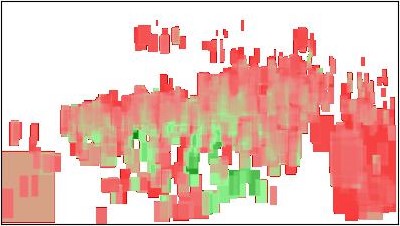}
      \includegraphics[width=0.16\textwidth]{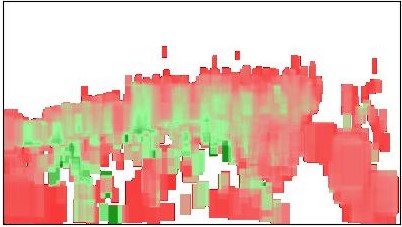} 
     \includegraphics[width=0.16\textwidth]{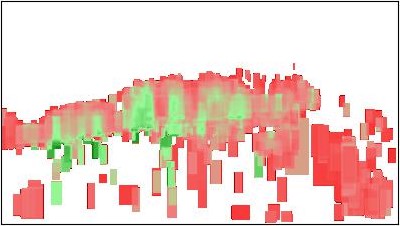}
      \includegraphics[width=0.16\textwidth]{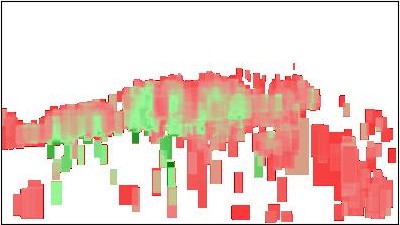}
      \subcaption{YOLOv3.} 
      \label{fig:castro-results-yolo}
    \end{minipage} 
    \newline
       \begin{minipage}{0.99\textwidth}
       \centering  
          \includegraphics[width=0.16\textwidth]{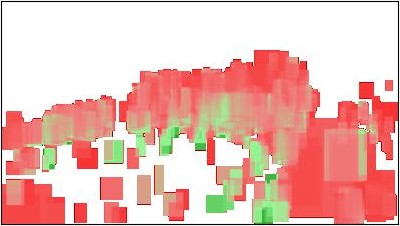} 
     \includegraphics[width=0.16\textwidth]{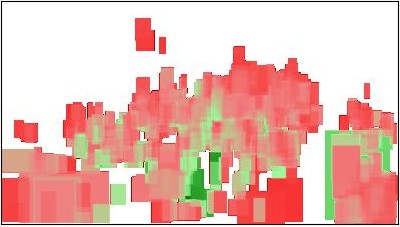}
      \includegraphics[width=0.16\textwidth]{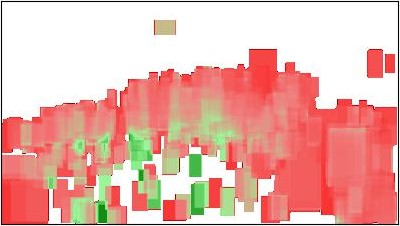} 
     \includegraphics[width=0.16\textwidth]{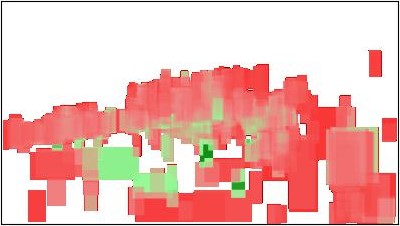}
      \includegraphics[width=0.16\textwidth]{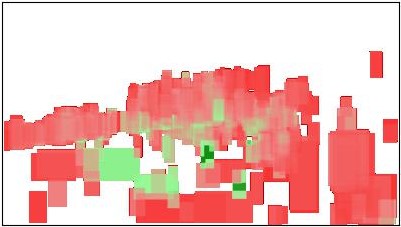}
      \subcaption{SSD.} 
      \label{fig:castro-results-ssd}
    \end{minipage}  
    \newline
       \begin{minipage}{0.99\textwidth}
       \centering  
          \includegraphics[width=0.16\textwidth]{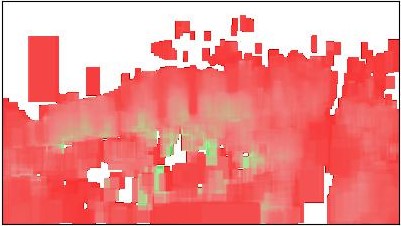} 
     \includegraphics[width=0.16\textwidth]{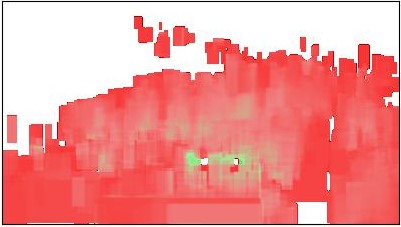}
      \includegraphics[width=0.16\textwidth]{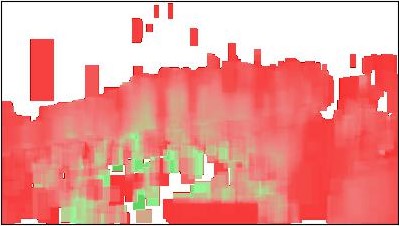} 
     \includegraphics[width=0.16\textwidth]{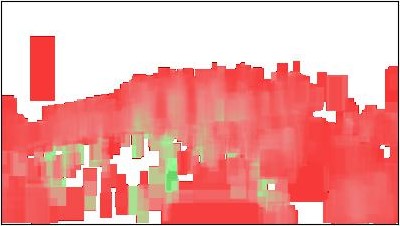}
      \includegraphics[width=0.16\textwidth]{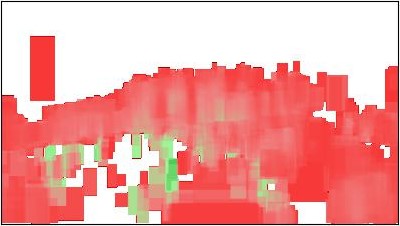}
      \subcaption{RTMDet.} 
      \label{fig:castro-results-rtm}
    \end{minipage} 
    \caption{The average confidence of pedestrian detection for each pixel in the observed area of Castro Street. The five videos, recorded in different lighting conditions (from left to right: sunrise, daytime, sunset, night, and night), are presented separately.}
\label{fig:castro-results-combined}
\end{figure*}

\subsubsection{Effect of Pedestrian Angle}
\label{sec:angle-analysis}

Here, we analyze the effect of a pedestrian's angle from a pedestrian detector's camera on the pedestrian detector's confidence.

\textbf{Experimental Setup:} We analyzed a person standing at eight different angles (0°, 45°, 90°, 135°, 180°, 225°, 270°, 315°) relative to a camera whose footage was fed to a Faster R-CNN object detector. These experiments were repeated with the person standing at three different distances (2.5m, 5m, 10m), and the camera positioned at three different heights (0.6m, 1.8m, 2.4m).

\textbf{Results:} The results of this analysis are presented in Table \ref{tab:standing-results-faster}. We found that when keeping other variables (distance, height) constant, the angle of the person resulted in varying confidence in the object detector, ranging from 0.55 to 1.0 in the most extreme case. In particular, we notice a significant confidence drop when the pedestrian is perpendicular to the camera (\textit{i.e.,} 90° and 270°).

\begin{insight} \label{insight:angle}
The confidence of an object detector is dependent on the \textit{angle} of the person relative to the object detector.
\end{insight}

\subsubsection{Effect of Pedestrian Detector Camera Height}
\label{sec:height-analysis}

Here, we analyze the effect of the camera's height on the pedestrian detector's confidence.

\textbf{Experimental Setup:} We analyzed a camera positioned at three different heights (0.6m, 1.8m, 2.4m), relative to a stationary person,whose footage was fed to a Faster R-CNN object detector. These experiments were repeated with the person standing at three distances (2.5m, 5m, 10m) and eight angles (0°, 45°, 90°, 135°, 180°, 225°, 270°, 315°).

\textbf{Results:} The results of this analysis are presented in Table \ref{tab:standing-results-faster}. We found that when keeping other variables (distance, angle) constant, the height of the camera resulted in varying confidence in the object detector, ranging from 0.57 to 1.0 in the most extreme case.

\begin{insight} \label{insight:height}
The confidence of an object detector is dependent on the \textit{height} of the camera recording the scene for the object detector.
\end{insight}

\subsection{Effect of Time of Day and Ambient Lighting}
\label{sec:ambient-light-analysis}

In this section, we analyze the effect of \textbf{time of day} and \textbf{ambient light} on the confidence of object detectors when detecting people at different angles, heights, and distances. In particular, we repeated our analysis from Section \ref{sec:position-analysis} at different times of day and with differing ambient light levels.

\textbf{Experimental Setup:} We repeated the experimental setup from 
Section \ref{sec:position-analysis}, varying the time of day and ambient light level. We repeated the experiment four times: (1) morning (approx. 11:00), (2) afternoon (approx. 16:00), and (3) night (approx. 20:00), and (4) lab conditions. The ambient light in the morning was the highest, with direct sunlight illuminating the scene, followed by the afternoon. The night had the lowest ambient light levels.

\textbf{Results:} Table \ref{tab:standing-results-faster} presents the results of this analysis. We can see that in the morning and the afternoon, the confidence ranges from 0.99 to 1.0. At night the confidence ranges from 0.55 to 1.0, and in lab conditions the confidence ranges from 0.57 to 1.0.

\begin{insight} \label{insight:ambient-light}
The phenomenon of varying model confidence depending on the angle, height, and distance of the person from the object detector is present in scenes lacking natural sunlight (indoors and night time).
\end{insight}

\subsubsection{Effect of Object Detector Model}
\label{sec:model-analysis}

Here, we analyze the impact of a person's position, as well as ambient light and time of day, on different object detector architectures.

\textbf{Experimental Setup:} We repeated the analyses from Sections \ref{sec:position-analysis} and \ref{sec:ambient-light-analysis} using object detectors with five architectures: (1) Faster R-CNN, (2) YOLOv3, (3) SSD, (4) DiffusionDet, and (5) RTMDet.

\textbf{Results:} The results of this analysis for Faster R-CNN, YOLOv3, SSD, and RTMDet are presented in Tables \ref{tab:standing-results-combined} and \ref{tab:standing-results-combined-2}. DiffusionDet is presented in Section \ref{sec:scene-analysis-appendix} in the appendix. We found that all five object detector architectures are affected by the pedestrian position and ambient light. 
In addition, we found that for three of the five models (YOLOv3, SSD, and RTMDet) the effect persists through all ambient light levels.

\begin{insight} \label{insight:architecture}
The effects of varying pedestrian position and ambient light impact the confidence levels of object detectors with a variety of architectures.
\end{insight}

\subsection{Location Effect on Pedestrian Detection Systems}
\label{sec:scene-analysis}

In our next analysis, we investigate the effect of a pedestrian’s location in an observed scene on an AI-based pedestrian detector’s confidence.

\subsubsection{Effect of Observed Location}
\label{sec:scene-analysis}

\textit{Experimental Setup:} We utilized publicly available stationary recordings of pedestrian traffic in a public area, recorded by an overlooking camera (example still frames displayed in Fig. \ref{fig:observed-scenes}). The footage was recorded in three public locations: (1) Shibuya Crossing in Tokyo, Japan (six recordings), (2) Broadway, New York (six recordings), and (3) Castro District, San Francisco, California (five recordings). Each recording contained approximately four hours of footage.

For each recording, we fed one frame from every two seconds of footage to a Faster R-CNN object detector.
For each pixel in the frame, we calculated the mean confidence of the object detector’s “person” detection bounding boxes that cover the pixel.

\textbf{Results:} The results of this analysis are presented in Figs. \ref{fig:shibuya-results-faster}, \ref{fig:new-york-results-faster}, and \ref{fig:castro-results-faster}. We found that in all three locations, the confidence of the object detector varied depending on the location of people in the frame. For instance, in the Shibuya Crossing footage, there are large areas of low confidence farther away from the camera, as well as closer to the camera, where a pole partially obscures passing pedestrians.

\begin{insight} \label{insight:location-detection}
The location of a person in a recorded scene impacts the confidence level of a pedestrian detector when detecting the person.
\end{insight}

\begin{insight} \label{insight:location-generality}
The effect of a person's location in a recorded scene on object detector confidence is present in various global locations.
\end{insight}

\subsubsection{Effect of Pedestrian Detector Model}
\label{sec:scene-analysis-object-detectors}

\textit{Experimental Setup:} We took the stationary recordings of pedestrian traffic from the previous analysis, and fed one frame from every two seconds of footage to five different object detectors: (1) Faster R-CNN, (2) YOLOv3, (3) SSD, (4) DiffusionDet, and (5) RTMDet.
For each object detector, and each pixel in the frame, we calculated the mean confidence of the object detector’s “person” detection bounding boxes that cover the pixel.

\textbf{Results:} The results of this analysis are presented in Figs. \ref{fig:shibuya-results-combined}, \ref{fig:new-york-results-combined}, and \ref{fig:castro-results-combined} for Faster R-CNN, YOLOv3, SSD, and RTMDet. Results for DiffusionDet are presented in Figs. \ref{fig:shibuya-results-dnet}, \ref{fig:new-york-results-dnet}, and \ref{fig:castro-results-dnet} in the appendix. We found that in all five object detectors, the confidence of the object detector varies depending on the location of people in the frame.

\begin{insight} \label{insight:location-generality-object-detectors}
The impact of a person's location in a recorded scene on object detector confidence is present when using various object detector architectures.
\end{insight}

\section{L-PET: Location-Based Privacy-Enhancing Technique}
\label{sec:evaluation-lpea}

In this section, we introduce and evaluate L-PET, our proposed location-based privacy-enhancing technique. We first review terminology and the problem definition, and then review the methodology and evaluations of L-PET.

\begin{figure*}[t]
    \centering
    \includegraphics[width=0.85\textwidth]{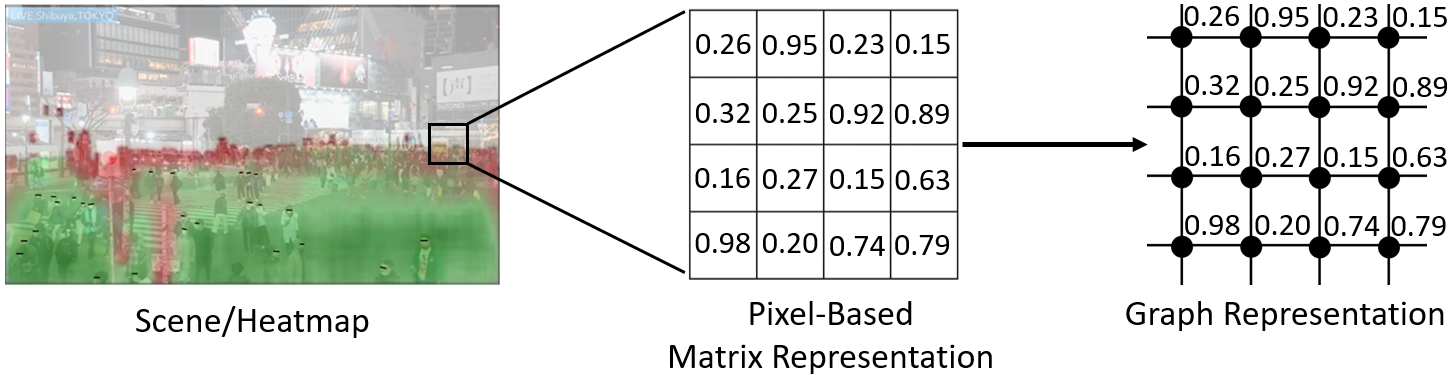}
    \caption{A visualization of the reduction of the scene/heatmap from a pixel-based matrix representation to a graph-based representation.}
    \label{fig:problem-reduction}
\end{figure*}

\textbf{Terminology:} We first define key terminology. A \textit{scene} is a physical world area observed by an object detector via camera stream/recording. 
A \textit{confidence heatmap} records for each pixel in the scene, the average confidence of the 'person' bounding boxes which fall on the pixel. 
Similar to a confidence heatmap, a \textit{detection heatmap} records for each pixel in the scene, the number of 'person' bounding boxes which fall on the pixel.

\textbf{Problem Definition:}

The problem of finding an optimal path between two points in the scene is reduced to finding an optimal path between two nodes in a graph representation. 
In order to avoid detection, the pedestrian would want to minimize: (1) the maximum confidence on the path, and (2) the average confidence of the nodes on the path.

\begin{algorithm}[t!]
\caption{Generate Confidence and Detection Heatmaps}
\begin{flushleft}
\textbf{Input:} video footage $V$, object detector $M$. \\
\textbf{Output:} $H_c$ - confidence heatmap of scene recorded in $V$, $H_d$ - detection heatmap.
\end{flushleft}
\begin{algorithmic}[1]
\State $H_c, H_d \gets \text{zero-value matrices, same resolution as } V$.
\State $F_V \gets \text{frames of } V$.
\State $BB \gets M(F_V) \text{, } M \text{'s 'person' bounding boxes for } F_V$.
\For{$bb \in BB$}
    \State $C_{bb} \gets M\text{'s confidence score for } bb$.
    \For{$(i,j) \in \text{pixels covered by } bb$}
    \State $H_c(i,j) \gets H_c(i,j) + C_{bb}$.
    \State $H_d(i,j) \gets H_d(i,j) + 1$.
    \EndFor
\EndFor
\State $H_c(i,j) \gets H_c(i,j) / H_d(i,j) \text{, } \forall (i,j) \text{ s.t. } H_d(i,j) \neq 0$.
\textbf{return:} $H_c, H_d$
\end{algorithmic}
\label{alg:generate-heatmaps}
\end{algorithm}

\begin{algorithm}[t!]
\caption{L-PET}
\begin{flushleft}
\textbf{Input:} $V, M, s, e$. \\
\textbf{Output:} $p$ - lowest-max-cost path between $s$ and $e$.
\end{flushleft}
\begin{algorithmic}[1]
\State $H_c, H_d \gets \text{GenerateHeatmaps}(V,M)$
\State $H_a \gets H_c[(i,j) \in H_c | H_d(i,j) > 0]$.
\State $Q \gets \text{priority queue.}$
\For{$p \in H_a$}
    \State $dist(p) \gets \infty$.
    \State $prev(p) \gets \text{UNDEFINED}$.
    \State $Q.add(p), \text{priority defined by $dist(p)$.}$
\EndFor
\State $dist(s) \gets 0$
\While{$Q \text{ is not empty:}$}
    \State $u \gets \text{node in $Q$ with min $dist(u)$.}$
    \State $Q.remove(u)$
    \For{$\{v \in Q | v \text{ is a neighbor of } u\}:$}
        \State $alt \gets min(max(dist(u), H_c(v)), dist(v))$.
        \If{$alt < dist(v)$}
        \State $dist(v) \gets alt$.
        \State $prev(v) \gets u$.
        \EndIf
    \EndFor
\EndWhile
\textbf{return:} $dist, prev$
\end{algorithmic}
\label{alg:lpea}
\end{algorithm}

\subsection{L-PET Methodology}
\label{sec:lpea-methodology}

In this section, we present the methodology of the L-PET privacy-enhancing technique.
L-PET receives four inputs: (1) $V$ - video footage of the scene, (2) $M$ - the object detection model analyzing the scene, (3) $s$ - a starting point for the path, and (4) $e$ - an end point for the path.

The first step of L-PET is to generate the confidence and detection heatmaps using $V$ and $M$.
We utilize 4 hour recordings of the scene for creating the confidence and detection heatmaps.
We feed one frame from every two seconds of footage to an object detector.
For each 4 hour recording, we calculate a confidence heatmap by calculating, for each pixel in the frame, the mean confidence of the object detector’s “person” detection bounding boxes that cover the pixel.
Similarly, we calculate a detection heatmap by calculating, for each pixel in the frame, the number of “person” detection bounding boxes that cover the pixel.
Algorithm \ref{alg:generate-heatmaps} presents the methodology used for generating the confidence and detection heatmaps.

The next step is to reduce the heatmap to a graph-based format. 
We perform a reduction from a pixel-based matrix representation to a graph-based representation. 
Each pixel is represented by a node in the graph-based representation, with two-way edges connecting each node to its cardinal and diagonal neighbors. 
Figure \ref{fig:problem-reduction} presents a demonstration of this reduction. 
Each node in the graph representation is defined with an associated weight equal to the confidence level recorded in the confidence heatmap’s respective pixel.

The path-finding step is a \textbf{modified version} of Dijkstra \cite{dijkstra1959note}, where an optimal path between two nodes is the path that crosses the lowest possible maximum confidence node.
Each path has at least one node possessing the highest confidence on the path; L-PET finds the path with such node possessing the lowest possible value, \textit{i.e., every other path contains a node with either equal or greater confidence}. 
To find this path, Dijkstra is performed according to a modified weight node update function.
Line 13 in Algorithm \ref{alg:lpea} records for each point in the scene, the minimum max confidence that is crossed in order to get to that point.
This contrasts with Dijkstra which minimizes the total path cost to a given point.
This calculation, at the end of running L-PET, finds the highest confidence value that must be crossed in order to get to each point.
The output of our proposed method is $p$, a path between the starting point and the end point of the path. 
Following this path, a pedestrian will achieve either: 1) minimal detection, or 2) no detection from the pedestrian detection system. 
Algorithm \ref{alg:lpea} presents the methodology of L-PET.

The resulting path can then be determined by starting from $e$, and following $prev(e), prev(prev(e)), prev(prev(...(prev(e)))) = s$. If there is any point $n$ on the path such that $prev(n)$ is undefined, L-PET returns no solution.

\begin{table*}[th!]
\centering
\caption{Average maximum confidence and average per-step confidence of 100 paths generated between 10 random start points and 10 random end points on the observed scene at different times of day. 100 paths were generated for each group: (1) paths generated with L-PET, (2) random direct paths of Manhattan distance from start to end, and (3) random paths. The paths were generated on a scene observed by an object detector.}
\label{tab:lpea-results-combined}
\resizebox{2.0\columnwidth}{!}{%
\begin{minipage}{0.6\textwidth}
\centering
\subcaption{Faster R-CNN}
\label{tab:lpea-results-faster}
\begin{tabular}{c|c|ccc|ccc|}
\cline{2-8}
                                                         &                                   & \multicolumn{3}{c|}{Max Confidence}                                                                                                           & \multicolumn{3}{c|}{Average Confidence}                                                                                                       \\ \cline{3-8} 
\multirow{-2}{*}{}                                       & \multirow{-2}{*}{Video}           & \multicolumn{1}{c|}{L-PET}                                & \multicolumn{1}{c|}{Man. Dist.}                   & Random                       & \multicolumn{1}{c|}{L-PET}                                & \multicolumn{1}{c|}{Man. Dist.}                   & Random                       \\ \hline
\multicolumn{1}{|c|}{}                                   & \cellcolor[HTML]{C0C0C0}Daytime 1 & \multicolumn{1}{c|}{\cellcolor[HTML]{C0C0C0}\textbf{0.73}} & \multicolumn{1}{c|}{\cellcolor[HTML]{C0C0C0}0.75} & \cellcolor[HTML]{C0C0C0}0.77 & \multicolumn{1}{c|}{\cellcolor[HTML]{C0C0C0}\textbf{0.54}} & \multicolumn{1}{c|}{\cellcolor[HTML]{C0C0C0}0.58} & \cellcolor[HTML]{C0C0C0}0.68 \\ \cline{2-8} 
\multicolumn{1}{|c|}{}                                   & \cellcolor[HTML]{EFEFEF}Daytime 2 & \multicolumn{1}{c|}{\cellcolor[HTML]{EFEFEF}\textbf{0.89}} & \multicolumn{1}{c|}{\cellcolor[HTML]{EFEFEF}0.90} & \cellcolor[HTML]{EFEFEF}0.91 & \multicolumn{1}{c|}{\cellcolor[HTML]{EFEFEF}\textbf{0.64}} & \multicolumn{1}{c|}{\cellcolor[HTML]{EFEFEF}0.70} & \cellcolor[HTML]{EFEFEF}0.82 \\ \cline{2-8} 
\multicolumn{1}{|c|}{}                                   & \cellcolor[HTML]{C0C0C0}Afternoon & \multicolumn{1}{c|}{\cellcolor[HTML]{C0C0C0}\textbf{0.62}} & \multicolumn{1}{c|}{\cellcolor[HTML]{C0C0C0}0.69} & \cellcolor[HTML]{C0C0C0}0.67 & \multicolumn{1}{c|}{\cellcolor[HTML]{C0C0C0}\textbf{0.49}} & \multicolumn{1}{c|}{\cellcolor[HTML]{C0C0C0}0.54} & \cellcolor[HTML]{C0C0C0}0.59 \\ \cline{2-8} 
\multicolumn{1}{|c|}{}                                   & \cellcolor[HTML]{EFEFEF}Evening   & \multicolumn{1}{c|}{\cellcolor[HTML]{EFEFEF}\textbf{0.64}} & \multicolumn{1}{c|}{\cellcolor[HTML]{EFEFEF}0.69} & \cellcolor[HTML]{EFEFEF}0.67 & \multicolumn{1}{c|}{\cellcolor[HTML]{EFEFEF}\textbf{0.48}} & \multicolumn{1}{c|}{\cellcolor[HTML]{EFEFEF}0.54} & \cellcolor[HTML]{EFEFEF}0.58 \\ \cline{2-8} 
\multicolumn{1}{|c|}{}                                   & \cellcolor[HTML]{C0C0C0}Night 1   & \multicolumn{1}{c|}{\cellcolor[HTML]{C0C0C0}\textbf{0.77}} & \multicolumn{1}{c|}{\cellcolor[HTML]{C0C0C0}0.78} & \cellcolor[HTML]{C0C0C0}0.80 & \multicolumn{1}{c|}{\cellcolor[HTML]{C0C0C0}\textbf{0.54}} & \multicolumn{1}{c|}{\cellcolor[HTML]{C0C0C0}0.61} & \cellcolor[HTML]{C0C0C0}0.71 \\ \cline{2-8} 
\multicolumn{1}{|c|}{\multirow{-6}{*}{\rotatebox[origin=c]{90}{Shibuya}}} & \cellcolor[HTML]{EFEFEF}Night 2   & \multicolumn{1}{c|}{\cellcolor[HTML]{EFEFEF}\textbf{0.69}} & \multicolumn{1}{c|}{\cellcolor[HTML]{EFEFEF}0.70} & \cellcolor[HTML]{EFEFEF}0.71 & \multicolumn{1}{c|}{\cellcolor[HTML]{EFEFEF}\textbf{0.49}} & \multicolumn{1}{c|}{\cellcolor[HTML]{EFEFEF}0.55} & \cellcolor[HTML]{EFEFEF}0.64 \\ \hline
\multicolumn{1}{|c|}{}                                   & \cellcolor[HTML]{C0C0C0}Daytime 1 & \multicolumn{1}{c|}{\cellcolor[HTML]{C0C0C0}\textbf{0.75}} & \multicolumn{1}{c|}{\cellcolor[HTML]{C0C0C0}0.86} & \cellcolor[HTML]{C0C0C0}0.84 & \multicolumn{1}{c|}{\cellcolor[HTML]{C0C0C0}\textbf{0.51}} & \multicolumn{1}{c|}{\cellcolor[HTML]{C0C0C0}0.62} & \cellcolor[HTML]{C0C0C0}0.70 \\ \cline{2-8} 
\multicolumn{1}{|c|}{}                                   & \cellcolor[HTML]{EFEFEF}Daytime 2 & \multicolumn{1}{c|}{\cellcolor[HTML]{EFEFEF}\textbf{0.64}} & \multicolumn{1}{c|}{\cellcolor[HTML]{EFEFEF}0.71} & \cellcolor[HTML]{EFEFEF}0.73 & \multicolumn{1}{c|}{\cellcolor[HTML]{EFEFEF}\textbf{0.46}} & \multicolumn{1}{c|}{\cellcolor[HTML]{EFEFEF}0.54} & \cellcolor[HTML]{EFEFEF}0.59 \\ \cline{2-8} 
\multicolumn{1}{|c|}{}                                   & \cellcolor[HTML]{C0C0C0}Daytime 3 & \multicolumn{1}{c|}{\cellcolor[HTML]{C0C0C0}\textbf{0.78}} & \multicolumn{1}{c|}{\cellcolor[HTML]{C0C0C0}0.83} & \cellcolor[HTML]{C0C0C0}0.85 & \multicolumn{1}{c|}{\cellcolor[HTML]{C0C0C0}\textbf{0.57}} & \multicolumn{1}{c|}{\cellcolor[HTML]{C0C0C0}0.67} & \cellcolor[HTML]{C0C0C0}0.69 \\ \cline{2-8} 
\multicolumn{1}{|c|}{}                                   & \cellcolor[HTML]{EFEFEF}Night 1   & \multicolumn{1}{c|}{\cellcolor[HTML]{EFEFEF}\textbf{0.69}} & \multicolumn{1}{c|}{\cellcolor[HTML]{EFEFEF}0.75} & \cellcolor[HTML]{EFEFEF}0.75 & \multicolumn{1}{c|}{\cellcolor[HTML]{EFEFEF}\textbf{0.45}} & \multicolumn{1}{c|}{\cellcolor[HTML]{EFEFEF}0.57} & \cellcolor[HTML]{EFEFEF}0.62 \\ \cline{2-8} 
\multicolumn{1}{|c|}{}                                   & \cellcolor[HTML]{C0C0C0}Night 2   & \multicolumn{1}{c|}{\cellcolor[HTML]{C0C0C0}\textbf{0.59}} & \multicolumn{1}{c|}{\cellcolor[HTML]{C0C0C0}0.66} & \cellcolor[HTML]{C0C0C0}0.68 & \multicolumn{1}{c|}{\cellcolor[HTML]{C0C0C0}\textbf{0.41}} & \multicolumn{1}{c|}{\cellcolor[HTML]{C0C0C0}0.49} & \cellcolor[HTML]{C0C0C0}0.58 \\ \cline{2-8} 
\multicolumn{1}{|c|}{\multirow{-6}{*}{\rotatebox[origin=c]{90}{Broadway}}}         & \cellcolor[HTML]{EFEFEF}Night 3   & \multicolumn{1}{c|}{\cellcolor[HTML]{EFEFEF}\textbf{0.62}} & \multicolumn{1}{c|}{\cellcolor[HTML]{EFEFEF}0.73} & \cellcolor[HTML]{EFEFEF}0.73 & \multicolumn{1}{c|}{\cellcolor[HTML]{EFEFEF}\textbf{0.48}} & \multicolumn{1}{c|}{\cellcolor[HTML]{EFEFEF}0.55} & \cellcolor[HTML]{EFEFEF}0.58 \\ \hline
\multicolumn{1}{|c|}{}                                   & \cellcolor[HTML]{C0C0C0}Sunrise   & \multicolumn{1}{c|}{\cellcolor[HTML]{C0C0C0}\textbf{0.77}} & \multicolumn{1}{c|}{\cellcolor[HTML]{C0C0C0}0.92} & \cellcolor[HTML]{C0C0C0}0.89 & \multicolumn{1}{c|}{\cellcolor[HTML]{C0C0C0}\textbf{0.57}} & \multicolumn{1}{c|}{\cellcolor[HTML]{C0C0C0}0.68} & \cellcolor[HTML]{C0C0C0}0.68 \\ \cline{2-8} 
\multicolumn{1}{|c|}{}                                   & \cellcolor[HTML]{EFEFEF}Daytime   & \multicolumn{1}{c|}{\cellcolor[HTML]{EFEFEF}\textbf{0.70}} & \multicolumn{1}{c|}{\cellcolor[HTML]{EFEFEF}0.90} & \cellcolor[HTML]{EFEFEF}0.85 & \multicolumn{1}{c|}{\cellcolor[HTML]{EFEFEF}\textbf{0.54}} & \multicolumn{1}{c|}{\cellcolor[HTML]{EFEFEF}0.61} & \cellcolor[HTML]{EFEFEF}0.62 \\ \cline{2-8} 
\multicolumn{1}{|c|}{}                                   & \cellcolor[HTML]{C0C0C0}Sunset    & \multicolumn{1}{c|}{\cellcolor[HTML]{C0C0C0}\textbf{0.73}} & \multicolumn{1}{c|}{\cellcolor[HTML]{C0C0C0}0.87} & \cellcolor[HTML]{C0C0C0}0.87 & \multicolumn{1}{c|}{\cellcolor[HTML]{C0C0C0}\textbf{0.54}} & \multicolumn{1}{c|}{\cellcolor[HTML]{C0C0C0}0.58} & \cellcolor[HTML]{C0C0C0}0.62 \\ \cline{2-8} 
\multicolumn{1}{|c|}{}                                   & \cellcolor[HTML]{EFEFEF}Night 1   & \multicolumn{1}{c|}{\cellcolor[HTML]{EFEFEF}\textbf{0.75}} & \multicolumn{1}{c|}{\cellcolor[HTML]{EFEFEF}0.90} & \cellcolor[HTML]{EFEFEF}0.91 & \multicolumn{1}{c|}{\cellcolor[HTML]{EFEFEF}\textbf{0.54}} & \multicolumn{1}{c|}{\cellcolor[HTML]{EFEFEF}0.62} & \cellcolor[HTML]{EFEFEF}0.64 \\ \cline{2-8} 
\multicolumn{1}{|c|}{\multirow{-5}{*}{\rotatebox[origin=c]{90}{Castro}}}    & \cellcolor[HTML]{C0C0C0}Night 2   & \multicolumn{1}{c|}{\cellcolor[HTML]{C0C0C0}\textbf{0.78}} & \multicolumn{1}{c|}{\cellcolor[HTML]{C0C0C0}0.93} & \cellcolor[HTML]{C0C0C0}0.93 & \multicolumn{1}{c|}{\cellcolor[HTML]{C0C0C0}\textbf{0.54}} & \multicolumn{1}{c|}{\cellcolor[HTML]{C0C0C0}0.64} & \cellcolor[HTML]{C0C0C0}0.64 \\ \hline
\multicolumn{1}{l|}{}                                    & Average                           & \multicolumn{1}{c|}{\textbf{0.71}}                                  & \multicolumn{1}{c|}{0.80}                         & 0.80                         & \multicolumn{1}{c|}{\textbf{0.52}}                                  & \multicolumn{1}{c|}{0.59}                         &  0.65                       \\ \cline{2-8}
\end{tabular}%
\end{minipage}
\hspace{5pt}
\begin{minipage}{0.6\textwidth}
\centering
\subcaption{YOLOv3}
\label{tab:lpea-results-yolo}
\begin{tabular}{c|c|ccc|ccc|}
\cline{2-8}
                                                         &                                   & \multicolumn{3}{c|}{Max Confidence}                                                                                                           & \multicolumn{3}{c|}{Average Confidence}                                                                                                       \\ \cline{3-8} 
\multirow{-2}{*}{}                                       & \multirow{-2}{*}{Video}           & \multicolumn{1}{c|}{L-PET}                                & \multicolumn{1}{c|}{Man. Dist.}                   & Random                       & \multicolumn{1}{c|}{L-PET}                                & \multicolumn{1}{c|}{Man. Dist.}                   & Random                       \\ \hline
\multicolumn{1}{|c|}{}                                   & \cellcolor[HTML]{C0C0C0}Daytime 1 & \multicolumn{1}{c|}{\cellcolor[HTML]{C0C0C0}\textbf{0.54}} & \multicolumn{1}{c|}{\cellcolor[HTML]{C0C0C0}0.55} & \cellcolor[HTML]{C0C0C0}0.56 & \multicolumn{1}{c|}{\cellcolor[HTML]{C0C0C0}\textbf{0.39}} & \multicolumn{1}{c|}{\cellcolor[HTML]{C0C0C0}0.42} & \cellcolor[HTML]{C0C0C0}0.48 \\ \cline{2-8} 
\multicolumn{1}{|c|}{}                                   & \cellcolor[HTML]{EFEFEF}Daytime 2 & \multicolumn{1}{c|}{\cellcolor[HTML]{EFEFEF}\textbf{0.58}} & \multicolumn{1}{c|}{\cellcolor[HTML]{EFEFEF}0.61} & \cellcolor[HTML]{EFEFEF}0.60 & \multicolumn{1}{c|}{\cellcolor[HTML]{EFEFEF}\textbf{0.44}} & \multicolumn{1}{c|}{\cellcolor[HTML]{EFEFEF}0.47} & \cellcolor[HTML]{EFEFEF}0.51 \\ \cline{2-8} 
\multicolumn{1}{|c|}{}                                   & \cellcolor[HTML]{C0C0C0}Afternoon & \multicolumn{1}{c|}{\cellcolor[HTML]{C0C0C0}\textbf{0.48}} & \multicolumn{1}{c|}{\cellcolor[HTML]{C0C0C0}0.57} & \cellcolor[HTML]{C0C0C0}0.51 & \multicolumn{1}{c|}{\cellcolor[HTML]{C0C0C0}\textbf{0.37}} & \multicolumn{1}{c|}{\cellcolor[HTML]{C0C0C0}0.41} & \cellcolor[HTML]{C0C0C0}0.45 \\ \cline{2-8} 
\multicolumn{1}{|c|}{}                                   & \cellcolor[HTML]{EFEFEF}Evening   & \multicolumn{1}{c|}{\cellcolor[HTML]{EFEFEF}\textbf{0.47}} & \multicolumn{1}{c|}{\cellcolor[HTML]{EFEFEF}0.60} & \cellcolor[HTML]{EFEFEF}0.51 & \multicolumn{1}{c|}{\cellcolor[HTML]{EFEFEF}\textbf{0.38}} & \multicolumn{1}{c|}{\cellcolor[HTML]{EFEFEF}0.41} & \cellcolor[HTML]{EFEFEF}0.44 \\ \cline{2-8} 
\multicolumn{1}{|c|}{}                                   & \cellcolor[HTML]{C0C0C0}Night 1   & \multicolumn{1}{c|}{\cellcolor[HTML]{C0C0C0}\textbf{0.55}} & \multicolumn{1}{c|}{\cellcolor[HTML]{C0C0C0}0.59} & \cellcolor[HTML]{C0C0C0}0.57 & \multicolumn{1}{c|}{\cellcolor[HTML]{C0C0C0}\textbf{0.41}} & \multicolumn{1}{c|}{\cellcolor[HTML]{C0C0C0}0.45} & \cellcolor[HTML]{C0C0C0}0.49 \\ \cline{2-8} 
\multicolumn{1}{|c|}{\multirow{-6}{*}{\rotatebox[origin=c]{90}{Shibuya}}} & \cellcolor[HTML]{EFEFEF}Night 2   & \multicolumn{1}{c|}{\cellcolor[HTML]{EFEFEF}\textbf{0.51}} & \multicolumn{1}{c|}{\cellcolor[HTML]{EFEFEF}0.55} & \cellcolor[HTML]{EFEFEF}0.55 & \multicolumn{1}{c|}{\cellcolor[HTML]{EFEFEF}\textbf{0.38}} & \multicolumn{1}{c|}{\cellcolor[HTML]{EFEFEF}0.43} & \cellcolor[HTML]{EFEFEF}0.49 \\ \hline
\multicolumn{1}{|c|}{}                                   & \cellcolor[HTML]{C0C0C0}Daytime 1 & \multicolumn{1}{c|}{\cellcolor[HTML]{C0C0C0}\textbf{0.44}} & \multicolumn{1}{c|}{\cellcolor[HTML]{C0C0C0}0.51} & \cellcolor[HTML]{C0C0C0}0.54 & \multicolumn{1}{c|}{\cellcolor[HTML]{C0C0C0}\textbf{0.34}} & \multicolumn{1}{c|}{\cellcolor[HTML]{C0C0C0}0.39} & \cellcolor[HTML]{C0C0C0}0.42 \\ \cline{2-8} 
\multicolumn{1}{|c|}{}                                   & \cellcolor[HTML]{EFEFEF}Daytime 2 & \multicolumn{1}{c|}{\cellcolor[HTML]{EFEFEF}\textbf{0.46}} & \multicolumn{1}{c|}{\cellcolor[HTML]{EFEFEF}0.53} & \cellcolor[HTML]{EFEFEF}0.54 & \multicolumn{1}{c|}{\cellcolor[HTML]{EFEFEF}\textbf{0.34}} & \multicolumn{1}{c|}{\cellcolor[HTML]{EFEFEF}0.39} & \cellcolor[HTML]{EFEFEF}0.43 \\ \cline{2-8} 
\multicolumn{1}{|c|}{}                                   & \cellcolor[HTML]{C0C0C0}Daytime 3 & \multicolumn{1}{c|}{\cellcolor[HTML]{C0C0C0}\textbf{0.47}} & \multicolumn{1}{c|}{\cellcolor[HTML]{C0C0C0}0.55} & \cellcolor[HTML]{C0C0C0}0.58 & \multicolumn{1}{c|}{\cellcolor[HTML]{C0C0C0}\textbf{0.37}} & \multicolumn{1}{c|}{\cellcolor[HTML]{C0C0C0}0.42} & \cellcolor[HTML]{C0C0C0}0.45 \\ \cline{2-8} 
\multicolumn{1}{|c|}{}                                   & \cellcolor[HTML]{EFEFEF}Night 1   & \multicolumn{1}{c|}{\cellcolor[HTML]{EFEFEF}\textbf{0.45}} & \multicolumn{1}{c|}{\cellcolor[HTML]{EFEFEF}0.55} & \cellcolor[HTML]{EFEFEF}0.55 & \multicolumn{1}{c|}{\cellcolor[HTML]{EFEFEF}\textbf{0.35}} & \multicolumn{1}{c|}{\cellcolor[HTML]{EFEFEF}0.41} & \cellcolor[HTML]{EFEFEF}0.43 \\ \cline{2-8} 
\multicolumn{1}{|c|}{}                                   & \cellcolor[HTML]{C0C0C0}Night 2   & \multicolumn{1}{c|}{\cellcolor[HTML]{C0C0C0}\textbf{0.43}} & \multicolumn{1}{c|}{\cellcolor[HTML]{C0C0C0}0.49} & \cellcolor[HTML]{C0C0C0}0.51 & \multicolumn{1}{c|}{\cellcolor[HTML]{C0C0C0}\textbf{0.33}} & \multicolumn{1}{c|}{\cellcolor[HTML]{C0C0C0}0.38} & \cellcolor[HTML]{C0C0C0}0.42 \\ \cline{2-8} 
\multicolumn{1}{|c|}{\multirow{-6}{*}{\rotatebox[origin=c]{90}{Broadway}}}         & \cellcolor[HTML]{EFEFEF}Night 3   & \multicolumn{1}{c|}{\cellcolor[HTML]{EFEFEF}\textbf{0.41}} & \multicolumn{1}{c|}{\cellcolor[HTML]{EFEFEF}0.52} & \cellcolor[HTML]{EFEFEF}0.53 & \multicolumn{1}{c|}{\cellcolor[HTML]{EFEFEF}\textbf{0.34}} & \multicolumn{1}{c|}{\cellcolor[HTML]{EFEFEF}0.39} & \cellcolor[HTML]{EFEFEF}0.40 \\ \hline
\multicolumn{1}{|c|}{}                                   & \cellcolor[HTML]{C0C0C0}Sunrise   & \multicolumn{1}{c|}{\cellcolor[HTML]{C0C0C0}\textbf{0.56}} & \multicolumn{1}{c|}{\cellcolor[HTML]{C0C0C0}0.68} & \cellcolor[HTML]{C0C0C0}0.68 & \multicolumn{1}{c|}{\cellcolor[HTML]{C0C0C0}\textbf{0.38}} & \multicolumn{1}{c|}{\cellcolor[HTML]{C0C0C0}0.43} & \cellcolor[HTML]{C0C0C0}0.46 \\ \cline{2-8} 
\multicolumn{1}{|c|}{}                                   & \cellcolor[HTML]{EFEFEF}Daytime   & \multicolumn{1}{c|}{\cellcolor[HTML]{EFEFEF}\textbf{0.52}} & \multicolumn{1}{c|}{\cellcolor[HTML]{EFEFEF}0.65} & \cellcolor[HTML]{EFEFEF}0.66 & \multicolumn{1}{c|}{\cellcolor[HTML]{EFEFEF}\textbf{0.38}} & \multicolumn{1}{c|}{\cellcolor[HTML]{EFEFEF}0.41} & \cellcolor[HTML]{EFEFEF}0.45 \\ \cline{2-8} 
\multicolumn{1}{|c|}{}                                   & \cellcolor[HTML]{C0C0C0}Sunset    & \multicolumn{1}{c|}{\cellcolor[HTML]{C0C0C0}\textbf{0.53}} & \multicolumn{1}{c|}{\cellcolor[HTML]{C0C0C0}0.70} & \cellcolor[HTML]{C0C0C0}0.71 & \multicolumn{1}{c|}{\cellcolor[HTML]{C0C0C0}\textbf{0.39}} & \multicolumn{1}{c|}{\cellcolor[HTML]{C0C0C0}0.44} & \cellcolor[HTML]{C0C0C0}0.46 \\ \cline{2-8} 
\multicolumn{1}{|c|}{}                                   & \cellcolor[HTML]{EFEFEF}Night 1   & \multicolumn{1}{c|}{\cellcolor[HTML]{EFEFEF}\textbf{0.50}} & \multicolumn{1}{c|}{\cellcolor[HTML]{EFEFEF}0.59} & \cellcolor[HTML]{EFEFEF}0.60 & \multicolumn{1}{c|}{\cellcolor[HTML]{EFEFEF}\textbf{0.38}} & \multicolumn{1}{c|}{\cellcolor[HTML]{EFEFEF}0.42} & \cellcolor[HTML]{EFEFEF}0.44 \\ \cline{2-8} 
\multicolumn{1}{|c|}{\multirow{-5}{*}{\rotatebox[origin=c]{90}{Castro}}}    & \cellcolor[HTML]{C0C0C0}Night 2   & \multicolumn{1}{c|}{\cellcolor[HTML]{C0C0C0}\textbf{0.56}} & \multicolumn{1}{c|}{\cellcolor[HTML]{C0C0C0}0.62} & \cellcolor[HTML]{C0C0C0}0.65 & \multicolumn{1}{c|}{\cellcolor[HTML]{C0C0C0}\textbf{0.41}} & \multicolumn{1}{c|}{\cellcolor[HTML]{C0C0C0}0.44} & \cellcolor[HTML]{C0C0C0}0.48 \\ \hline
\multicolumn{1}{l|}{}                                    & Average                           & \multicolumn{1}{c|}{\textbf{0.50}}                                  & \multicolumn{1}{c|}{0.58}                         & 0.58                         & \multicolumn{1}{c|}{\textbf{0.38}}                                  & \multicolumn{1}{c|}{0.42}                         & 0.45                         \\ \cline{2-8}
\end{tabular}%
\end{minipage}
}
\newline
\vspace*{0.15 cm}
\newline
\resizebox{2.0\columnwidth}{!}{%
\noindent
\begin{minipage}{0.6\textwidth}
\centering
\subcaption{SSD}
\label{tab:lpea-results-ssd}
\begin{tabular}{c|c|ccc|ccc|}
\cline{2-8}
                                                         &                                   & \multicolumn{3}{c|}{Max Confidence}                                                                                                           & \multicolumn{3}{c|}{Average Confidence}                                                                                                       \\ \cline{3-8} 
\multirow{-2}{*}{}                                       & \multirow{-2}{*}{Video}           & \multicolumn{1}{c|}{L-PET}                                & \multicolumn{1}{c|}{Man. Dist.}                   & Random                       & \multicolumn{1}{c|}{L-PET}                                & \multicolumn{1}{c|}{Man. Dist.}                   & Random                       \\ \hline
\multicolumn{1}{|c|}{}                                   & \cellcolor[HTML]{C0C0C0}Daytime 1 & \multicolumn{1}{c|}{\cellcolor[HTML]{C0C0C0}\textbf{0.43}} & \multicolumn{1}{c|}{\cellcolor[HTML]{C0C0C0}0.44} & \cellcolor[HTML]{C0C0C0}0.44 & \multicolumn{1}{c|}{\cellcolor[HTML]{C0C0C0}\textbf{0.38}} & \multicolumn{1}{c|}{\cellcolor[HTML]{C0C0C0}0.39} & \cellcolor[HTML]{C0C0C0}0.41 \\ \cline{2-8} 
\multicolumn{1}{|c|}{}                                   & \cellcolor[HTML]{EFEFEF}Daytime 2 & \multicolumn{1}{c|}{\cellcolor[HTML]{EFEFEF}\textbf{0.45}} & \multicolumn{1}{c|}{\cellcolor[HTML]{EFEFEF}0.51} & \cellcolor[HTML]{EFEFEF}0.48 & \multicolumn{1}{c|}{\cellcolor[HTML]{EFEFEF}\textbf{0.4}}  & \multicolumn{1}{c|}{\cellcolor[HTML]{EFEFEF}0.42} & \cellcolor[HTML]{EFEFEF}0.44 \\ \cline{2-8} 
\multicolumn{1}{|c|}{}                                   & \cellcolor[HTML]{C0C0C0}Afternoon & \multicolumn{1}{c|}{\cellcolor[HTML]{C0C0C0}\textbf{0.39}} & \multicolumn{1}{c|}{\cellcolor[HTML]{C0C0C0}0.46} & \cellcolor[HTML]{C0C0C0}0.40 & \multicolumn{1}{c|}{\cellcolor[HTML]{C0C0C0}\textbf{0.37}} & \multicolumn{1}{c|}{\cellcolor[HTML]{C0C0C0}0.38} & \cellcolor[HTML]{C0C0C0}0.38 \\ \cline{2-8} 
\multicolumn{1}{|c|}{}                                   & \cellcolor[HTML]{EFEFEF}Evening   & \multicolumn{1}{c|}{\cellcolor[HTML]{EFEFEF}\textbf{0.37}} & \multicolumn{1}{c|}{\cellcolor[HTML]{EFEFEF}0.47} & \cellcolor[HTML]{EFEFEF}0.39 & \multicolumn{1}{c|}{\cellcolor[HTML]{EFEFEF}\textbf{0.33}} & \multicolumn{1}{c|}{\cellcolor[HTML]{EFEFEF}0.34} & \cellcolor[HTML]{EFEFEF}0.36 \\ \cline{2-8} 
\multicolumn{1}{|c|}{}                                   & \cellcolor[HTML]{C0C0C0}Night 1   & \multicolumn{1}{c|}{\cellcolor[HTML]{C0C0C0}\textbf{0.41}} & \multicolumn{1}{c|}{\cellcolor[HTML]{C0C0C0}0.44} & \cellcolor[HTML]{C0C0C0}0.43 & \multicolumn{1}{c|}{\cellcolor[HTML]{C0C0C0}\textbf{0.36}} & \multicolumn{1}{c|}{\cellcolor[HTML]{C0C0C0}0.37} & \cellcolor[HTML]{C0C0C0}0.39 \\ \cline{2-8} 
\multicolumn{1}{|c|}{\multirow{-6}{*}{\rotatebox[origin=c]{90}{Shibuya}}} & \cellcolor[HTML]{EFEFEF}Night 2   & \multicolumn{1}{c|}{\cellcolor[HTML]{EFEFEF}\textbf{0.40}} & \multicolumn{1}{c|}{\cellcolor[HTML]{EFEFEF}0.43} & \cellcolor[HTML]{EFEFEF}0.41 & \multicolumn{1}{c|}{\cellcolor[HTML]{EFEFEF}\textbf{0.35}} & \multicolumn{1}{c|}{\cellcolor[HTML]{EFEFEF}0.37} & \cellcolor[HTML]{EFEFEF}0.38 \\ \hline
\multicolumn{1}{|c|}{}                                   & \cellcolor[HTML]{C0C0C0}Daytime 1 & \multicolumn{1}{c|}{\cellcolor[HTML]{C0C0C0}\textbf{0.40}} & \multicolumn{1}{c|}{\cellcolor[HTML]{C0C0C0}0.51} & \cellcolor[HTML]{C0C0C0}0.47 & \multicolumn{1}{c|}{\cellcolor[HTML]{C0C0C0}\textbf{0.33}} & \multicolumn{1}{c|}{\cellcolor[HTML]{C0C0C0}0.35} & \cellcolor[HTML]{C0C0C0}0.36 \\ \cline{2-8} 
\multicolumn{1}{|c|}{}                                   & \cellcolor[HTML]{EFEFEF}Daytime 2 & \multicolumn{1}{c|}{\cellcolor[HTML]{EFEFEF}\textbf{0.37}} & \multicolumn{1}{c|}{\cellcolor[HTML]{EFEFEF}0.46} & \cellcolor[HTML]{EFEFEF}0.45 & \multicolumn{1}{c|}{\cellcolor[HTML]{EFEFEF}\textbf{0.31}} & \multicolumn{1}{c|}{\cellcolor[HTML]{EFEFEF}0.34} & \cellcolor[HTML]{EFEFEF}0.36 \\ \cline{2-8} 
\multicolumn{1}{|c|}{}                                   & \cellcolor[HTML]{C0C0C0}Daytime 3 & \multicolumn{1}{c|}{\cellcolor[HTML]{C0C0C0}\textbf{0.43}} & \multicolumn{1}{c|}{\cellcolor[HTML]{C0C0C0}0.51} & \cellcolor[HTML]{C0C0C0}0.53 & \multicolumn{1}{c|}{\cellcolor[HTML]{C0C0C0}\textbf{0.33}} & \multicolumn{1}{c|}{\cellcolor[HTML]{C0C0C0}0.37} & \cellcolor[HTML]{C0C0C0}0.40 \\ \cline{2-8} 
\multicolumn{1}{|c|}{}                                   & \cellcolor[HTML]{EFEFEF}Night 1   & \multicolumn{1}{c|}{\cellcolor[HTML]{EFEFEF}\textbf{0.38}} & \multicolumn{1}{c|}{\cellcolor[HTML]{EFEFEF}0.49} & \cellcolor[HTML]{EFEFEF}0.47 & \multicolumn{1}{c|}{\cellcolor[HTML]{EFEFEF}\textbf{0.32}} & \multicolumn{1}{c|}{\cellcolor[HTML]{EFEFEF}0.34} & \cellcolor[HTML]{EFEFEF}0.35 \\ \cline{2-8} 
\multicolumn{1}{|c|}{}                                   & \cellcolor[HTML]{C0C0C0}Night 2   & \multicolumn{1}{c|}{\cellcolor[HTML]{C0C0C0}\textbf{0.39}} & \multicolumn{1}{c|}{\cellcolor[HTML]{C0C0C0}0.45} & \cellcolor[HTML]{C0C0C0}0.43 & \multicolumn{1}{c|}{\cellcolor[HTML]{C0C0C0}\textbf{0.33}} & \multicolumn{1}{c|}{\cellcolor[HTML]{C0C0C0}0.36} & \cellcolor[HTML]{C0C0C0}0.37 \\ \cline{2-8} 
\multicolumn{1}{|c|}{\multirow{-6}{*}{\rotatebox[origin=c]{90}{Broadway}}}         & \cellcolor[HTML]{EFEFEF}Night 3   & \multicolumn{1}{c|}{\cellcolor[HTML]{EFEFEF}\textbf{0.39}} & \multicolumn{1}{c|}{\cellcolor[HTML]{EFEFEF}0.50} & \cellcolor[HTML]{EFEFEF}0.49 & \multicolumn{1}{c|}{\cellcolor[HTML]{EFEFEF}\textbf{0.32}} & \multicolumn{1}{c|}{\cellcolor[HTML]{EFEFEF}0.36} & \cellcolor[HTML]{EFEFEF}0.37 \\ \hline
\multicolumn{1}{|c|}{}                                   & \cellcolor[HTML]{C0C0C0}Sunrise   & \multicolumn{1}{c|}{\cellcolor[HTML]{C0C0C0}\textbf{0.45}} & \multicolumn{1}{c|}{\cellcolor[HTML]{C0C0C0}0.58} & \cellcolor[HTML]{C0C0C0}0.53 & \multicolumn{1}{c|}{\cellcolor[HTML]{C0C0C0}\textbf{0.33}} & \multicolumn{1}{c|}{\cellcolor[HTML]{C0C0C0}0.37} & \cellcolor[HTML]{C0C0C0}0.38 \\ \cline{2-8} 
\multicolumn{1}{|c|}{}                                   & \cellcolor[HTML]{EFEFEF}Daytime   & \multicolumn{1}{c|}{\cellcolor[HTML]{EFEFEF}\textbf{0.49}} & \multicolumn{1}{c|}{\cellcolor[HTML]{EFEFEF}0.68} & \cellcolor[HTML]{EFEFEF}0.64 & \multicolumn{1}{c|}{\cellcolor[HTML]{EFEFEF}\textbf{0.37}} & \multicolumn{1}{c|}{\cellcolor[HTML]{EFEFEF}0.43} & \cellcolor[HTML]{EFEFEF}0.43 \\ \cline{2-8} 
\multicolumn{1}{|c|}{}                                   & \cellcolor[HTML]{C0C0C0}Sunset    & \multicolumn{1}{c|}{\cellcolor[HTML]{C0C0C0}\textbf{0.55}} & \multicolumn{1}{c|}{\cellcolor[HTML]{C0C0C0}0.67} & \cellcolor[HTML]{C0C0C0}0.63 & \multicolumn{1}{c|}{\cellcolor[HTML]{C0C0C0}\textbf{0.36}} & \multicolumn{1}{c|}{\cellcolor[HTML]{C0C0C0}0.39} & \cellcolor[HTML]{C0C0C0}0.42 \\ \cline{2-8} 
\multicolumn{1}{|c|}{}                                   & \cellcolor[HTML]{EFEFEF}Night 1   & \multicolumn{1}{c|}{\cellcolor[HTML]{EFEFEF}\textbf{0.46}} & \multicolumn{1}{c|}{\cellcolor[HTML]{EFEFEF}0.62} & \cellcolor[HTML]{EFEFEF}0.58 & \multicolumn{1}{c|}{\cellcolor[HTML]{EFEFEF}\textbf{0.34}} & \multicolumn{1}{c|}{\cellcolor[HTML]{EFEFEF}0.39} & \cellcolor[HTML]{EFEFEF}0.40 \\ \cline{2-8} 
\multicolumn{1}{|c|}{\multirow{-5}{*}{\rotatebox[origin=c]{90}{Castro}}}    & \cellcolor[HTML]{C0C0C0}Night 2   & \multicolumn{1}{c|}{\cellcolor[HTML]{C0C0C0}\textbf{0.47}} & \multicolumn{1}{c|}{\cellcolor[HTML]{C0C0C0}0.64} & \cellcolor[HTML]{C0C0C0}0.60 & \multicolumn{1}{c|}{\cellcolor[HTML]{C0C0C0}\textbf{0.35}} & \multicolumn{1}{c|}{\cellcolor[HTML]{C0C0C0}0.40} & \cellcolor[HTML]{C0C0C0}0.41 \\ \hline
\multicolumn{1}{l|}{}                                    & Average                           & \multicolumn{1}{c|}{\textbf{0.43}}                                  & \multicolumn{1}{c|}{0.52}                         & 0.49                         & \multicolumn{1}{c|}{\textbf{0.35}}                                  & \multicolumn{1}{c|}{0.37}                         & 0.39                         \\ \cline{2-8}
\end{tabular}%
\end{minipage}

\hspace{5pt}

\begin{minipage}{0.6\textwidth}
\centering
\subcaption{RTMDet}
\label{tab:lpea-results-rtm}
\begin{tabular}{c|c|ccc|ccc|}
\cline{2-8}
                                                         &                                   & \multicolumn{3}{c|}{Max Confidence}                                                                                                           & \multicolumn{3}{c|}{Average Confidence}                                                                                                       \\ \cline{3-8} 
\multirow{-2}{*}{}                                       & \multirow{-2}{*}{Video}           & \multicolumn{1}{c|}{L-PET}                                & \multicolumn{1}{c|}{Man. Dist.}                   & Random                       & \multicolumn{1}{c|}{L-PET}                                & \multicolumn{1}{c|}{Man. Dist.}                   & Random                       \\ \hline
\multicolumn{1}{|c|}{}                                   & \cellcolor[HTML]{C0C0C0}Daytime 1 & \multicolumn{1}{c|}{\cellcolor[HTML]{C0C0C0}\textbf{0.43}} & \multicolumn{1}{c|}{\cellcolor[HTML]{C0C0C0}0.45} & \cellcolor[HTML]{C0C0C0}0.44 & \multicolumn{1}{c|}{\cellcolor[HTML]{C0C0C0}\textbf{0.34}} & \multicolumn{1}{c|}{\cellcolor[HTML]{C0C0C0}0.37} & \cellcolor[HTML]{C0C0C0}0.39 \\ \cline{2-8} 
\multicolumn{1}{|c|}{}                                   & \cellcolor[HTML]{EFEFEF}Daytime 2 & \multicolumn{1}{c|}{\cellcolor[HTML]{EFEFEF}\textbf{0.51}} & \multicolumn{1}{c|}{\cellcolor[HTML]{EFEFEF}0.52} & \cellcolor[HTML]{EFEFEF}0.53 & \multicolumn{1}{c|}{\cellcolor[HTML]{EFEFEF}\textbf{0.36}} & \multicolumn{1}{c|}{\cellcolor[HTML]{EFEFEF}0.41} & \cellcolor[HTML]{EFEFEF}0.46 \\ \cline{2-8} 
\multicolumn{1}{|c|}{}                                   & \cellcolor[HTML]{C0C0C0}Afternoon & \multicolumn{1}{c|}{\cellcolor[HTML]{C0C0C0}\textbf{0.36}} & \multicolumn{1}{c|}{\cellcolor[HTML]{C0C0C0}0.38} & \cellcolor[HTML]{C0C0C0}0.38 & \multicolumn{1}{c|}{\cellcolor[HTML]{C0C0C0}\textbf{0.32}} & \multicolumn{1}{c|}{\cellcolor[HTML]{C0C0C0}0.33} & \cellcolor[HTML]{C0C0C0}0.34 \\ \cline{2-8} 
\multicolumn{1}{|c|}{}                                   & \cellcolor[HTML]{EFEFEF}Evening   & \multicolumn{1}{c|}{\cellcolor[HTML]{EFEFEF}\textbf{0.38}} & \multicolumn{1}{c|}{\cellcolor[HTML]{EFEFEF}0.44} & \cellcolor[HTML]{EFEFEF}0.40 & \multicolumn{1}{c|}{\cellcolor[HTML]{EFEFEF}\textbf{0.32}} & \multicolumn{1}{c|}{\cellcolor[HTML]{EFEFEF}0.34} & \cellcolor[HTML]{EFEFEF}0.36 \\ \cline{2-8} 
\multicolumn{1}{|c|}{}                                   & \cellcolor[HTML]{C0C0C0}Night 1   & \multicolumn{1}{c|}{\cellcolor[HTML]{C0C0C0}\textbf{0.42}} & \multicolumn{1}{c|}{\cellcolor[HTML]{C0C0C0}0.44} & \cellcolor[HTML]{C0C0C0}0.44 & \multicolumn{1}{c|}{\cellcolor[HTML]{C0C0C0}\textbf{0.33}} & \multicolumn{1}{c|}{\cellcolor[HTML]{C0C0C0}0.36} & \cellcolor[HTML]{C0C0C0}0.39 \\ \cline{2-8} 
\multicolumn{1}{|c|}{\multirow{-6}{*}{\rotatebox[origin=c]{90}{Shibuya}}} & \cellcolor[HTML]{EFEFEF}Night 2   & \multicolumn{1}{c|}{\cellcolor[HTML]{EFEFEF}\textbf{0.42}} & \multicolumn{1}{c|}{\cellcolor[HTML]{EFEFEF}0.49} & \cellcolor[HTML]{EFEFEF}0.43 & \multicolumn{1}{c|}{\cellcolor[HTML]{EFEFEF}\textbf{0.33}} & \multicolumn{1}{c|}{\cellcolor[HTML]{EFEFEF}0.36} & \cellcolor[HTML]{EFEFEF}0.39 \\ \hline
\multicolumn{1}{|c|}{}                                   & \cellcolor[HTML]{C0C0C0}Daytime 1 & \multicolumn{1}{c|}{\cellcolor[HTML]{C0C0C0}\textbf{0.38}} & \multicolumn{1}{c|}{\cellcolor[HTML]{C0C0C0}0.41} & \cellcolor[HTML]{C0C0C0}0.41 & \multicolumn{1}{c|}{\cellcolor[HTML]{C0C0C0}\textbf{0.31}} & \multicolumn{1}{c|}{\cellcolor[HTML]{C0C0C0}0.34} & \cellcolor[HTML]{C0C0C0}0.36 \\ \cline{2-8} 
\multicolumn{1}{|c|}{}                                   & \cellcolor[HTML]{EFEFEF}Daytime 2 & \multicolumn{1}{c|}{\cellcolor[HTML]{EFEFEF}\textbf{0.36}} & \multicolumn{1}{c|}{\cellcolor[HTML]{EFEFEF}0.41} & \cellcolor[HTML]{EFEFEF}0.41 & \multicolumn{1}{c|}{\cellcolor[HTML]{EFEFEF}\textbf{0.31}} & \multicolumn{1}{c|}{\cellcolor[HTML]{EFEFEF}0.33} & \cellcolor[HTML]{EFEFEF}0.35 \\ \cline{2-8} 
\multicolumn{1}{|c|}{}                                   & \cellcolor[HTML]{C0C0C0}Daytime 3 & \multicolumn{1}{c|}{\cellcolor[HTML]{C0C0C0}\textbf{0.38}} & \multicolumn{1}{c|}{\cellcolor[HTML]{C0C0C0}0.42} & \cellcolor[HTML]{C0C0C0}0.43 & \multicolumn{1}{c|}{\cellcolor[HTML]{C0C0C0}\textbf{0.32}} & \multicolumn{1}{c|}{\cellcolor[HTML]{C0C0C0}0.35} & \cellcolor[HTML]{C0C0C0}0.36 \\ \cline{2-8} 
\multicolumn{1}{|c|}{}                                   & \cellcolor[HTML]{EFEFEF}Night 1   & \multicolumn{1}{c|}{\cellcolor[HTML]{EFEFEF}\textbf{0.35}} & \multicolumn{1}{c|}{\cellcolor[HTML]{EFEFEF}0.38} & \cellcolor[HTML]{EFEFEF}0.39 & \multicolumn{1}{c|}{\cellcolor[HTML]{EFEFEF}\textbf{0.30}} & \multicolumn{1}{c|}{\cellcolor[HTML]{EFEFEF}0.32} & \cellcolor[HTML]{EFEFEF}0.34 \\ \cline{2-8} 
\multicolumn{1}{|c|}{}                                   & \cellcolor[HTML]{C0C0C0}Night 2   & \multicolumn{1}{c|}{\cellcolor[HTML]{C0C0C0}\textbf{0.35}} & \multicolumn{1}{c|}{\cellcolor[HTML]{C0C0C0}0.37} & \cellcolor[HTML]{C0C0C0}0.38 & \multicolumn{1}{c|}{\cellcolor[HTML]{C0C0C0}\textbf{0.30}} & \multicolumn{1}{c|}{\cellcolor[HTML]{C0C0C0}0.32} & \cellcolor[HTML]{C0C0C0}0.34 \\ \cline{2-8} 
\multicolumn{1}{|c|}{\multirow{-6}{*}{\rotatebox[origin=c]{90}{Broadway}}}         & \cellcolor[HTML]{EFEFEF}Night 3   & \multicolumn{1}{c|}{\cellcolor[HTML]{EFEFEF}\textbf{0.36}} & \multicolumn{1}{c|}{\cellcolor[HTML]{EFEFEF}0.37} & \cellcolor[HTML]{EFEFEF}0.39 & \multicolumn{1}{c|}{\cellcolor[HTML]{EFEFEF}\textbf{0.31}} & \multicolumn{1}{c|}{\cellcolor[HTML]{EFEFEF}0.33} & \cellcolor[HTML]{EFEFEF}0.34 \\ \hline
\multicolumn{1}{|c|}{}                                   & \cellcolor[HTML]{C0C0C0}Sunrise   & \multicolumn{1}{c|}{\cellcolor[HTML]{C0C0C0}\textbf{0.38}} & \multicolumn{1}{c|}{\cellcolor[HTML]{C0C0C0}0.52} & \cellcolor[HTML]{C0C0C0}0.54 & \multicolumn{1}{c|}{\cellcolor[HTML]{C0C0C0}\textbf{0.31}} & \multicolumn{1}{c|}{\cellcolor[HTML]{C0C0C0}0.35} & \cellcolor[HTML]{C0C0C0}0.36 \\ \cline{2-8} 
\multicolumn{1}{|c|}{}                                   & \cellcolor[HTML]{EFEFEF}Daytime   & \multicolumn{1}{c|}{\cellcolor[HTML]{EFEFEF}\textbf{0.38}} & \multicolumn{1}{c|}{\cellcolor[HTML]{EFEFEF}0.45} & \cellcolor[HTML]{EFEFEF}0.45 & \multicolumn{1}{c|}{\cellcolor[HTML]{EFEFEF}\textbf{0.31}} & \multicolumn{1}{c|}{\cellcolor[HTML]{EFEFEF}0.34} & \cellcolor[HTML]{EFEFEF}0.36 \\ \cline{2-8} 
\multicolumn{1}{|c|}{}                                   & \cellcolor[HTML]{C0C0C0}Sunset    & \multicolumn{1}{c|}{\cellcolor[HTML]{C0C0C0}\textbf{0.44}} & \multicolumn{1}{c|}{\cellcolor[HTML]{C0C0C0}0.53} & \cellcolor[HTML]{C0C0C0}0.54 & \multicolumn{1}{c|}{\cellcolor[HTML]{C0C0C0}\textbf{0.33}} & \multicolumn{1}{c|}{\cellcolor[HTML]{C0C0C0}0.39} & \cellcolor[HTML]{C0C0C0}0.40 \\ \cline{2-8} 
\multicolumn{1}{|c|}{}                                   & \cellcolor[HTML]{EFEFEF}Night 1   & \multicolumn{1}{c|}{\cellcolor[HTML]{EFEFEF}\textbf{0.40}} & \multicolumn{1}{c|}{\cellcolor[HTML]{EFEFEF}0.51} & \cellcolor[HTML]{EFEFEF}0.53 & \multicolumn{1}{c|}{\cellcolor[HTML]{EFEFEF}\textbf{0.32}} & \multicolumn{1}{c|}{\cellcolor[HTML]{EFEFEF}0.35} & \cellcolor[HTML]{EFEFEF}0.37 \\ \cline{2-8} 
\multicolumn{1}{|c|}{\multirow{-5}{*}{\rotatebox[origin=c]{90}{Castro}}}    & \cellcolor[HTML]{C0C0C0}Night 2   & \multicolumn{1}{c|}{\cellcolor[HTML]{C0C0C0}\textbf{0.39}} & \multicolumn{1}{c|}{\cellcolor[HTML]{C0C0C0}0.50} & \cellcolor[HTML]{C0C0C0}0.53 & \multicolumn{1}{c|}{\cellcolor[HTML]{C0C0C0}\textbf{0.32}} & \multicolumn{1}{c|}{\cellcolor[HTML]{C0C0C0}0.35} & \cellcolor[HTML]{C0C0C0}0.37 \\ \hline
\multicolumn{1}{l|}{}                                    & Average                           & \multicolumn{1}{c|}{\textbf{0.39}}                                  & \multicolumn{1}{c|}{0.45}                         & 0.45                         & \multicolumn{1}{c|}{\textbf{0.32}}                                  & \multicolumn{1}{c|}{0.35}                         & 0.37                         \\ \cline{2-8}
\end{tabular}%
\end{minipage}
}
\end{table*}

\begin{table*}[th!]
\centering
\caption{Average maximum confidence and average per-step confidence of 100 paths generated between 10 random start points and 10 random end points on the observed scene at different times of day. 100 paths were generated for each group: (1) paths generated with L-PET, (2) random direct paths of Manhattan distance from start to end, and (3) random paths. The paths were generated on a scene observed by an object detector enhanced with the L-BAT countermeasure.}
\label{tab:lbat-lpea-results-combined}
\resizebox{2.0\columnwidth}{!}{%
\begin{minipage}{0.6\textwidth}
\centering
\subcaption{Faster R-CNN}
\label{tab:lbat-lpea-results-faster}
\begin{tabular}{c|c|ccc|ccc|}
\cline{2-8}
                                                         &                                   & \multicolumn{3}{c|}{Max Confidence}                                                                                                           & \multicolumn{3}{c|}{Average Confidence}                                                                                                       \\ \cline{3-8} 
\multirow{-2}{*}{}                                       & \multirow{-2}{*}{Video}           & \multicolumn{1}{c|}{L-PET}                                & \multicolumn{1}{c|}{Man. Dist.}                   & Random                       & \multicolumn{1}{c|}{L-PET}                                & \multicolumn{1}{c|}{Man. Dist.}                   & Random                       \\ \hline
\multicolumn{1}{|c|}{}                                   & \cellcolor[HTML]{C0C0C0}Daytime 1 & \multicolumn{1}{c|}{\cellcolor[HTML]{C0C0C0}\textbf{0.88}} & \multicolumn{1}{c|}{\cellcolor[HTML]{C0C0C0}0.89} & \cellcolor[HTML]{C0C0C0}0.90 & \multicolumn{1}{c|}{\cellcolor[HTML]{C0C0C0}\textbf{0.65}} & \multicolumn{1}{c|}{\cellcolor[HTML]{C0C0C0}0.72} & \cellcolor[HTML]{C0C0C0}0.84 \\ \cline{2-8} 
\multicolumn{1}{|c|}{}                                   & \cellcolor[HTML]{EFEFEF}Daytime 2 & \multicolumn{1}{c|}{\cellcolor[HTML]{EFEFEF}\textbf{0.90}} & \multicolumn{1}{c|}{\cellcolor[HTML]{EFEFEF}0.91} & \cellcolor[HTML]{EFEFEF}0.92 & \multicolumn{1}{c|}{\cellcolor[HTML]{EFEFEF}\textbf{0.67}} & \multicolumn{1}{c|}{\cellcolor[HTML]{EFEFEF}0.73} & \cellcolor[HTML]{EFEFEF}0.83 \\ \cline{2-8} 
\multicolumn{1}{|c|}{}                                   & \cellcolor[HTML]{C0C0C0}Afternoon & \multicolumn{1}{c|}{\cellcolor[HTML]{C0C0C0}\textbf{0.76}} & \multicolumn{1}{c|}{\cellcolor[HTML]{C0C0C0}0.80} & \cellcolor[HTML]{C0C0C0}0.79 & \multicolumn{1}{c|}{\cellcolor[HTML]{C0C0C0}\textbf{0.56}} & \multicolumn{1}{c|}{\cellcolor[HTML]{C0C0C0}0.62} & \cellcolor[HTML]{C0C0C0}0.70 \\
\end{tabular}%
\end{minipage}
\hspace{5pt}
\begin{minipage}{0.6\textwidth}
\centering
\subcaption{YOLOv3}
\label{tab:lbat-lpea-results-yolo}
\begin{tabular}{c|c|ccc|ccc|}
\cline{2-8}
                                                         &                                   & \multicolumn{3}{c|}{Max Confidence}                                                                                                           & \multicolumn{3}{c|}{Average Confidence}                                                                                                       \\ \cline{3-8} 
\multirow{-2}{*}{}                                       & \multirow{-2}{*}{Video}           & \multicolumn{1}{c|}{L-PET}                                & \multicolumn{1}{c|}{Man. Dist.}                   & Random                       & \multicolumn{1}{c|}{L-PET}                                & \multicolumn{1}{c|}{Man. Dist.}                   & Random                       \\ \hline
\multicolumn{1}{|c|}{}                                   & \cellcolor[HTML]{C0C0C0}Daytime 1 & \multicolumn{1}{c|}{\cellcolor[HTML]{C0C0C0}\textbf{0.61}} & \multicolumn{1}{c|}{\cellcolor[HTML]{C0C0C0}0.63} & \cellcolor[HTML]{C0C0C0}0.65 & \multicolumn{1}{c|}{\cellcolor[HTML]{C0C0C0}\textbf{0.41}} & \multicolumn{1}{c|}{\cellcolor[HTML]{C0C0C0}0.46} & \cellcolor[HTML]{C0C0C0}0.55 \\ \cline{2-8} 
\multicolumn{1}{|c|}{}                                   & \cellcolor[HTML]{EFEFEF}Daytime 2 & \multicolumn{1}{c|}{\cellcolor[HTML]{EFEFEF}\textbf{0.65}} & \multicolumn{1}{c|}{\cellcolor[HTML]{EFEFEF}0.72} & \cellcolor[HTML]{EFEFEF}0.69 & \multicolumn{1}{c|}{\cellcolor[HTML]{EFEFEF}\textbf{0.45}} & \multicolumn{1}{c|}{\cellcolor[HTML]{EFEFEF}0.50} & \cellcolor[HTML]{EFEFEF}0.57 \\ \cline{2-8} 
\multicolumn{1}{|c|}{}                                   & \cellcolor[HTML]{C0C0C0}Afternoon & \multicolumn{1}{c|}{\cellcolor[HTML]{C0C0C0}\textbf{0.51}} & \multicolumn{1}{c|}{\cellcolor[HTML]{C0C0C0}0.62} & \cellcolor[HTML]{C0C0C0}0.54 & \multicolumn{1}{c|}{\cellcolor[HTML]{C0C0C0}\textbf{0.37}} & \multicolumn{1}{c|}{\cellcolor[HTML]{C0C0C0}0.39} & \cellcolor[HTML]{C0C0C0}0.45 \\
\end{tabular}%
\end{minipage}
}
\newline
\vspace*{0.15 cm}
\newline
\resizebox{2.0\columnwidth}{!}{%
\noindent
\begin{minipage}{0.6\textwidth}
\centering
\subcaption{SSD}
\label{tab:lbat-lpea-results-ssd}
\begin{tabular}{c|c|ccc|ccc|}
\cline{2-8}
                                                         &                                   & \multicolumn{3}{c|}{Max Confidence}                                                                                                           & \multicolumn{3}{c|}{Average Confidence}                                                                                                       \\ \cline{3-8} 
\multirow{-2}{*}{}                                       & \multirow{-2}{*}{Video}           & \multicolumn{1}{c|}{L-PET}                                & \multicolumn{1}{c|}{Man. Dist.}                   & Random                       & \multicolumn{1}{c|}{L-PET}                                & \multicolumn{1}{c|}{Man. Dist.}                   & Random                       \\ \hline
\multicolumn{1}{|c|}{}                                   & \cellcolor[HTML]{C0C0C0}Daytime 1 & \multicolumn{1}{c|}{\cellcolor[HTML]{C0C0C0}\textbf{0.43}} & \multicolumn{1}{c|}{\cellcolor[HTML]{C0C0C0}0.45} & \cellcolor[HTML]{C0C0C0}0.44 & \multicolumn{1}{c|}{\cellcolor[HTML]{C0C0C0}\textbf{0.39}} & \multicolumn{1}{c|}{\cellcolor[HTML]{C0C0C0}0.40} & \cellcolor[HTML]{C0C0C0}0.41 \\ \cline{2-8} 
\multicolumn{1}{|c|}{}                                   & \cellcolor[HTML]{EFEFEF}Daytime 2 & \multicolumn{1}{c|}{\cellcolor[HTML]{EFEFEF}\textbf{0.48}} & \multicolumn{1}{c|}{\cellcolor[HTML]{EFEFEF}0.52} & \cellcolor[HTML]{EFEFEF}0.50 & \multicolumn{1}{c|}{\cellcolor[HTML]{EFEFEF}\textbf{0.42}} & \multicolumn{1}{c|}{\cellcolor[HTML]{EFEFEF}0.43} & \cellcolor[HTML]{EFEFEF}0.44 \\ \cline{2-8} 
\multicolumn{1}{|c|}{}                                   & \cellcolor[HTML]{C0C0C0}Afternoon & \multicolumn{1}{c|}{\cellcolor[HTML]{C0C0C0}\textbf{0.40}} & \multicolumn{1}{c|}{\cellcolor[HTML]{C0C0C0}0.45} & \cellcolor[HTML]{C0C0C0}0.41 & \multicolumn{1}{c|}{\cellcolor[HTML]{C0C0C0}\textbf{0.37}} & \multicolumn{1}{c|}{\cellcolor[HTML]{C0C0C0}0.38} & \cellcolor[HTML]{C0C0C0}0.38 \\
\end{tabular}%
\end{minipage}

\hspace{5pt}

\begin{minipage}{0.6\textwidth}
\centering
\subcaption{RTMDet}
\label{tab:lbat-lpea-results-rtm}
\begin{tabular}{c|c|ccc|ccc|}
\cline{2-8}
                                                         &                                   & \multicolumn{3}{c|}{Max Confidence}                                                                                                           & \multicolumn{3}{c|}{Average Confidence}                                                                                                       \\ \cline{3-8} 
\multirow{-2}{*}{}                                       & \multirow{-2}{*}{Video}           & \multicolumn{1}{c|}{L-PET}                                & \multicolumn{1}{c|}{Man. Dist.}                   & Random                       & \multicolumn{1}{c|}{L-PET}                                & \multicolumn{1}{c|}{Man. Dist.}                   & Random                       \\ \hline
\multicolumn{1}{|c|}{}                                   & \cellcolor[HTML]{C0C0C0}Daytime 1 & \multicolumn{1}{c|}{\cellcolor[HTML]{C0C0C0}\textbf{0.47}} & \multicolumn{1}{c|}{\cellcolor[HTML]{C0C0C0}0.48} & \cellcolor[HTML]{C0C0C0}0.48 & \multicolumn{1}{c|}{\cellcolor[HTML]{C0C0C0}\textbf{0.35}} & \multicolumn{1}{c|}{\cellcolor[HTML]{C0C0C0}0.38} & \cellcolor[HTML]{C0C0C0}0.42 \\ \cline{2-8} 
\multicolumn{1}{|c|}{}                                   & \cellcolor[HTML]{EFEFEF}Daytime 2 & \multicolumn{1}{c|}{\cellcolor[HTML]{EFEFEF}\textbf{0.51}} & \multicolumn{1}{c|}{\cellcolor[HTML]{EFEFEF}0.52} & \cellcolor[HTML]{EFEFEF}0.53 & \multicolumn{1}{c|}{\cellcolor[HTML]{EFEFEF}\textbf{0.37}} & \multicolumn{1}{c|}{\cellcolor[HTML]{EFEFEF}0.40} & \cellcolor[HTML]{EFEFEF}0.45 \\ \cline{2-8} 
\multicolumn{1}{|c|}{}                                   & \cellcolor[HTML]{C0C0C0}Afternoon & \multicolumn{1}{c|}{\cellcolor[HTML]{C0C0C0}\textbf{0.40}} & \multicolumn{1}{c|}{\cellcolor[HTML]{C0C0C0}0.40} & \cellcolor[HTML]{C0C0C0}0.40 & \multicolumn{1}{c|}{\cellcolor[HTML]{C0C0C0}\textbf{0.33}} & \multicolumn{1}{c|}{\cellcolor[HTML]{C0C0C0}0.35} & \cellcolor[HTML]{C0C0C0}0.37 \\
\end{tabular}%
\end{minipage}
}
\end{table*}

\begin{table*}[th!]
\centering
\caption{AUC score, true positive rate (TPR), false positive rate (FPR), and average true positive confidence of (1) an untouched object detection model,
and (2) an object detection model utilizing L-BAT.}
\label{tab:evaluation-results-lbat-performance-combined}
\resizebox{2.0\columnwidth}{!}{%
\begin{minipage}{0.65\textwidth}
\centering
\subcaption{Faster R-CNN}
\label{tab:evaluation-results-lbat-performance-faster}
\begin{tabular}{c|c|cccc|cccc|}
\cline{2-10}
                                                 &                                   & \multicolumn{4}{c|}{Original Model}                                                                                                                                                                                 & \multicolumn{4}{c|}{Original Model with L-BAT}                                                                                                                                                                      \\ \cline{3-10} 
\multirow{-2}{*}{}                               & \multirow{-2}{*}{Video}           & \multicolumn{1}{c|}{AUC}                                   & \multicolumn{1}{c|}{TPR}                                   & \multicolumn{1}{c|}{FPR}                                   & \makecell{Avg. TP \\ Confidence}           & \multicolumn{1}{c|}{AUC}                                   & \multicolumn{1}{c|}{TPR}                                   & \multicolumn{1}{c|}{FPR}                          & \makecell{Avg. TP \\ Confidence}                    \\ \hline
\multicolumn{1}{|c|}{}                           & \cellcolor[HTML]{C0C0C0}Daytime 1 & \multicolumn{1}{c|}{\cellcolor[HTML]{C0C0C0}0.83}          & \multicolumn{1}{c|}{\cellcolor[HTML]{C0C0C0}0.94}          & \multicolumn{1}{c|}{\cellcolor[HTML]{C0C0C0}\textbf{0.29}} & \cellcolor[HTML]{C0C0C0}0.76 & \multicolumn{1}{c|}{\cellcolor[HTML]{C0C0C0}\textbf{0.84}} & \multicolumn{1}{c|}{\cellcolor[HTML]{C0C0C0}\textbf{0.97}} & \multicolumn{1}{c|}{\cellcolor[HTML]{C0C0C0}0.32} & \cellcolor[HTML]{C0C0C0}\textbf{0.93} \\
\end{tabular}%
\end{minipage}
\hspace{5pt}
\begin{minipage}{0.65\textwidth}
\centering
\subcaption{YOLOv3}
\label{tab:evaluation-results-lbat-performance-yolo}
\begin{tabular}{c|c|cccc|cccc|}
\cline{2-10}
                                                 &                                   & \multicolumn{4}{c|}{Original Model}                                                                                                                                                               & \multicolumn{4}{c|}{Original Model with L-BAT}                                                                                                                                                                      \\ \cline{3-10} 
\multirow{-2}{*}{}                               & \multirow{-2}{*}{Video}           & \multicolumn{1}{c|}{AUC}                          & \multicolumn{1}{c|}{TPR}                          & \multicolumn{1}{c|}{FPR}                                   & \makecell{Avg. TP \\ Confidence}           & \multicolumn{1}{c|}{AUC}                                   & \multicolumn{1}{c|}{TPR}                                   & \multicolumn{1}{c|}{FPR}                          & \makecell{Avg. TP \\ Confidence}                    \\ \hline
\multicolumn{1}{|c|}{}                           & \cellcolor[HTML]{C0C0C0}Daytime 1 & \multicolumn{1}{c|}{\cellcolor[HTML]{C0C0C0}0.62} & \multicolumn{1}{c|}{\cellcolor[HTML]{C0C0C0}0.35} & \multicolumn{1}{c|}{\cellcolor[HTML]{C0C0C0}\textbf{0.10}} & \cellcolor[HTML]{C0C0C0}0.45 & \multicolumn{1}{c|}{\cellcolor[HTML]{C0C0C0}\textbf{0.74}} & \multicolumn{1}{c|}{\cellcolor[HTML]{C0C0C0}\textbf{0.66}} & \multicolumn{1}{c|}{\cellcolor[HTML]{C0C0C0}0.17} & \cellcolor[HTML]{C0C0C0}\textbf{0.67} \\
\end{tabular}%
\end{minipage}
}
\newline
\vspace*{0.15 cm}
\newline
\resizebox{2.0\columnwidth}{!}{%
\noindent
\begin{minipage}{0.65\textwidth}
\centering
\subcaption{SSD}
\label{tab:evaluation-results-lbat-performance-ssd}
\begin{tabular}{c|c|cccc|cccc|}
\cline{2-10}
                                                 &                                   & \multicolumn{4}{c|}{Original Model}                                                                                                                                                               & \multicolumn{4}{c|}{Original Model with L-BAT}                                                                                                                                                                      \\ \cline{3-10} 
\multirow{-2}{*}{}                               & \multirow{-2}{*}{Video}           & \multicolumn{1}{c|}{AUC}                          & \multicolumn{1}{c|}{TPR}                          & \multicolumn{1}{c|}{FPR}                                   & \makecell{Avg. TP \\ Confidence}           & \multicolumn{1}{c|}{AUC}                                   & \multicolumn{1}{c|}{TPR}                                   & \multicolumn{1}{c|}{FPR}                          & \makecell{Avg. TP \\ Confidence}                    \\ \hline
\multicolumn{1}{|c|}{}                           & \cellcolor[HTML]{C0C0C0}Daytime 1 & \multicolumn{1}{c|}{\cellcolor[HTML]{C0C0C0}0.63} & \multicolumn{1}{c|}{\cellcolor[HTML]{C0C0C0}0.37} & \multicolumn{1}{c|}{\cellcolor[HTML]{C0C0C0}\textbf{0.11}} & \cellcolor[HTML]{C0C0C0}0.40 & \multicolumn{1}{c|}{\cellcolor[HTML]{C0C0C0}\textbf{0.80}} & \multicolumn{1}{c|}{\cellcolor[HTML]{C0C0C0}\textbf{0.87}} & \multicolumn{1}{c|}{\cellcolor[HTML]{C0C0C0}0.29} & \cellcolor[HTML]{C0C0C0}\textbf{0.63} \\
\end{tabular}%
\end{minipage}

\hspace{5pt}

\begin{minipage}{0.65\textwidth}
\centering
\subcaption{RTMDet}
\label{tab:evaluation-results-lbat-performance-rtm}
\begin{tabular}{c|c|cccc|cccc|}
\cline{2-10}
                                                 &                                   & \multicolumn{4}{c|}{Original Model}                                                                                                                                                               & \multicolumn{4}{c|}{Original Model with L-BAT}                                                                                                                                                                      \\ \cline{3-10} 
\multirow{-2}{*}{}                               & \multirow{-2}{*}{Video}           & \multicolumn{1}{c|}{AUC}                          & \multicolumn{1}{c|}{TPR}                          & \multicolumn{1}{c|}{FPR}                                   & \makecell{Avg. TP \\ Confidence}           & \multicolumn{1}{c|}{AUC}                                   & \multicolumn{1}{c|}{TPR}                                   & \multicolumn{1}{c|}{FPR}                          & \makecell{Avg. TP \\ Confidence}                    \\ \hline
\multicolumn{1}{|c|}{}                           & \cellcolor[HTML]{C0C0C0}Daytime 1 & \multicolumn{1}{c|}{\cellcolor[HTML]{C0C0C0}0.64} & \multicolumn{1}{c|}{\cellcolor[HTML]{C0C0C0}0.47} & \multicolumn{1}{c|}{\cellcolor[HTML]{C0C0C0}\textbf{0.17}} & \cellcolor[HTML]{C0C0C0}0.37 & \multicolumn{1}{c|}{\cellcolor[HTML]{C0C0C0}\textbf{0.82}} & \multicolumn{1}{c|}{\cellcolor[HTML]{C0C0C0}\textbf{0.95}} & \multicolumn{1}{c|}{\cellcolor[HTML]{C0C0C0}0.32} & \cellcolor[HTML]{C0C0C0}\textbf{0.76} \\
\end{tabular}%
\end{minipage}
}
\end{table*}

\subsection{Experimental Setup:}

We utilized publicly available stationary recordings of pedestrian traffic through three public locations: (1) Shibuya Crossing in Tokyo, Japan, (2) Broadway, New York, and (3) Castro Street, San Francisco, California, displayed in Fig. \ref{fig:observed-scenes}.
We utilized 4 hour recordings of the scenes in various lighting conditions (day, night, afternoon, evening, etc.), as well as five different object detectors: (1) Faster R-CNN, (2) YOLOv3, (3) SSD, (4) DiffusionDet, and (5) RTMDet.
For each 4 hour recording and object detector, we calculated a confidence heatmap, and detection heatmap according to the methodology presented in 
Algorithm \ref{alg:generate-heatmaps} and Section \ref{sec:lpea-methodology}.

For each scene represented by a confidence heatmap, we took 10 randomly-chosen start and end points on the scene, each represented by coordinates $(x,y)$ on the confidence heatmap. On the set of 100 paths between each pair of start and end points, we calculated the average: (1) max node confidence of the path, and (2) per-step confidence of the path. We calculated these values for paths generated using three path-finding strategies: (1) L-PET, (2) a random shortest-distance path, and (3) a random length path.

\subsection{Evaluation and Results}

Table \ref{tab:lpea-results-combined} presents the results of this evaluation for the five object detectors. We can see that paths generated by L-PET consistently have an average lower max path confidence and average confidence. For Faster R-CNN, L-PET produces as much as a 0.23 drop in max confidence, compared to random direct/indirect paths. In addition, L-PET produces as much as a 0.19 drop in average confidence, compared to random direct/indirect paths. These results are similar for the other object detectors as well.

\textbf{Impact of Object Detector:}
We can see in the results presented in Table \ref{tab:lpea-results-combined} that, on average, the max and average path confidence score reductions caused by L-PET were highest for Faster R-CNN (0.09 and 0.13), but there were also similar max/average path confidence reductions for the other models: YOLOv3 (0.08 and 0.07), DiffusionDet (0.08 and 0.09), SSD (0.09 and 0.04), and RTMDet (0.06 and 0.05). This indicates that L-PET finds paths of reduced confidence for a variety of object detection models.

\textbf{Impact of Time of Day:}
According to the results presented in Table \ref{tab:lpea-results-combined}, for Shibuya Crossing footage analyzed by Faster R-CNN, we notice the max and average path confidences tends to be approximately 0.05 higher in the daytime/afternoon videos than the night videos. However, the average max confidence reductions (0.03) and average confidence reductions (0.14) are the same for both times of day. We note that similar confidence reduction behavior also occurs in the other evaluations with different locations and object detectors. These findings indicate that L-PET is effective at finding paths of reduced confidence in various lighting conditions/times of day.

\textbf{Impact of Location:}
According to the results presented in Table \ref{tab:lpea-results-combined}, for footage analyzed by Faster R-CNN, we notice reductions of max/average path confidence in footage recorded in Shibuya Crossing (0.04 and 0.14), Broadway (0.08 and 0.15) and Castro Street (0.15 and 0.09). While the exact reduction amount differs, we note both max and average path confidence reduction occurs across all evaluated locations.

\textbf{Conclusions:} Compared to direct (Manhattan distance) and random paths, paths generated by L-PET have lower: (1) maximum path confidence, and (2) average path confidence. These findings demonstrate that a privacy-concerned pedestrian can exploit the confidence level of an object detector by leveraging a person’s location as they traverse the scene. In addition, these findings persist across various object detectors, times of day, and global locations.

\section{L-BAT: Location-Based Adaptive Threshold}
\label{sec:evaluation-lbat}

In this section, we introduce and evaluate L-BAT, our proposed countermeasure. L-BAT is intended to reduce the inherent location-based vulnerability of object detectors that is leveraged by L-PET, and require investment in more costly and sophisticated methods to exploit object detectors. We evaluate L-BAT's impact on: (1) the performance of L-PET, and (2) the performance of the underlying pedestrian detector.

\begin{algorithm}[t!]
\caption{L-BAT}
\begin{flushleft}
\textbf{Input:} $I, M, V$. \\
\textbf{Output:} $BB_{new}$ - bounding boxes with updated confidence scores.
\end{flushleft}
\begin{algorithmic}[1]
\State $H_c, H_d \gets \text{GenerateHeatmaps}(V,M)$
\State $BB \gets \text{bounding boxes from feeding $I$ to $M$}$.
\State $BB_{new} \gets \{\}$
\For{$bb \in BB, bb\text{ is a person bounding box}$}
    \State $x_1,x_2,y_1,y_2 \gets \text{coordinates of $bb$}$.
    \State $c_{bb} \gets \text{confidence score of $bb$}$.
    \State $scores \gets \{H_c(x,y)| x \in [x_1, .. ,x_2], y \in [y_1, .. ,y_2]\}$
    \State $t \gets \frac{\Sigma scores}{|scores|}$
    \State $c_{new} \gets min(1, c_{bb} / t)$
    \State $BB_{new}.append(bb\text{ with confidence } c_{new})$
\EndFor
\textbf{return:} $BB_{new}$
\end{algorithmic}
\label{alg:lbat}
\end{algorithm}

\subsection{L-BAT Methodology}
\label{sec:lbat-methodology}

In this section, we present the methodology of our countermeasure, the Location-Based Adaptive Threshold (L-BAT) for Pedestrian Detection Systems.
L-BAT receives three inputs: (1) $I$ - an image of the scene the pedestrian detection system is observing, (2) $M$ - the AI model utilized in the pedestrian detection system, and (3) $V$ - prerecorded footage of the scene.

The first step of L-BAT is to generate the confidence heatmap $H_c$ from the model $M$ and prerecorded footage $V$. This process is the same as described above in Sec. \ref{sec:lpea-methodology}.

The next step of L-BAT is the inference step. $I$ is fed to $M$, and the resulting 'person' bounding boxes' confidences are then updated according to their location in the scene. 
$H_c$ records for each pixel the average confidence score of 'person' detections landing on the pixel.
For each bounding box from $I$, L-BAT calculates the average heatmap confidence of the pixels covered by the bounding box by averaging the bounding box pixels' respective confidence measurements in $H_c$.
We then divide the bounding box's original confidence by this value. This updated confidence measurement increases the confidence of detections in 'blind spots' with low confidence, mitigating the impact of location-based confidence reduction.
This confidence update is performed for each 'person' detection bounding box.

The output of L-BAT is an enhanced pedestrian detection system, which updates the confidence scores of pedestrian detections according to the location of the pedestrian detection on the heat-map.
This is performed with a dynamic location-dependent detection threshold to update the confidence scores, rather than depending on a universal detection threshold set in the object detector.
Algorithm \ref{alg:lbat} presents the methodology of L-BAT.

\subsection{Impact on L-PET Performance}

In this section, we evaluate the impact of L-BAT, our proposed countermeasure, against the L-PET privacy-enhancing technique.

\subsubsection{Experimental Setup}

We repeated the experimental setup from Section \ref{sec:evaluation-lpea}, replacing the unprotected pedestrian detector model with a model defended by L-BAT.

\subsubsection{Evaluation and Results} Table \ref{tab:lbat-lpea-results-combined} presents the results of this evaluation for the five object detectors. For Faster R-CNN, we can see the range of max confidences for paths generated by L-PET increased, from [0.59, 0.89] to [0.71, 0.90]. On average, the max confidence of paths generated by L-PET increased by 0.09. Similarly, the range of average confidences for paths generated by L-PET increased, from [0.41, 0.64] to [0.51, 0.67]. On average, the average confidence of paths generated by L-PET increased by 0.05. In addition, the max/average confidences of direct paths and random paths increased by 0.06/0.07 and 0.06/0.07, respectively. We also see similar results for the other object detectors.

\textbf{Impact of Object Detector:}
We can see in the results presented in Table \ref{tab:lbat-lpea-results-combined} that, on average, the max and average path confidence score increases caused by L-BAT were highest for Faster R-CNN (0.09 and 0.05), but there were also similar max/average path confidence reductions for the other models: YOLOv3 (0.04 and 0.01), DiffusionDet (0.03 and 0.01), SSD (<0.01 and <0.01), and RTMDet (0.02 and <0.01).
We note that the increases in max path confidence tended to be larger than the increases in average path confidence. This is an indication that while L-PET finds paths with similar average confidence, it is still confronted with larger max confidence bottlenecks which must be crossed, increasing the likelihood of detection.
We also note that object detectors which originally had lower path confidences (e.g., SSD, RTMDet) had less significant confidence improvements. We hypothesize that this is due to L-PET succeeding to find more successful paths in an area with already reduced confidence. We recommend taking these object detectors' low confidence behavior into account to ensure proper detection.
These findings indicate that L-BAT increases the confidence of L-PET paths for a variety of object detection models.

\textbf{Impact of Time of Day:}
According to the results presented in Table \ref{tab:lbat-lpea-results-combined}, for Shibuya Crossing footage analyzed by Faster R-CNN, we notice that after applying L-BAT, the max and average path confidences tends to be approximately 0.05 higher in the daytime/afternoon videos than the night videos. However, the average max confidence increase (0.10/0.09) and average confidence increase (0.07/0.05) are slightly higher for the daytime videos. These findings indicate that L-BAT is effective at finding paths of reduced confidence in various lighting conditions/times of day.

\textbf{Impact of Location:}
According to the results presented in Table \ref{tab:lbat-lpea-results-combined}, for footage analyzed by Faster R-CNN, we notice L-BAT increases the L-PET max/average path confidence in footage from Shibuya Crossing (0.10 and 0.06), Broadway (0.10 and 0.06) and Castro Street (0.05 and 0.04). While the exact increase differs, we note both max and average path confidence increase occurs across all evaluated locations.

\textbf{Conclusions:} Based on our findings, we conclude that L-BAT is an effective countermeasure against L-PET, successfully raising the average max path confidence by 0.09 and the average path confidence by 0.05 on Faster R-CNN. 
In addition, these findings persist across various object detectors, times of day, and global locations.

\subsection{Impact on Pedestrian Detection Performance}

Here, we evaluate the impact of L-BAT on the performance of the protected pedestrian detector.

\subsubsection{Experimental Setup}

We utilized the 4 hour videos of public pedestrian traffic used in the previous evaluations. We built a dataset of pedestrian images according to the following methodology: (1) We ran one frame $f$ every two seconds in an SSD object detector. (2) For each $f$, we chose one person bounding box as a positive sample $p_f$, and designated a random rectangular area where no people were detected in the frame as a negative sample $n_f$. (3) We added these two samples (one positive and one negative) to the dataset.

We then evaluated two models, a Faster R-CNN model not utilizing L-BAT, and a Faster R-CNN model utilizing L-BAT. We determined the models' performance according to the following methodology: (1) for every frame $f$, if the model generated a bounding box which intersected with $p_f$, $p_f$ is considered a \textbf{true positive}, otherwise it's considered a \textbf{false negative}. (2) for every frame $f$, if the model generated a bounding box which intersected with $n_f$, $n_f$ is considered a \textbf{false positive}, otherwise it's considered a \textbf{true negative}. Using these values, we calculated the AUC of each model, as well as: (1) true positive rate (TPR), (2) false positive rate (FPR), and (3) average true positive (TP) confidence. We repeated this evaluation for each of the six videos of Shibuya Crossing used in the previous experiments.

\subsubsection{Evaluation and Results}

Table \ref{tab:evaluation-results-lbat-performance-combined} presents the results of this evaluation. We found that using L-BAT with a Faster R-CNN model maintains an average AUC of 0.89, while increasing the average TPR from 0.93 to 0.96, and the average true positive (TP) confidence increasing from 0.79 to 0.94. This performance improvement comes at the cost of an increased average FPR, from 0.19 to 0.24.

\textbf{Impact of Object Detector:}
We can see in the results presented in Table \ref{tab:evaluation-results-lbat-performance-combined} that, on average, the AUC increase caused by L-BAT is between 0.10 and 0.20 for every object detector, with the exception of Faster R-CNN, which maintains the same AUC. We hypothesize that this is due to Faster R-CNN starting with a very high AUC (0.89). In addition, all the evaluated object detectors demonstrated significant increases in TPR (increase of between 0.03-0.54) and average TP confidence (increase of between 0.15-0.42). These performance increases come at a cost of an increased FPR (increase of between 0.04-0.15). This indicates that L-BAT's performance improvements persist for a variety of object detection models.

\textbf{Impact of Time of Day:}
According to the results presented in Table \ref{tab:evaluation-results-lbat-performance-combined}, for Shibuya Crossing footage analyzed by Faster R-CNN, we notice the AUC, TPR, and TP confidence tend to be higher, and the FPR lower, in the daytime/afternoon videos than the night videos. We see similar improvements in AUC (0.01 and 0.01), TPR (0.03 and 0.06), and TP confidence (0.16 and 0.18), as well as similar penalties in FPR (0.04 and 0.05). We note that similar performance improvements also occur in the other evaluations with different locations and object detectors. These findings indicate that L-BAT is effective at reducing location-based drops in performance in various lighting conditions/times of day.

\textbf{Impact of Location:}
According to the results presented in Table \ref{tab:evaluation-results-lbat-performance-combined}, for footage analyzed by Faster R-CNN, we notice performance improvements in AUC/TPR/TP confidence in footage recorded in Shibuya Crossing (0.02/0.05/0.17), Broadway (0.0/0.02/0.15) and Castro Street (0.01/0.02/0.11). While the exact performance improvement differs, we note improvement in AUC, TPR, and TP confidence, along with a slight penalty to FPR, occurs across all evaluated locations.

\textbf{Conclusions:} Based on these results, we conclude that utilizing L-BAT improves the performance of a pedestrian detector by improving the ability to detect people in weaker detection areas, where a pedestrian detector would either not detect pedestrians or detect them with low confidence. L-BAT improves the AUC, TPR, and average TP confidence (at the cost of a slightly increased FPR), which represents an improved ability of detecting pedestrians across a diverse scene. In addition, these findings persist across various object detectors, times of day, and global locations.

\section{Related Work}
\label{sec:related-works}

Security attacks on AI models are attacks which exploit vulnerabilities in AI models to cause undesired behavior. This is often performed with adversarial patches, which utilize the model's gradient to produce a pattern which triggers unexpected behavior in the model. One type of security attacks are \textit{evasion attacks}, which leverage adversarial patches and/or target detection sensors to change/avoid detection by AI systems.

There are multiple evasion attacks that generate patches to evade an object detector, leveraging algorithms such as Expectation over Transformation \cite{athalye2018synthesizing} in order to produce adversarial patches which can evade facial detection systems \cite{sharif2016accessorize,Komkov_2021,ijcai2021p173,zolfi2022adversarial,wang2020avdpattern}, as well as autonomous driver assistance systems (ADASs) \cite{sitawarin2018darts,Zhao2019SeeingIB,274691}, automatic check-out systems \cite{liu2020biasbaseduniversaladversarialpatch,wang2022adversarial}, and object detectors \cite{kurakin2016adversarial,athalye2018synthesizing,Chen_2019,eykholt2018robust,thys2019fooling,wu2020making,xu2020adversarial,Hu2022Adversarial,doan2022tnt,eykholt2018physicaladversarialexamplesobject,lee2019physicaladversarialpatchesobject,Yang_Tsai_Yu_Ho_Jin_2020,zolfi2020translucentpatchphysicaluniversal,hoory2020dynamicadversarialpatchevading}. 

The method presented in \cite{sharif2016accessorize} is a misclassification attack against facial recognition systems utilizing adversarial patterns printed on eyeglass frames to misclassify a face randomly or misclassify as a target face. In addition, \cite{Komkov_2021} performs a misclassification attack on facial recognition models by placing an adversarial patch on a hat. Also, \cite{zolfi2022adversarial} performs a universal evasion attack against facial recognition models with an adversarial perturbation printed on a face mask.
Against autonomous vehicle systems, evasion attacks can be utilized to cause these systems to not classify or misclassify traffic signs \cite{sitawarin2018darts,Zhao2019SeeingIB} or falsely perceive road lanes \cite{274691}. \cite{athalye2018synthesizing} presented physical patches to induce misclassification of a turtle as a rifle, while \cite{Chen_2019} and \cite{eykholt2018robust} apply patches to street signs to prevent detection by an object detector. Finally, \cite{thys2019fooling,Hu2022Adversarial,xu2020adversarial,wu2020making} generate adversarial patches to avoid detection of people by an object detector.
These attacks have proven to be a threat to AI systems deployed in the real world and used for critical safety tasks, such as traffic sign detection \cite{9565475}.

There are also attacks which, rather than generate adversarial patches, target sensors in order to evade detection by an object detector. These methods utilize various tools, such as projectors \cite{lovisotto2021slap,shen2019vla,nassi2020phantom}, lasers \cite{sato2024invisible} and infrared light \cite{zhou2018invisible,sato2024invisible,10.1145/3460120.3484766,Zhu_Li_Li_Wang_Hu_2021,zhu2022infraredinvisibleclothinghidinginfrared} in order to cause undesired behavior in object detectors. This behavior includes: (1) failure to detect faces \cite{zhou2018invisible}, (2) failure to detect cars and street signs \cite{10.1145/3460120.3484766,sato2024invisible,lovisotto2021slap}, (3) falsely perceive a traffic light \cite{10.1145/3460120.3484766}, and (4) falsely perceive the surrounding environment \cite{10.1145/3460120.3484766,nassi2020phantom}.
\section{Limitations}
\label{sec:limitations}

The limitations of the L-BAT countermeasure are as follows:
\textbf{(1) Lack of heatmap transferability} - The confidence heatmaps utilized in L-BAT are generated using video footage of a scene at a particular time of day. This results in the heatmap being relevant only for the same location and lighting conditions. In particular, the confidence heatmaps cannot be used for different times of day and/or locations.
\textbf{(2) Not robust against adversarial attacks} - L-BAT is designed to counter L-PET, which is a privacy-enhancing technique not requiring sophisticated adversarial methods (e.g., generating patches). L-BAT is intended to require investing more efforts in sophisticated methods in order to successfully exploit an object detector. In particular, L-BAT is not intended to be robust against sophisticated adversarial attacks on object detectors.
\textbf{(3) Lack of heatmap transferability to new model architectures} - The confidence heatmaps utilized in L-BAT are intended for one specific object detector model architecture. In particular, confidence heatmaps generated for different object detector architectures can't be interchanged.

\section{Conclusions and Future Work}
\label{sec:conclusion}

In this paper, we demonstrate a novel vulnerability in AI-based pedestrian detection systems, that the position of a pedestrian in an observed scene directly impacts the confidence of the model.
We demonstrate that the angle and distance of a pedestrian, as well as the height of the pedestrian detector and ambient light level, result in varying confidence levels in the pedestrian detector.
We then demonstrate that in areas with large amounts of pedestrian traffic, this phenomenon can be observed in areas of weaker detection in the scene, where pedestrian detections have lower confidence than in other areas of the scene.

Given these findings, we propose L-PET, a location-based privacy-enhancing technique which generates a path between two points in the scene, leveraging weak detection areas to achieve lower (1) max confidence, and (2) average confidence than both direct and randomly generated paths.
To counter L-PET, we propose the L-BAT countermeasure, which compensates weak areas in a scene by defining the confidence of a pedestrian detection be a weighted function of: (1) the pedestrian detector's confidence and (2) the average confidence of detections from the same area.
We demonstrate that L-BAT mitigates the performance of L-PET, while improving the ability of the pedestrian object detector to detect pedestrians across the scene.

For future work, we propose research on the location-based confidence reduction phenomenon in additional use-cases, such as autonomous vehicles.

\bibliographystyle{IEEEtran}
\bibliography{bibliography}

\pagebreak
\section{Appendix}
\label{sec:appendix}

\subsection{Effect of Pedestrian Position on Pedestrian Detection Systems - Additional Figures}
\label{sec:scene-analysis-appendix}

\begin{figure}[h]
\centering
       \begin{minipage}{0.99\textwidth}
          \includegraphics[width=0.16\textwidth]{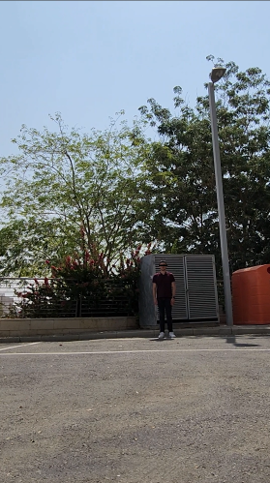} 
     \includegraphics[width=0.16\textwidth]{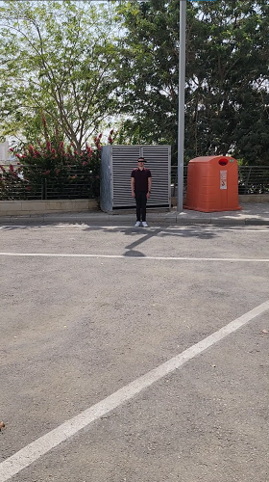}
      \includegraphics[width=0.16\textwidth]{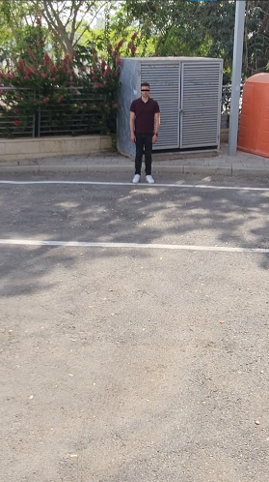} 
    \end{minipage} 
    \newline
    \vspace{2pt}
       \begin{minipage}{0.99\textwidth}
          \includegraphics[width=0.16\textwidth]{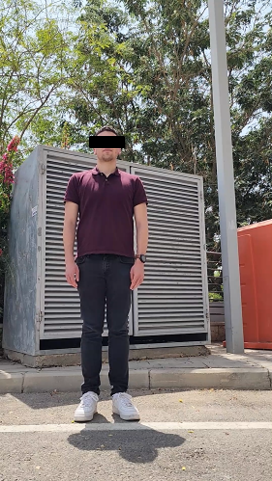} 
     \includegraphics[width=0.16\textwidth]{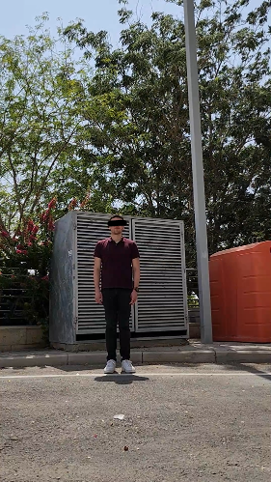}
      \includegraphics[width=0.16\textwidth]{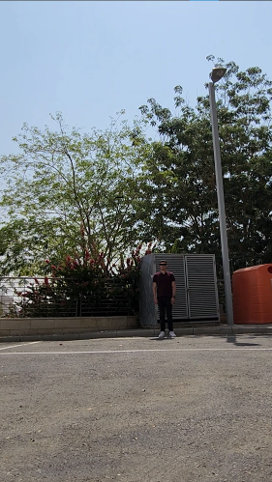} 
    \end{minipage} 
    \newline
    \vspace{2pt}
       \begin{minipage}{0.99\textwidth}
          \includegraphics[width=0.16\textwidth]{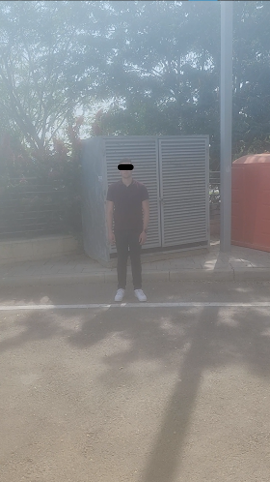} 
     \includegraphics[width=0.16\textwidth]{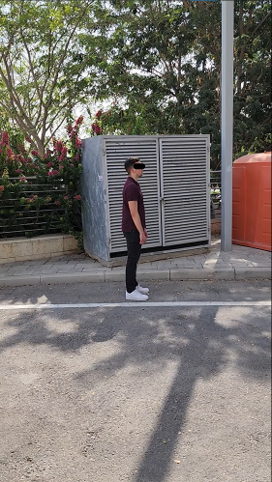}
      \includegraphics[width=0.16\textwidth]{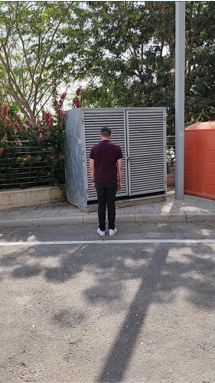} 
    \end{minipage}  
    \caption{The positioning of the person and camera in the analysis. Top row: the three analyzed heights. Middle row: the three analyzed distances. Bottom row: Three of the eight analyzed angles.}
    \label{fig:exp-setup-visualization}
\end{figure}

\onecolumn

\begin{minipage}{\textwidth}

\captionof{table}{The confidence of the DiffusionDet object detector for each recorded position (height, distance, angle). Each row represents a configuration of camera height and distance, and each column represents an angle the person stood at relative to the camera. Cells representing positions which recorded average confidences of \textgreater{}0.8, 0.6-0.8, and \textless{}0.6 are presented in green, yellow, and red, respectively.}
\label{tab:standing-results-dnet}
\resizebox{1.0\columnwidth}{!}{%
\begin{tabular}{cccccccccc}
\multicolumn{10}{c}{Lab Conditions}                                                                                                                                                                                                                                                                                                                                                                                                                                                                                              \\ \hline
\multicolumn{1}{|c|}{}                         & \multicolumn{1}{c|}{}                           & \multicolumn{8}{c|}{Angle}                                                                                                                                                                                                                                                                                                                                                                                                    \\ \cline{3-10} 
\multicolumn{1}{|c|}{\multirow{-2}{*}{Height}} & \multicolumn{1}{c|}{\multirow{-2}{*}{Distance}} & \multicolumn{1}{c|}{0\degree}                            & \multicolumn{1}{c|}{45\degree}                           & \multicolumn{1}{c|}{90\degree}                           & \multicolumn{1}{c|}{135\degree}                          & \multicolumn{1}{c|}{180\degree}                          & \multicolumn{1}{c|}{225\degree}                          & \multicolumn{1}{c|}{270\degree}                          & \multicolumn{1}{c|}{315\degree}                          \\ \hline
\multicolumn{1}{|c|}{}                         & \multicolumn{1}{c|}{2.5m}                       & \multicolumn{1}{c|}{\cellcolor[HTML]{67FD9A}0.97} & \multicolumn{1}{c|}{\cellcolor[HTML]{67FD9A}0.96} & \multicolumn{1}{c|}{\cellcolor[HTML]{67FD9A}0.94} & \multicolumn{1}{c|}{\cellcolor[HTML]{67FD9A}0.95} & \multicolumn{1}{c|}{\cellcolor[HTML]{67FD9A}0.96} & \multicolumn{1}{c|}{\cellcolor[HTML]{67FD9A}0.97} & \multicolumn{1}{c|}{\cellcolor[HTML]{67FD9A}0.97} & \multicolumn{1}{c|}{\cellcolor[HTML]{67FD9A}0.97} \\ \cline{2-10} 
\multicolumn{1}{|c|}{}                         & \multicolumn{1}{c|}{5m}                         & \multicolumn{1}{c|}{\cellcolor[HTML]{67FD9A}0.94} & \multicolumn{1}{c|}{\cellcolor[HTML]{67FD9A}0.96} & \multicolumn{1}{c|}{\cellcolor[HTML]{67FD9A}0.89} & \multicolumn{1}{c|}{\cellcolor[HTML]{67FD9A}0.95} & \multicolumn{1}{c|}{\cellcolor[HTML]{67FD9A}0.95} & \multicolumn{1}{c|}{\cellcolor[HTML]{67FD9A}0.94} & \multicolumn{1}{c|}{\cellcolor[HTML]{67FD9A}0.90} & \multicolumn{1}{c|}{\cellcolor[HTML]{67FD9A}0.94} \\ \cline{2-10} 
\multicolumn{1}{|c|}{\multirow{-3}{*}{0.6m}}   & \multicolumn{1}{c|}{10m}                        & \multicolumn{1}{c|}{\cellcolor[HTML]{67FD9A}0.87} & \multicolumn{1}{c|}{\cellcolor[HTML]{67FD9A}0.94} & \multicolumn{1}{c|}{\cellcolor[HTML]{67FD9A}0.88} & \multicolumn{1}{c|}{\cellcolor[HTML]{67FD9A}0.89} & \multicolumn{1}{c|}{\cellcolor[HTML]{67FD9A}0.83} & \multicolumn{1}{c|}{\cellcolor[HTML]{67FD9A}0.92} & \multicolumn{1}{c|}{\cellcolor[HTML]{67FD9A}0.93} & \multicolumn{1}{c|}{\cellcolor[HTML]{67FD9A}0.95} \\ \hline
\multicolumn{1}{|c|}{}                         & \multicolumn{1}{c|}{2.5m}                       & \multicolumn{1}{c|}{\cellcolor[HTML]{67FD9A}0.97} & \multicolumn{1}{c|}{\cellcolor[HTML]{67FD9A}0.97} & \multicolumn{1}{c|}{\cellcolor[HTML]{67FD9A}0.97} & \multicolumn{1}{c|}{\cellcolor[HTML]{67FD9A}0.97} & \multicolumn{1}{c|}{\cellcolor[HTML]{67FD9A}0.96} & \multicolumn{1}{c|}{\cellcolor[HTML]{67FD9A}0.86} & \multicolumn{1}{c|}{\cellcolor[HTML]{67FD9A}0.97} & \multicolumn{1}{c|}{\cellcolor[HTML]{67FD9A}0.97} \\ \cline{2-10} 
\multicolumn{1}{|c|}{}                         & \multicolumn{1}{c|}{5m}                         & \multicolumn{1}{c|}{\cellcolor[HTML]{67FD9A}0.95} & \multicolumn{1}{c|}{\cellcolor[HTML]{67FD9A}0.97} & \multicolumn{1}{c|}{\cellcolor[HTML]{67FD9A}0.94} & \multicolumn{1}{c|}{\cellcolor[HTML]{67FD9A}0.94} & \multicolumn{1}{c|}{\cellcolor[HTML]{67FD9A}0.89} & \multicolumn{1}{c|}{\cellcolor[HTML]{67FD9A}0.94} & \multicolumn{1}{c|}{\cellcolor[HTML]{67FD9A}0.87} & \multicolumn{1}{c|}{\cellcolor[HTML]{67FD9A}0.95} \\ \cline{2-10} 
\multicolumn{1}{|c|}{\multirow{-3}{*}{1.8m}}   & \multicolumn{1}{c|}{10m}                        & \multicolumn{1}{c|}{\cellcolor[HTML]{67FD9A}0.93} & \multicolumn{1}{c|}{\cellcolor[HTML]{67FD9A}0.93} & \multicolumn{1}{c|}{\cellcolor[HTML]{67FD9A}0.93} & \multicolumn{1}{c|}{\cellcolor[HTML]{67FD9A}0.87} & \multicolumn{1}{c|}{\cellcolor[HTML]{FFFE65}0.75} & \multicolumn{1}{c|}{\cellcolor[HTML]{67FD9A}0.88} & \multicolumn{1}{c|}{\cellcolor[HTML]{67FD9A}0.95} & \multicolumn{1}{c|}{\cellcolor[HTML]{67FD9A}0.96} \\ \hline
\multicolumn{1}{|c|}{}                         & \multicolumn{1}{c|}{2.5m}                       & \multicolumn{1}{c|}{\cellcolor[HTML]{67FD9A}0.94} & \multicolumn{1}{c|}{\cellcolor[HTML]{67FD9A}0.93} & \multicolumn{1}{c|}{\cellcolor[HTML]{67FD9A}0.95} & \multicolumn{1}{c|}{\cellcolor[HTML]{67FD9A}0.97} & \multicolumn{1}{c|}{\cellcolor[HTML]{67FD9A}0.96} & \multicolumn{1}{c|}{\cellcolor[HTML]{67FD9A}0.96} & \multicolumn{1}{c|}{\cellcolor[HTML]{67FD9A}0.96} & \multicolumn{1}{c|}{\cellcolor[HTML]{67FD9A}0.96} \\ \cline{2-10} 
\multicolumn{1}{|c|}{}                         & \multicolumn{1}{c|}{5m}                         & \multicolumn{1}{c|}{\cellcolor[HTML]{67FD9A}0.92} & \multicolumn{1}{c|}{\cellcolor[HTML]{67FD9A}0.95} & \multicolumn{1}{c|}{\cellcolor[HTML]{67FD9A}0.91} & \multicolumn{1}{c|}{\cellcolor[HTML]{67FD9A}0.94} & \multicolumn{1}{c|}{\cellcolor[HTML]{67FD9A}0.92} & \multicolumn{1}{c|}{\cellcolor[HTML]{67FD9A}0.93} & \multicolumn{1}{c|}{\cellcolor[HTML]{67FD9A}0.91} & \multicolumn{1}{c|}{\cellcolor[HTML]{67FD9A}0.94} \\ \cline{2-10} 
\multicolumn{1}{|c|}{\multirow{-3}{*}{2.4m}}   & \multicolumn{1}{c|}{10m}                        & \multicolumn{1}{c|}{\cellcolor[HTML]{67FD9A}0.92} & \multicolumn{1}{c|}{\cellcolor[HTML]{67FD9A}0.91} & \multicolumn{1}{c|}{\cellcolor[HTML]{67FD9A}0.93} & \multicolumn{1}{c|}{\cellcolor[HTML]{67FD9A}0.91} & \multicolumn{1}{c|}{\cellcolor[HTML]{FFFE65}0.65} & \multicolumn{1}{c|}{\cellcolor[HTML]{67FD9A}0.90} & \multicolumn{1}{c|}{\cellcolor[HTML]{67FD9A}0.94} & \multicolumn{1}{c|}{\cellcolor[HTML]{67FD9A}0.93} \\ \hline
\end{tabular}%
\hspace{.1pt}
\begin{tabular}{cccccccccc}
\multicolumn{10}{c}{Morning}                                                                                                                                                                                                                                                                                                                                                                                                                                                                                                     \\ \hline
\multicolumn{1}{|c|}{}                         & \multicolumn{1}{c|}{}                           & \multicolumn{8}{c|}{Angle}                                                                                                                                                                                                                                                                                                                                                                                                    \\ \cline{3-10} 
\multicolumn{1}{|c|}{\multirow{-2}{*}{Height}} & \multicolumn{1}{c|}{\multirow{-2}{*}{Distance}} & \multicolumn{1}{c|}{0\degree}                            & \multicolumn{1}{c|}{45\degree}                           & \multicolumn{1}{c|}{90\degree}                           & \multicolumn{1}{c|}{135\degree}                          & \multicolumn{1}{c|}{180\degree}                          & \multicolumn{1}{c|}{225\degree}                          & \multicolumn{1}{c|}{270\degree}                          & \multicolumn{1}{c|}{315\degree}                          \\ \hline
\multicolumn{1}{|c|}{}                         & \multicolumn{1}{c|}{2.5m}                       & \multicolumn{1}{c|}{\cellcolor[HTML]{67FD9A}0.97} & \multicolumn{1}{c|}{\cellcolor[HTML]{67FD9A}0.96} & \multicolumn{1}{c|}{\cellcolor[HTML]{67FD9A}0.95} & \multicolumn{1}{c|}{\cellcolor[HTML]{67FD9A}0.96} & \multicolumn{1}{c|}{\cellcolor[HTML]{67FD9A}0.97} & \multicolumn{1}{c|}{\cellcolor[HTML]{67FD9A}0.97} & \multicolumn{1}{c|}{\cellcolor[HTML]{67FD9A}0.97} & \multicolumn{1}{c|}{\cellcolor[HTML]{67FD9A}0.98} \\ \cline{2-10} 
\multicolumn{1}{|c|}{}                         & \multicolumn{1}{c|}{5m}                         & \multicolumn{1}{c|}{\cellcolor[HTML]{67FD9A}0.97} & \multicolumn{1}{c|}{\cellcolor[HTML]{67FD9A}0.97} & \multicolumn{1}{c|}{\cellcolor[HTML]{67FD9A}0.96} & \multicolumn{1}{c|}{\cellcolor[HTML]{67FD9A}0.96} & \multicolumn{1}{c|}{\cellcolor[HTML]{67FD9A}0.96} & \multicolumn{1}{c|}{\cellcolor[HTML]{67FD9A}0.96} & \multicolumn{1}{c|}{\cellcolor[HTML]{67FD9A}0.97} & \multicolumn{1}{c|}{\cellcolor[HTML]{67FD9A}0.97} \\ \cline{2-10} 
\multicolumn{1}{|c|}{\multirow{-3}{*}{0.6m}}   & \multicolumn{1}{c|}{10m}                        & \multicolumn{1}{c|}{\cellcolor[HTML]{67FD9A}0.97} & \multicolumn{1}{c|}{\cellcolor[HTML]{67FD9A}0.97} & \multicolumn{1}{c|}{\cellcolor[HTML]{67FD9A}0.97} & \multicolumn{1}{c|}{\cellcolor[HTML]{67FD9A}0.97} & \multicolumn{1}{c|}{\cellcolor[HTML]{67FD9A}0.97} & \multicolumn{1}{c|}{\cellcolor[HTML]{67FD9A}0.96} & \multicolumn{1}{c|}{\cellcolor[HTML]{67FD9A}0.96} & \multicolumn{1}{c|}{\cellcolor[HTML]{FE0000}0.58} \\ \hline
\multicolumn{1}{|c|}{}                         & \multicolumn{1}{c|}{2.5m}                       & \multicolumn{1}{c|}{\cellcolor[HTML]{67FD9A}0.97} & \multicolumn{1}{c|}{\cellcolor[HTML]{67FD9A}0.97} & \multicolumn{1}{c|}{\cellcolor[HTML]{67FD9A}0.97} & \multicolumn{1}{c|}{\cellcolor[HTML]{67FD9A}0.97} & \multicolumn{1}{c|}{\cellcolor[HTML]{67FD9A}0.97} & \multicolumn{1}{c|}{\cellcolor[HTML]{67FD9A}0.97} & \multicolumn{1}{c|}{\cellcolor[HTML]{67FD9A}0.96} & \multicolumn{1}{c|}{\cellcolor[HTML]{67FD9A}0.97} \\ \cline{2-10} 
\multicolumn{1}{|c|}{}                         & \multicolumn{1}{c|}{5m}                         & \multicolumn{1}{c|}{\cellcolor[HTML]{67FD9A}0.96} & \multicolumn{1}{c|}{\cellcolor[HTML]{67FD9A}0.97} & \multicolumn{1}{c|}{\cellcolor[HTML]{67FD9A}0.97} & \multicolumn{1}{c|}{\cellcolor[HTML]{67FD9A}0.97} & \multicolumn{1}{c|}{\cellcolor[HTML]{67FD9A}0.96} & \multicolumn{1}{c|}{\cellcolor[HTML]{67FD9A}0.96} & \multicolumn{1}{c|}{\cellcolor[HTML]{67FD9A}0.97} & \multicolumn{1}{c|}{\cellcolor[HTML]{67FD9A}0.97} \\ \cline{2-10} 
\multicolumn{1}{|c|}{\multirow{-3}{*}{1.8m}}   & \multicolumn{1}{c|}{10m}                        & \multicolumn{1}{c|}{\cellcolor[HTML]{67FD9A}0.95} & \multicolumn{1}{c|}{\cellcolor[HTML]{67FD9A}0.97} & \multicolumn{1}{c|}{\cellcolor[HTML]{67FD9A}0.97} & \multicolumn{1}{c|}{\cellcolor[HTML]{67FD9A}0.97} & \multicolumn{1}{c|}{\cellcolor[HTML]{67FD9A}0.96} & \multicolumn{1}{c|}{\cellcolor[HTML]{67FD9A}0.97} & \multicolumn{1}{c|}{\cellcolor[HTML]{67FD9A}0.97} & \multicolumn{1}{c|}{\cellcolor[HTML]{67FD9A}0.97} \\ \hline
\multicolumn{1}{|c|}{}                         & \multicolumn{1}{c|}{2.5m}                       & \multicolumn{1}{c|}{\cellcolor[HTML]{67FD9A}0.97} & \multicolumn{1}{c|}{\cellcolor[HTML]{67FD9A}0.97} & \multicolumn{1}{c|}{\cellcolor[HTML]{67FD9A}0.96} & \multicolumn{1}{c|}{\cellcolor[HTML]{67FD9A}0.97} & \multicolumn{1}{c|}{\cellcolor[HTML]{67FD9A}0.98} & \multicolumn{1}{c|}{\cellcolor[HTML]{67FD9A}0.98} & \multicolumn{1}{c|}{\cellcolor[HTML]{67FD9A}0.97} & \multicolumn{1}{c|}{\cellcolor[HTML]{67FD9A}0.97} \\ \cline{2-10} 
\multicolumn{1}{|c|}{}                         & \multicolumn{1}{c|}{5m}                         & \multicolumn{1}{c|}{\cellcolor[HTML]{67FD9A}0.96} & \multicolumn{1}{c|}{\cellcolor[HTML]{67FD9A}0.97} & \multicolumn{1}{c|}{\cellcolor[HTML]{67FD9A}0.96} & \multicolumn{1}{c|}{\cellcolor[HTML]{67FD9A}0.96} & \multicolumn{1}{c|}{\cellcolor[HTML]{67FD9A}0.97} & \multicolumn{1}{c|}{\cellcolor[HTML]{67FD9A}0.98} & \multicolumn{1}{c|}{\cellcolor[HTML]{67FD9A}0.98} & \multicolumn{1}{c|}{\cellcolor[HTML]{67FD9A}0.97} \\ \cline{2-10} 
\multicolumn{1}{|c|}{\multirow{-3}{*}{2.4m}}   & \multicolumn{1}{c|}{10m}                        & \multicolumn{1}{c|}{\cellcolor[HTML]{67FD9A}0.97} & \multicolumn{1}{c|}{\cellcolor[HTML]{67FD9A}0.97} & \multicolumn{1}{c|}{\cellcolor[HTML]{67FD9A}0.97} & \multicolumn{1}{c|}{\cellcolor[HTML]{67FD9A}0.97} & \multicolumn{1}{c|}{\cellcolor[HTML]{67FD9A}0.98} & \multicolumn{1}{c|}{\cellcolor[HTML]{67FD9A}0.98} & \multicolumn{1}{c|}{\cellcolor[HTML]{67FD9A}0.97} & \multicolumn{1}{c|}{\cellcolor[HTML]{67FD9A}0.98} \\ \hline
\end{tabular}%
}
\newline
\vspace*{0.03 cm}
\newline
\resizebox{1.0\columnwidth}{!}{%
\noindent
\begin{tabular}{cccccccccc}
\multicolumn{10}{c}{Afternoon}                                                                                                                                                                                                                                                                                                                                                                                                                                                                                                   \\ \hline
\multicolumn{1}{|c|}{}                         & \multicolumn{1}{c|}{}                           & \multicolumn{8}{c|}{Angle}                                                                                                                                                                                                                                                                                                                                                                                                    \\ \cline{3-10} 
\multicolumn{1}{|c|}{\multirow{-2}{*}{Height}} & \multicolumn{1}{c|}{\multirow{-2}{*}{Distance}} & \multicolumn{1}{c|}{0\degree}                            & \multicolumn{1}{c|}{45\degree}                           & \multicolumn{1}{c|}{90\degree}                           & \multicolumn{1}{c|}{135\degree}                          & \multicolumn{1}{c|}{180\degree}                          & \multicolumn{1}{c|}{225\degree}                          & \multicolumn{1}{c|}{270\degree}                          & \multicolumn{1}{c|}{315\degree}                          \\ \hline
\multicolumn{1}{|c|}{}                         & \multicolumn{1}{c|}{2.5m}                       & \multicolumn{1}{c|}{\cellcolor[HTML]{67FD9A}0.98} & \multicolumn{1}{c|}{\cellcolor[HTML]{67FD9A}0.96} & \multicolumn{1}{c|}{\cellcolor[HTML]{67FD9A}0.94} & \multicolumn{1}{c|}{\cellcolor[HTML]{67FD9A}0.96} & \multicolumn{1}{c|}{\cellcolor[HTML]{67FD9A}0.97} & \multicolumn{1}{c|}{\cellcolor[HTML]{67FD9A}0.97} & \multicolumn{1}{c|}{\cellcolor[HTML]{67FD9A}0.96} & \multicolumn{1}{c|}{\cellcolor[HTML]{67FD9A}0.97} \\ \cline{2-10} 
\multicolumn{1}{|c|}{}                         & \multicolumn{1}{c|}{5m}                         & \multicolumn{1}{c|}{\cellcolor[HTML]{67FD9A}0.97} & \multicolumn{1}{c|}{\cellcolor[HTML]{67FD9A}0.97} & \multicolumn{1}{c|}{\cellcolor[HTML]{67FD9A}0.97} & \multicolumn{1}{c|}{\cellcolor[HTML]{67FD9A}0.97} & \multicolumn{1}{c|}{\cellcolor[HTML]{67FD9A}0.97} & \multicolumn{1}{c|}{\cellcolor[HTML]{67FD9A}0.97} & \multicolumn{1}{c|}{\cellcolor[HTML]{67FD9A}0.98} & \multicolumn{1}{c|}{\cellcolor[HTML]{67FD9A}0.97} \\ \cline{2-10} 
\multicolumn{1}{|c|}{\multirow{-3}{*}{0.6m}}   & \multicolumn{1}{c|}{10m}                        & \multicolumn{1}{c|}{\cellcolor[HTML]{67FD9A}0.98} & \multicolumn{1}{c|}{\cellcolor[HTML]{67FD9A}0.97} & \multicolumn{1}{c|}{\cellcolor[HTML]{67FD9A}0.97} & \multicolumn{1}{c|}{\cellcolor[HTML]{67FD9A}0.97} & \multicolumn{1}{c|}{\cellcolor[HTML]{67FD9A}0.98} & \multicolumn{1}{c|}{\cellcolor[HTML]{67FD9A}0.97} & \multicolumn{1}{c|}{\cellcolor[HTML]{67FD9A}0.96} & \multicolumn{1}{c|}{\cellcolor[HTML]{67FD9A}0.97} \\ \hline
\multicolumn{1}{|c|}{}                         & \multicolumn{1}{c|}{2.5m}                       & \multicolumn{1}{c|}{\cellcolor[HTML]{67FD9A}0.97} & \multicolumn{1}{c|}{\cellcolor[HTML]{67FD9A}0.97} & \multicolumn{1}{c|}{\cellcolor[HTML]{67FD9A}0.97} & \multicolumn{1}{c|}{\cellcolor[HTML]{67FD9A}0.98} & \multicolumn{1}{c|}{\cellcolor[HTML]{67FD9A}0.97} & \multicolumn{1}{c|}{\cellcolor[HTML]{67FD9A}0.98} & \multicolumn{1}{c|}{\cellcolor[HTML]{67FD9A}0.97} & \multicolumn{1}{c|}{\cellcolor[HTML]{67FD9A}0.96} \\ \cline{2-10} 
\multicolumn{1}{|c|}{}                         & \multicolumn{1}{c|}{5m}                         & \multicolumn{1}{c|}{\cellcolor[HTML]{67FD9A}0.97} & \multicolumn{1}{c|}{\cellcolor[HTML]{67FD9A}0.98} & \multicolumn{1}{c|}{\cellcolor[HTML]{67FD9A}0.97} & \multicolumn{1}{c|}{\cellcolor[HTML]{67FD9A}0.96} & \multicolumn{1}{c|}{\cellcolor[HTML]{67FD9A}0.97} & \multicolumn{1}{c|}{\cellcolor[HTML]{67FD9A}0.97} & \multicolumn{1}{c|}{\cellcolor[HTML]{67FD9A}0.97} & \multicolumn{1}{c|}{\cellcolor[HTML]{67FD9A}0.97} \\ \cline{2-10} 
\multicolumn{1}{|c|}{\multirow{-3}{*}{1.8m}}   & \multicolumn{1}{c|}{10m}                        & \multicolumn{1}{c|}{\cellcolor[HTML]{67FD9A}0.97} & \multicolumn{1}{c|}{\cellcolor[HTML]{67FD9A}0.98} & \multicolumn{1}{c|}{\cellcolor[HTML]{67FD9A}0.96} & \multicolumn{1}{c|}{\cellcolor[HTML]{67FD9A}0.98} & \multicolumn{1}{c|}{\cellcolor[HTML]{67FD9A}0.97} & \multicolumn{1}{c|}{\cellcolor[HTML]{67FD9A}0.97} & \multicolumn{1}{c|}{\cellcolor[HTML]{67FD9A}0.97} & \multicolumn{1}{c|}{\cellcolor[HTML]{67FD9A}0.97} \\ \hline
\multicolumn{1}{|c|}{}                         & \multicolumn{1}{c|}{2.5m}                       & \multicolumn{1}{c|}{\cellcolor[HTML]{67FD9A}0.96} & \multicolumn{1}{c|}{\cellcolor[HTML]{67FD9A}0.97} & \multicolumn{1}{c|}{\cellcolor[HTML]{67FD9A}0.97} & \multicolumn{1}{c|}{\cellcolor[HTML]{67FD9A}0.97} & \multicolumn{1}{c|}{\cellcolor[HTML]{67FD9A}0.97} & \multicolumn{1}{c|}{\cellcolor[HTML]{67FD9A}0.97} & \multicolumn{1}{c|}{\cellcolor[HTML]{67FD9A}0.97} & \multicolumn{1}{c|}{\cellcolor[HTML]{67FD9A}0.97} \\ \cline{2-10} 
\multicolumn{1}{|c|}{}                         & \multicolumn{1}{c|}{5m}                         & \multicolumn{1}{c|}{\cellcolor[HTML]{67FD9A}0.96} & \multicolumn{1}{c|}{\cellcolor[HTML]{67FD9A}0.98} & \multicolumn{1}{c|}{\cellcolor[HTML]{67FD9A}0.96} & \multicolumn{1}{c|}{\cellcolor[HTML]{67FD9A}0.97} & \multicolumn{1}{c|}{\cellcolor[HTML]{67FD9A}0.97} & \multicolumn{1}{c|}{\cellcolor[HTML]{67FD9A}0.97} & \multicolumn{1}{c|}{\cellcolor[HTML]{67FD9A}0.97} & \multicolumn{1}{c|}{\cellcolor[HTML]{67FD9A}0.97} \\ \cline{2-10} 
\multicolumn{1}{|c|}{\multirow{-3}{*}{2.4m}}   & \multicolumn{1}{c|}{10m}                        & \multicolumn{1}{c|}{\cellcolor[HTML]{67FD9A}0.97} & \multicolumn{1}{c|}{\cellcolor[HTML]{67FD9A}0.97} & \multicolumn{1}{c|}{\cellcolor[HTML]{67FD9A}0.97} & \multicolumn{1}{c|}{\cellcolor[HTML]{67FD9A}0.97} & \multicolumn{1}{c|}{\cellcolor[HTML]{67FD9A}0.96} & \multicolumn{1}{c|}{\cellcolor[HTML]{67FD9A}0.97} & \multicolumn{1}{c|}{\cellcolor[HTML]{67FD9A}0.98} & \multicolumn{1}{c|}{\cellcolor[HTML]{67FD9A}0.96} \\ \hline
\end{tabular}%

\hspace{0.1pt}

\begin{tabular}{cccccccccc}
\multicolumn{10}{c}{Night}                                                                                                                                                                                                                                                                                                                                                                                                                                                                                                       \\ \hline
\multicolumn{1}{|c|}{}                         & \multicolumn{1}{c|}{}                           & \multicolumn{8}{c|}{Angle}                                                                                                                                                                                                                                                                                                                                                                                                    \\ \cline{3-10} 
\multicolumn{1}{|c|}{\multirow{-2}{*}{Height}} & \multicolumn{1}{c|}{\multirow{-2}{*}{Distance}} & \multicolumn{1}{c|}{0\degree}                            & \multicolumn{1}{c|}{45\degree}                           & \multicolumn{1}{c|}{90\degree}                           & \multicolumn{1}{c|}{135\degree}                          & \multicolumn{1}{c|}{180\degree}                          & \multicolumn{1}{c|}{225\degree}                          & \multicolumn{1}{c|}{270\degree}                          & \multicolumn{1}{c|}{315\degree}                          \\ \hline
\multicolumn{1}{|c|}{}                         & \multicolumn{1}{c|}{2.5m}                       & \multicolumn{1}{c|}{\cellcolor[HTML]{67FD9A}0.97} & \multicolumn{1}{c|}{\cellcolor[HTML]{67FD9A}0.97} & \multicolumn{1}{c|}{\cellcolor[HTML]{67FD9A}0.96} & \multicolumn{1}{c|}{\cellcolor[HTML]{67FD9A}0.97} & \multicolumn{1}{c|}{\cellcolor[HTML]{67FD9A}0.97} & \multicolumn{1}{c|}{\cellcolor[HTML]{67FD9A}0.97} & \multicolumn{1}{c|}{\cellcolor[HTML]{67FD9A}0.88} & \multicolumn{1}{c|}{\cellcolor[HTML]{67FD9A}0.98} \\ \cline{2-10} 
\multicolumn{1}{|c|}{}                         & \multicolumn{1}{c|}{5m}                         & \multicolumn{1}{c|}{\cellcolor[HTML]{67FD9A}0.95} & \multicolumn{1}{c|}{\cellcolor[HTML]{67FD9A}0.97} & \multicolumn{1}{c|}{\cellcolor[HTML]{67FD9A}0.97} & \multicolumn{1}{c|}{\cellcolor[HTML]{67FD9A}0.97} & \multicolumn{1}{c|}{\cellcolor[HTML]{67FD9A}0.97} & \multicolumn{1}{c|}{\cellcolor[HTML]{67FD9A}0.97} & \multicolumn{1}{c|}{\cellcolor[HTML]{FE0000}0.54} & \multicolumn{1}{c|}{\cellcolor[HTML]{67FD9A}0.97} \\ \cline{2-10} 
\multicolumn{1}{|c|}{\multirow{-3}{*}{0.6m}}   & \multicolumn{1}{c|}{10m}                        & \multicolumn{1}{c|}{\cellcolor[HTML]{67FD9A}0.98} & \multicolumn{1}{c|}{\cellcolor[HTML]{67FD9A}0.97} & \multicolumn{1}{c|}{\cellcolor[HTML]{67FD9A}0.97} & \multicolumn{1}{c|}{\cellcolor[HTML]{67FD9A}0.97} & \multicolumn{1}{c|}{\cellcolor[HTML]{67FD9A}0.96} & \multicolumn{1}{c|}{\cellcolor[HTML]{FFFE65}0.67} & \multicolumn{1}{c|}{\cellcolor[HTML]{67FD9A}0.97} & \multicolumn{1}{c|}{\cellcolor[HTML]{67FD9A}0.98} \\ \hline
\multicolumn{1}{|c|}{}                         & \multicolumn{1}{c|}{2.5m}                       & \multicolumn{1}{c|}{\cellcolor[HTML]{67FD9A}0.97} & \multicolumn{1}{c|}{\cellcolor[HTML]{67FD9A}0.97} & \multicolumn{1}{c|}{\cellcolor[HTML]{67FD9A}0.96} & \multicolumn{1}{c|}{\cellcolor[HTML]{67FD9A}0.97} & \multicolumn{1}{c|}{\cellcolor[HTML]{67FD9A}0.96} & \multicolumn{1}{c|}{\cellcolor[HTML]{67FD9A}0.97} & \multicolumn{1}{c|}{\cellcolor[HTML]{67FD9A}0.96} & \multicolumn{1}{c|}{\cellcolor[HTML]{67FD9A}0.97} \\ \cline{2-10} 
\multicolumn{1}{|c|}{}                         & \multicolumn{1}{c|}{5m}                         & \multicolumn{1}{c|}{\cellcolor[HTML]{67FD9A}0.97} & \multicolumn{1}{c|}{\cellcolor[HTML]{FE0000}0.46} & \multicolumn{1}{c|}{\cellcolor[HTML]{FE0000}0.42} & \multicolumn{1}{c|}{\cellcolor[HTML]{FFFE65}0.61} & \multicolumn{1}{c|}{\cellcolor[HTML]{67FD9A}0.97} & \multicolumn{1}{c|}{\cellcolor[HTML]{FE0000}0.52} & \multicolumn{1}{c|}{\cellcolor[HTML]{FE0000}0.45} & \multicolumn{1}{c|}{\cellcolor[HTML]{FFFE65}0.64} \\ \cline{2-10} 
\multicolumn{1}{|c|}{\multirow{-3}{*}{1.8m}}   & \multicolumn{1}{c|}{10m}                        & \multicolumn{1}{c|}{\cellcolor[HTML]{67FD9A}0.97} & \multicolumn{1}{c|}{\cellcolor[HTML]{FE0000}0.54} & \multicolumn{1}{c|}{\cellcolor[HTML]{FE0000}0.45} & \multicolumn{1}{c|}{\cellcolor[HTML]{67FD9A}0.96} & \multicolumn{1}{c|}{\cellcolor[HTML]{67FD9A}0.96} & \multicolumn{1}{c|}{\cellcolor[HTML]{67FD9A}0.97} & \multicolumn{1}{c|}{\cellcolor[HTML]{67FD9A}0.96} & \multicolumn{1}{c|}{\cellcolor[HTML]{67FD9A}0.98} \\ \hline
\multicolumn{1}{|c|}{}                         & \multicolumn{1}{c|}{2.5m}                       & \multicolumn{1}{c|}{\cellcolor[HTML]{67FD9A}0.97} & \multicolumn{1}{c|}{\cellcolor[HTML]{67FD9A}0.97} & \multicolumn{1}{c|}{\cellcolor[HTML]{FFFE65}0.72} & \multicolumn{1}{c|}{\cellcolor[HTML]{67FD9A}0.97} & \multicolumn{1}{c|}{\cellcolor[HTML]{67FD9A}0.96} & \multicolumn{1}{c|}{\cellcolor[HTML]{67FD9A}0.97} & \multicolumn{1}{c|}{\cellcolor[HTML]{67FD9A}0.96} & \multicolumn{1}{c|}{\cellcolor[HTML]{67FD9A}0.97} \\ \cline{2-10} 
\multicolumn{1}{|c|}{}                         & \multicolumn{1}{c|}{5m}                         & \multicolumn{1}{c|}{\cellcolor[HTML]{67FD9A}0.97} & \multicolumn{1}{c|}{\cellcolor[HTML]{67FD9A}0.96} & \multicolumn{1}{c|}{\cellcolor[HTML]{FFFE65}0.69} & \multicolumn{1}{c|}{\cellcolor[HTML]{67FD9A}0.96} & \multicolumn{1}{c|}{\cellcolor[HTML]{67FD9A}0.98} & \multicolumn{1}{c|}{\cellcolor[HTML]{67FD9A}0.97} & \multicolumn{1}{c|}{\cellcolor[HTML]{FE0000}0.55} & \multicolumn{1}{c|}{\cellcolor[HTML]{67FD9A}0.97} \\ \cline{2-10} 
\multicolumn{1}{|c|}{\multirow{-3}{*}{2.4m}}   & \multicolumn{1}{c|}{10m}                        & \multicolumn{1}{c|}{\cellcolor[HTML]{67FD9A}0.97} & \multicolumn{1}{c|}{\cellcolor[HTML]{67FD9A}0.97} & \multicolumn{1}{c|}{\cellcolor[HTML]{67FD9A}0.98} & \multicolumn{1}{c|}{\cellcolor[HTML]{67FD9A}0.97} & \multicolumn{1}{c|}{\cellcolor[HTML]{67FD9A}0.97} & \multicolumn{1}{c|}{\cellcolor[HTML]{67FD9A}0.97} & \multicolumn{1}{c|}{\cellcolor[HTML]{67FD9A}0.97} & \multicolumn{1}{c|}{\cellcolor[HTML]{67FD9A}0.97} \\ \hline
\end{tabular}%

}

\subsection{Location Effect on Pedestrian Detection
Systems - Additional Object Detector Results}

       \begin{minipage}{0.99\textwidth}
       \centering  
          \includegraphics[width=0.16\textwidth]{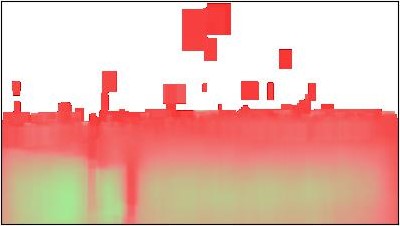} 
     \includegraphics[width=0.16\textwidth]{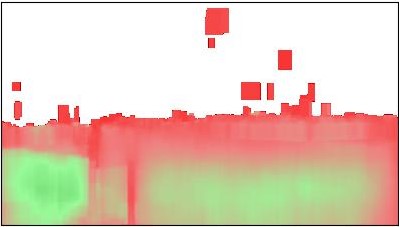}
      \includegraphics[width=0.16\textwidth]{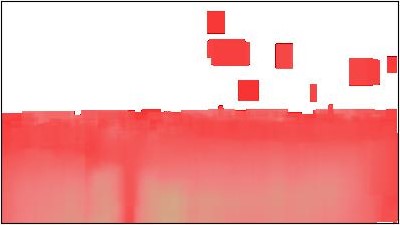} 
     \includegraphics[width=0.16\textwidth]{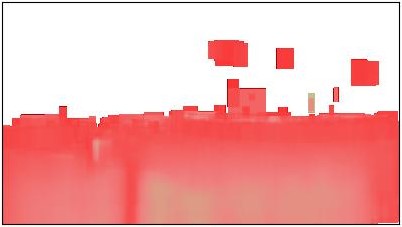}
      \includegraphics[width=0.16\textwidth]{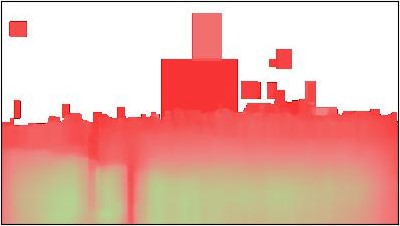}
      \includegraphics[width=0.16\textwidth]{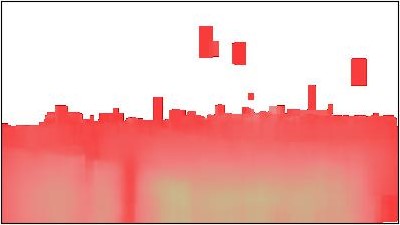}
      \captionof{figure}{The average DiffusionDet confidence of pedestrian detection for each pixel in the observed area of Shibuya Crossing. The six videos, recorded in different lighting conditions (from left to right: daytime, daytime, afternoon, evening, night, and night), are presented separately.} 
      \label{fig:shibuya-results-dnet}
    \end{minipage} 
    \newline
       \begin{minipage}{0.99\textwidth}
       \centering  
          \includegraphics[width=0.16\textwidth]{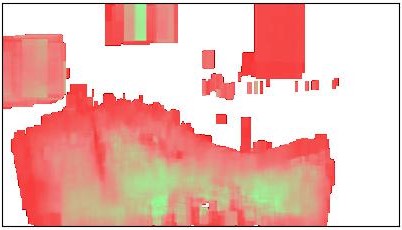} 
     \includegraphics[width=0.16\textwidth]{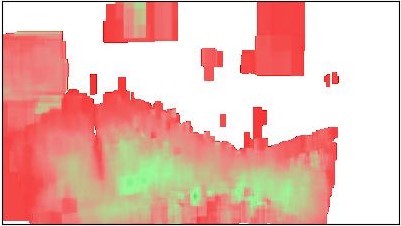}
      \includegraphics[width=0.16\textwidth]{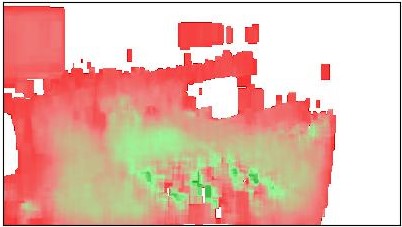} 
     \includegraphics[width=0.16\textwidth]{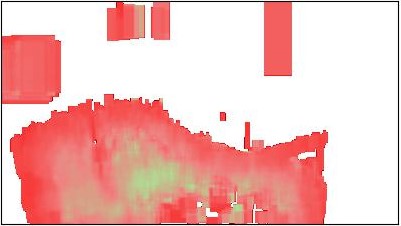}
      \includegraphics[width=0.16\textwidth]{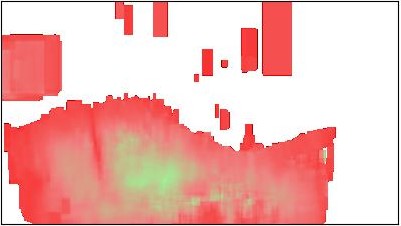}
      \includegraphics[width=0.16\textwidth]{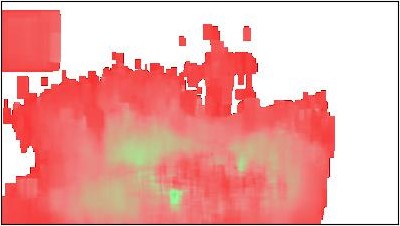}
      \captionof{figure}{The average DiffusionDet confidence of pedestrian detection for each pixel in the observed area of Broadway. The six videos, recorded in different lighting conditions (from left to right: daytime, daytime, daytime, night, night, and night), are presented separately.} 
      \label{fig:new-york-results-dnet}
    \end{minipage} 
    \newline
       \begin{minipage}{0.99\textwidth}
       \centering  
          \includegraphics[width=0.16\textwidth]{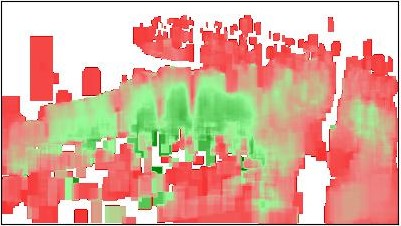} 
     \includegraphics[width=0.16\textwidth]{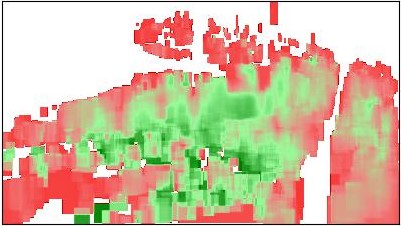}
      \includegraphics[width=0.16\textwidth]{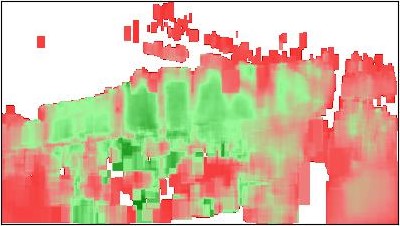} 
     \includegraphics[width=0.16\textwidth]{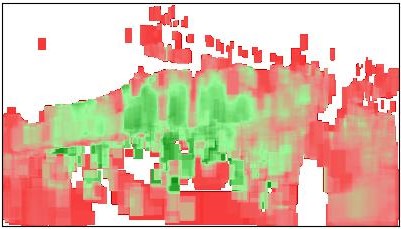}
      \includegraphics[width=0.16\textwidth]{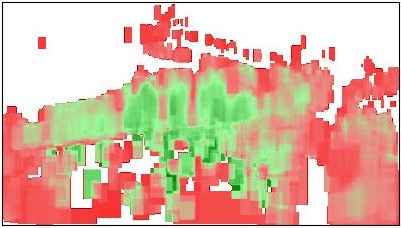}
      \captionof{figure}{The average DiffusionDet confidence of pedestrian detection for each pixel in the observed area of Castro Street. The five videos, recorded in different lighting conditions (from left to right: sunrise, daytime, sunset, night, and night), are presented separately.} 
      \label{fig:castro-results-dnet}
    \end{minipage}

\end{minipage}

\twocolumn

\subsection{Evaluation of L-PET - Additional Object Detector Results}
\label{sec:evaluation-lpea-appendix}

\begin{table}[h!]
\centering
\caption{Average maximum confidence and average per-step confidence of 100 paths generated between 10 random start points and 10 random end points on the observed scene at different times of day. 100 paths were generated for each group: (1) paths generated with L-PET, (2) random direct paths of Manhattan distance from start to end, and (3) random paths. The paths were generated on a scene observed by a DiffusionDet object detector.}
\label{tab:lpea-results-dnet}
\resizebox{1.0\columnwidth}{!}{%
\begin{tabular}{c|c|ccc|ccc|}
\cline{2-8}
                                                         &                                   & \multicolumn{3}{c|}{Max Confidence}                                                                                                           & \multicolumn{3}{c|}{Average Confidence}                                                                                                       \\ \cline{3-8} 
\multirow{-2}{*}{}                                       & \multirow{-2}{*}{Video}           & \multicolumn{1}{c|}{L-PET}                                & \multicolumn{1}{c|}{Man. Dist.}                   & Random                       & \multicolumn{1}{c|}{L-PET}                                & \multicolumn{1}{c|}{Man. Dist.}                   & Random                       \\ \hline
\multicolumn{1}{|c|}{}                                   & \cellcolor[HTML]{C0C0C0}Daytime 1 & \multicolumn{1}{c|}{\cellcolor[HTML]{C0C0C0}\textbf{0.47}} & \multicolumn{1}{c|}{\cellcolor[HTML]{C0C0C0}\textbf{0.47}} & \cellcolor[HTML]{C0C0C0}0.49 & \multicolumn{1}{c|}{\cellcolor[HTML]{C0C0C0}\textbf{0.35}} & \multicolumn{1}{c|}{\cellcolor[HTML]{C0C0C0}0.39} & \cellcolor[HTML]{C0C0C0}0.44 \\ \cline{2-8} 
\multicolumn{1}{|c|}{}                                   & \cellcolor[HTML]{EFEFEF}Daytime 2 & \multicolumn{1}{c|}{\cellcolor[HTML]{EFEFEF}\textbf{0.51}} & \multicolumn{1}{c|}{\cellcolor[HTML]{EFEFEF}0.57} & \cellcolor[HTML]{EFEFEF}0.54 & \multicolumn{1}{c|}{\cellcolor[HTML]{EFEFEF}\textbf{0.37}} & \multicolumn{1}{c|}{\cellcolor[HTML]{EFEFEF}0.41} & \cellcolor[HTML]{EFEFEF}0.47 \\ \cline{2-8} 
\multicolumn{1}{|c|}{}                                   & \cellcolor[HTML]{C0C0C0}Afternoon & \multicolumn{1}{c|}{\cellcolor[HTML]{C0C0C0}\textbf{0.38}} & \multicolumn{1}{c|}{\cellcolor[HTML]{C0C0C0}0.40} & \cellcolor[HTML]{C0C0C0}0.40 & \multicolumn{1}{c|}{\cellcolor[HTML]{C0C0C0}\textbf{0.32}} & \multicolumn{1}{c|}{\cellcolor[HTML]{C0C0C0}0.34} & \cellcolor[HTML]{C0C0C0}0.37 \\ \cline{2-8} 
\multicolumn{1}{|c|}{}                                   & \cellcolor[HTML]{EFEFEF}Evening   & \multicolumn{1}{c|}{\cellcolor[HTML]{EFEFEF}\textbf{0.38}} & \multicolumn{1}{c|}{\cellcolor[HTML]{EFEFEF}0.40} & \cellcolor[HTML]{EFEFEF}0.40 & \multicolumn{1}{c|}{\cellcolor[HTML]{EFEFEF}\textbf{0.32}} & \multicolumn{1}{c|}{\cellcolor[HTML]{EFEFEF}0.34} & \cellcolor[HTML]{EFEFEF}0.36 \\ \cline{2-8} 
\multicolumn{1}{|c|}{}                                   & \cellcolor[HTML]{C0C0C0}Night 1   & \multicolumn{1}{c|}{\cellcolor[HTML]{C0C0C0}\textbf{0.49}} & \multicolumn{1}{c|}{\cellcolor[HTML]{C0C0C0}0.51} & \cellcolor[HTML]{C0C0C0}0.51 & \multicolumn{1}{c|}{\cellcolor[HTML]{C0C0C0}\textbf{0.36}} & \multicolumn{1}{c|}{\cellcolor[HTML]{C0C0C0}0.41} & \cellcolor[HTML]{C0C0C0}0.44 \\ \cline{2-8} 
\multicolumn{1}{|c|}{\multirow{-6}{*}{\rotatebox[origin=c]{90}{Shibuya}}} & \cellcolor[HTML]{EFEFEF}Night 2   & \multicolumn{1}{c|}{\cellcolor[HTML]{EFEFEF}\textbf{0.42}} & \multicolumn{1}{c|}{\cellcolor[HTML]{EFEFEF}0.50} & \cellcolor[HTML]{EFEFEF}0.44 & \multicolumn{1}{c|}{\cellcolor[HTML]{EFEFEF}\textbf{0.32}} & \multicolumn{1}{c|}{\cellcolor[HTML]{EFEFEF}0.35} & \cellcolor[HTML]{EFEFEF}0.40 \\ \hline
\multicolumn{1}{|c|}{}                                   & \cellcolor[HTML]{C0C0C0}Daytime 1 & \multicolumn{1}{c|}{\cellcolor[HTML]{C0C0C0}\textbf{0.47}} & \multicolumn{1}{c|}{\cellcolor[HTML]{C0C0C0}0.51} & \cellcolor[HTML]{C0C0C0}0.52 & \multicolumn{1}{c|}{\cellcolor[HTML]{C0C0C0}\textbf{0.34}} & \multicolumn{1}{c|}{\cellcolor[HTML]{C0C0C0}0.40} & \cellcolor[HTML]{C0C0C0}0.44 \\ \cline{2-8} 
\multicolumn{1}{|c|}{}                                   & \cellcolor[HTML]{EFEFEF}Daytime 2 & \multicolumn{1}{c|}{\cellcolor[HTML]{EFEFEF}\textbf{0.46}} & \multicolumn{1}{c|}{\cellcolor[HTML]{EFEFEF}0.55} & \cellcolor[HTML]{EFEFEF}0.55 & \multicolumn{1}{c|}{\cellcolor[HTML]{EFEFEF}\textbf{0.35}} & \multicolumn{1}{c|}{\cellcolor[HTML]{EFEFEF}0.40} & \cellcolor[HTML]{EFEFEF}0.44 \\ \cline{2-8} 
\multicolumn{1}{|c|}{}                                   & \cellcolor[HTML]{C0C0C0}Daytime 3 & \multicolumn{1}{c|}{\cellcolor[HTML]{C0C0C0}\textbf{0.50}} & \multicolumn{1}{c|}{\cellcolor[HTML]{C0C0C0}0.57} & \cellcolor[HTML]{C0C0C0}0.62 & \multicolumn{1}{c|}{\cellcolor[HTML]{C0C0C0}\textbf{0.39}} & \multicolumn{1}{c|}{\cellcolor[HTML]{C0C0C0}0.46} & \cellcolor[HTML]{C0C0C0}0.49 \\ \cline{2-8} 
\multicolumn{1}{|c|}{}                                   & \cellcolor[HTML]{EFEFEF}Night 1   & \multicolumn{1}{c|}{\cellcolor[HTML]{EFEFEF}\textbf{0.40}} & \multicolumn{1}{c|}{\cellcolor[HTML]{EFEFEF}0.51} & \cellcolor[HTML]{EFEFEF}0.48 & \multicolumn{1}{c|}{\cellcolor[HTML]{EFEFEF}\textbf{0.33}} & \multicolumn{1}{c|}{\cellcolor[HTML]{EFEFEF}0.37} & \cellcolor[HTML]{EFEFEF}0.39 \\ \cline{2-8} 
\multicolumn{1}{|c|}{}                                   & \cellcolor[HTML]{C0C0C0}Night 2   & \multicolumn{1}{c|}{\cellcolor[HTML]{C0C0C0}\textbf{0.39}} & \multicolumn{1}{c|}{\cellcolor[HTML]{C0C0C0}0.44} & \cellcolor[HTML]{C0C0C0}0.45 & \multicolumn{1}{c|}{\cellcolor[HTML]{C0C0C0}\textbf{0.31}} & \multicolumn{1}{c|}{\cellcolor[HTML]{C0C0C0}0.34} & \cellcolor[HTML]{C0C0C0}0.38 \\ \cline{2-8} 
\multicolumn{1}{|c|}{\multirow{-6}{*}{\rotatebox[origin=c]{90}{Broadway}}}         & \cellcolor[HTML]{EFEFEF}Night 3   & \multicolumn{1}{c|}{\cellcolor[HTML]{EFEFEF}\textbf{0.43}} & \multicolumn{1}{c|}{\cellcolor[HTML]{EFEFEF}0.47} & \cellcolor[HTML]{EFEFEF}0.48 & \multicolumn{1}{c|}{\cellcolor[HTML]{EFEFEF}\textbf{0.34}} & \multicolumn{1}{c|}{\cellcolor[HTML]{EFEFEF}0.40} & \cellcolor[HTML]{EFEFEF}0.41 \\ \hline
\multicolumn{1}{|c|}{}                                   & \cellcolor[HTML]{C0C0C0}Sunrise   & \multicolumn{1}{c|}{\cellcolor[HTML]{C0C0C0}\textbf{0.56}} & \multicolumn{1}{c|}{\cellcolor[HTML]{C0C0C0}0.74} & \cellcolor[HTML]{C0C0C0}0.72 & \multicolumn{1}{c|}{\cellcolor[HTML]{C0C0C0}\textbf{0.39}} & \multicolumn{1}{c|}{\cellcolor[HTML]{C0C0C0}0.48} & \cellcolor[HTML]{C0C0C0}0.52 \\ \cline{2-8} 
\multicolumn{1}{|c|}{}                                   & \cellcolor[HTML]{EFEFEF}Daytime   & \multicolumn{1}{c|}{\cellcolor[HTML]{EFEFEF}\textbf{0.62}} & \multicolumn{1}{c|}{\cellcolor[HTML]{EFEFEF}0.78} & \cellcolor[HTML]{EFEFEF}0.76 & \multicolumn{1}{c|}{\cellcolor[HTML]{EFEFEF}\textbf{0.42}} & \multicolumn{1}{c|}{\cellcolor[HTML]{EFEFEF}0.52} & \cellcolor[HTML]{EFEFEF}0.55 \\ \cline{2-8} 
\multicolumn{1}{|c|}{}                                   & \cellcolor[HTML]{C0C0C0}Sunset    & \multicolumn{1}{c|}{\cellcolor[HTML]{C0C0C0}\textbf{0.59}} & \multicolumn{1}{c|}{\cellcolor[HTML]{C0C0C0}0.70} & \cellcolor[HTML]{C0C0C0}0.69 & \multicolumn{1}{c|}{\cellcolor[HTML]{C0C0C0}\textbf{0.42}} & \multicolumn{1}{c|}{\cellcolor[HTML]{C0C0C0}0.47} & \cellcolor[HTML]{C0C0C0}0.51 \\ \cline{2-8} 
\multicolumn{1}{|c|}{}                                   & \cellcolor[HTML]{EFEFEF}Night 1   & \multicolumn{1}{c|}{\cellcolor[HTML]{EFEFEF}\textbf{0.58}} & \multicolumn{1}{c|}{\cellcolor[HTML]{EFEFEF}0.74} & \cellcolor[HTML]{EFEFEF}0.74 & \multicolumn{1}{c|}{\cellcolor[HTML]{EFEFEF}\textbf{0.42}} & \multicolumn{1}{c|}{\cellcolor[HTML]{EFEFEF}0.51} & \cellcolor[HTML]{EFEFEF}0.52 \\ \cline{2-8} 
\multicolumn{1}{|c|}{\multirow{-5}{*}{\rotatebox[origin=c]{90}{Castro}}}    & \cellcolor[HTML]{C0C0C0}Night 2   & \multicolumn{1}{c|}{\cellcolor[HTML]{C0C0C0}\textbf{0.55}} & \multicolumn{1}{c|}{\cellcolor[HTML]{C0C0C0}0.72} & \cellcolor[HTML]{C0C0C0}0.72 & \multicolumn{1}{c|}{\cellcolor[HTML]{C0C0C0}\textbf{0.39}} & \multicolumn{1}{c|}{\cellcolor[HTML]{C0C0C0}0.48} & \cellcolor[HTML]{C0C0C0}0.49 \\ \hline
\multicolumn{1}{l|}{}                                    & Average                           & \multicolumn{1}{c|}{\textbf{0.48}}                                  & \multicolumn{1}{c|}{0.56}                         & 0.56                         & \multicolumn{1}{c|}{\textbf{0.36}}                                  & \multicolumn{1}{c|}{0.42}                         & 0.45                         \\ \cline{2-8}
\end{tabular}%
}
\end{table}

\newpage
\subsection{Evaluation of L-BAT - Additional Object Detector Results}

\begin{table}[h!]
\centering
\caption{Average maximum confidence and average per-step confidence of 100 paths generated between 10 random start points and 10 random end points on the observed scene at different times of day. 100 paths were generated for each group: (1) paths generated with L-PET, (2) random direct paths of Manhattan distance from start to end, and (3) random paths. The paths were generated on a scene observed by a DiffusionDet object detector enhanced with the L-BAT countermeasure.}
\label{tab:lbat-lpea-results-faster}
\resizebox{1.0\columnwidth}{!}{%
\begin{tabular}{c|c|ccc|ccc|}
\cline{2-8}
                                                         &                                   & \multicolumn{3}{c|}{Max Confidence}                                                                                                           & \multicolumn{3}{c|}{Average Confidence}                                                                                                       \\ \cline{3-8} 
\multirow{-2}{*}{}                                       & \multirow{-2}{*}{Video}           & \multicolumn{1}{c|}{L-PET}                                & \multicolumn{1}{c|}{Man. Dist.}                   & Random                       & \multicolumn{1}{c|}{L-PET}                                & \multicolumn{1}{c|}{Man. Dist.}                   & Random                       \\ \hline
\multicolumn{1}{|c|}{}                                   & \cellcolor[HTML]{C0C0C0}Daytime 1 & \multicolumn{1}{c|}{\cellcolor[HTML]{C0C0C0}\textbf{0.46}} & \multicolumn{1}{c|}{\cellcolor[HTML]{C0C0C0}0.49} & \cellcolor[HTML]{C0C0C0}0.50 & \multicolumn{1}{c|}{\cellcolor[HTML]{C0C0C0}\textbf{0.35}} & \multicolumn{1}{c|}{\cellcolor[HTML]{C0C0C0}0.38} & \cellcolor[HTML]{C0C0C0}0.43 \\ \cline{2-8} 
\multicolumn{1}{|c|}{}                                   & \cellcolor[HTML]{EFEFEF}Daytime 2 & \multicolumn{1}{c|}{\cellcolor[HTML]{EFEFEF}\textbf{0.54}} & \multicolumn{1}{c|}{\cellcolor[HTML]{EFEFEF}0.57} & \cellcolor[HTML]{EFEFEF}0.59 & \multicolumn{1}{c|}{\cellcolor[HTML]{EFEFEF}\textbf{0.38}} & \multicolumn{1}{c|}{\cellcolor[HTML]{EFEFEF}0.41} & \cellcolor[HTML]{EFEFEF}0.48 \\ \cline{2-8} 
\multicolumn{1}{|c|}{}                                   & \cellcolor[HTML]{C0C0C0}Afternoon & \multicolumn{1}{c|}{\cellcolor[HTML]{C0C0C0}\textbf{0.39}} & \multicolumn{1}{c|}{\cellcolor[HTML]{C0C0C0}0.40} & \cellcolor[HTML]{C0C0C0}0.40 & \multicolumn{1}{c|}{\cellcolor[HTML]{C0C0C0}\textbf{0.32}} & \multicolumn{1}{c|}{\cellcolor[HTML]{C0C0C0}0.34} & \cellcolor[HTML]{C0C0C0}0.36 \\ \cline{2-8} 
\multicolumn{1}{|c|}{}                                   & \cellcolor[HTML]{EFEFEF}Evening   & \multicolumn{1}{c|}{\cellcolor[HTML]{EFEFEF}\textbf{0.41}} & \multicolumn{1}{c|}{\cellcolor[HTML]{EFEFEF}0.42} & \cellcolor[HTML]{EFEFEF}0.43 & \multicolumn{1}{c|}{\cellcolor[HTML]{EFEFEF}\textbf{0.33}} & \multicolumn{1}{c|}{\cellcolor[HTML]{EFEFEF}0.36} & \cellcolor[HTML]{EFEFEF}0.39 \\ \cline{2-8} 
\multicolumn{1}{|c|}{}                                   & \cellcolor[HTML]{C0C0C0}Night 1   & \multicolumn{1}{c|}{\cellcolor[HTML]{C0C0C0}\textbf{0.45}} & \multicolumn{1}{c|}{\cellcolor[HTML]{C0C0C0}0.47} & \cellcolor[HTML]{C0C0C0}0.49 & \multicolumn{1}{c|}{\cellcolor[HTML]{C0C0C0}\textbf{0.34}} & \multicolumn{1}{c|}{\cellcolor[HTML]{C0C0C0}0.38} & \cellcolor[HTML]{C0C0C0}0.43 \\ \cline{2-8} 
\multicolumn{1}{|c|}{\multirow{-6}{*}{\rotatebox[origin=c]{90}{Shibuya}}} & \cellcolor[HTML]{EFEFEF}Night 2   & \multicolumn{1}{c|}{\cellcolor[HTML]{EFEFEF}\textbf{0.41}} & \multicolumn{1}{c|}{\cellcolor[HTML]{EFEFEF}0.44} & \cellcolor[HTML]{EFEFEF}0.43 & \multicolumn{1}{c|}{\cellcolor[HTML]{EFEFEF}\textbf{0.33}} & \multicolumn{1}{c|}{\cellcolor[HTML]{EFEFEF}0.35} & \cellcolor[HTML]{EFEFEF}0.39 \\ \hline
\multicolumn{1}{|c|}{}                                   & \cellcolor[HTML]{C0C0C0}Daytime 1 & \multicolumn{1}{c|}{\cellcolor[HTML]{C0C0C0}\textbf{0.46}} & \multicolumn{1}{c|}{\cellcolor[HTML]{C0C0C0}0.50} & \cellcolor[HTML]{C0C0C0}0.52 & \multicolumn{1}{c|}{\cellcolor[HTML]{C0C0C0}\textbf{0.33}} & \multicolumn{1}{c|}{\cellcolor[HTML]{C0C0C0}0.39} & \cellcolor[HTML]{C0C0C0}0.43 \\ \cline{2-8} 
\multicolumn{1}{|c|}{}                                   & \cellcolor[HTML]{EFEFEF}Daytime 2 & \multicolumn{1}{c|}{\cellcolor[HTML]{EFEFEF}\textbf{0.51}} & \multicolumn{1}{c|}{\cellcolor[HTML]{EFEFEF}0.55} & \cellcolor[HTML]{EFEFEF}0.58 & \multicolumn{1}{c|}{\cellcolor[HTML]{EFEFEF}\textbf{0.36}} & \multicolumn{1}{c|}{\cellcolor[HTML]{EFEFEF}0.42} & \cellcolor[HTML]{EFEFEF}0.46 \\ \cline{2-8} 
\multicolumn{1}{|c|}{}                                   & \cellcolor[HTML]{C0C0C0}Daytime 3 & \multicolumn{1}{c|}{\cellcolor[HTML]{C0C0C0}\textbf{0.54}} & \multicolumn{1}{c|}{\cellcolor[HTML]{C0C0C0}0.58} & \cellcolor[HTML]{C0C0C0}0.61 & \multicolumn{1}{c|}{\cellcolor[HTML]{C0C0C0}\textbf{0.38}} & \multicolumn{1}{c|}{\cellcolor[HTML]{C0C0C0}0.47} & \cellcolor[HTML]{C0C0C0}0.49 \\ \cline{2-8} 
\multicolumn{1}{|c|}{}                                   & \cellcolor[HTML]{EFEFEF}Night 1   & \multicolumn{1}{c|}{\cellcolor[HTML]{EFEFEF}\textbf{0.44}} & \multicolumn{1}{c|}{\cellcolor[HTML]{EFEFEF}0.55} & \cellcolor[HTML]{EFEFEF}0.50 & \multicolumn{1}{c|}{\cellcolor[HTML]{EFEFEF}\textbf{0.32}} & \multicolumn{1}{c|}{\cellcolor[HTML]{EFEFEF}0.38} & \cellcolor[HTML]{EFEFEF}0.41 \\ \cline{2-8} 
\multicolumn{1}{|c|}{}                                   & \cellcolor[HTML]{C0C0C0}Night 2   & \multicolumn{1}{c|}{\cellcolor[HTML]{C0C0C0}\textbf{0.42}} & \multicolumn{1}{c|}{\cellcolor[HTML]{C0C0C0}0.48} & \cellcolor[HTML]{C0C0C0}0.49 & \multicolumn{1}{c|}{\cellcolor[HTML]{C0C0C0}\textbf{0.32}} & \multicolumn{1}{c|}{\cellcolor[HTML]{C0C0C0}0.38} & \cellcolor[HTML]{C0C0C0}0.41 \\ \cline{2-8} 
\multicolumn{1}{|c|}{\multirow{-6}{*}{\rotatebox[origin=c]{90}{Broadway}}}         & \cellcolor[HTML]{EFEFEF}Night 3   & \multicolumn{1}{c|}{\cellcolor[HTML]{EFEFEF}\textbf{0.45}} & \multicolumn{1}{c|}{\cellcolor[HTML]{EFEFEF}0.51} & \cellcolor[HTML]{EFEFEF}0.52 & \multicolumn{1}{c|}{\cellcolor[HTML]{EFEFEF}\textbf{0.35}} & \multicolumn{1}{c|}{\cellcolor[HTML]{EFEFEF}0.40} & \cellcolor[HTML]{EFEFEF}0.41 \\ \hline
\multicolumn{1}{|c|}{}                                   & \cellcolor[HTML]{C0C0C0}Sunrise   & \multicolumn{1}{c|}{\cellcolor[HTML]{C0C0C0}\textbf{0.57}} & \multicolumn{1}{c|}{\cellcolor[HTML]{C0C0C0}0.77} & \cellcolor[HTML]{C0C0C0}0.76 & \multicolumn{1}{c|}{\cellcolor[HTML]{C0C0C0}\textbf{0.41}} & \multicolumn{1}{c|}{\cellcolor[HTML]{C0C0C0}0.54} & \cellcolor[HTML]{C0C0C0}0.54 \\ \cline{2-8} 
\multicolumn{1}{|c|}{}                                   & \cellcolor[HTML]{EFEFEF}Daytime   & \multicolumn{1}{c|}{\cellcolor[HTML]{EFEFEF}\textbf{0.72}} & \multicolumn{1}{c|}{\cellcolor[HTML]{EFEFEF}0.82} & \cellcolor[HTML]{EFEFEF}0.81 & \multicolumn{1}{c|}{\cellcolor[HTML]{EFEFEF}\textbf{0.50}} & \multicolumn{1}{c|}{\cellcolor[HTML]{EFEFEF}0.62} & \cellcolor[HTML]{EFEFEF}0.64 \\ \cline{2-8} 
\multicolumn{1}{|c|}{}                                   & \cellcolor[HTML]{C0C0C0}Sunset    & \multicolumn{1}{c|}{\cellcolor[HTML]{C0C0C0}\textbf{0.68}} & \multicolumn{1}{c|}{\cellcolor[HTML]{C0C0C0}0.76} & \cellcolor[HTML]{C0C0C0}0.77 & \multicolumn{1}{c|}{\cellcolor[HTML]{C0C0C0}\textbf{0.50}} & \multicolumn{1}{c|}{\cellcolor[HTML]{C0C0C0}0.56} & \cellcolor[HTML]{C0C0C0}0.57 \\ \cline{2-8} 
\multicolumn{1}{|c|}{}                                   & \cellcolor[HTML]{EFEFEF}Night 1   & \multicolumn{1}{c|}{\cellcolor[HTML]{EFEFEF}\textbf{0.59}} & \multicolumn{1}{c|}{\cellcolor[HTML]{EFEFEF}0.72} & \cellcolor[HTML]{EFEFEF}0.73 & \multicolumn{1}{c|}{\cellcolor[HTML]{EFEFEF}\textbf{0.44}} & \multicolumn{1}{c|}{\cellcolor[HTML]{EFEFEF}0.49} & \cellcolor[HTML]{EFEFEF}0.52 \\ \cline{2-8} 
\multicolumn{1}{|c|}{\multirow{-5}{*}{\rotatebox[origin=c]{90}{Castro}}}    & \cellcolor[HTML]{C0C0C0}Night 2   & \multicolumn{1}{c|}{\cellcolor[HTML]{C0C0C0}\textbf{0.58}} & \multicolumn{1}{c|}{\cellcolor[HTML]{C0C0C0}0.73} & \cellcolor[HTML]{C0C0C0}0.73 & \multicolumn{1}{c|}{\cellcolor[HTML]{C0C0C0}\textbf{0.40}} & \multicolumn{1}{c|}{\cellcolor[HTML]{C0C0C0}0.49} & \cellcolor[HTML]{C0C0C0}0.49 \\ \hline
\multicolumn{1}{l|}{}                                    & Average                           & \multicolumn{1}{c|}{\textbf{0.51}}                                  & \multicolumn{1}{c|}{0.57}                         & 0.58                         & \multicolumn{1}{c|}{\textbf{0.37}}                                  & \multicolumn{1}{c|}{0.43}                         & 0.46                         \\ \cline{2-8}
\end{tabular}%
}
\end{table}

\begin{table}[h!]
\centering
\caption{AUC score, true positive rate (TPR), false positive rate (FPR), and average true positive confidence of (1) an untouched DiffusionDet model,
and (2) a DiffusionDet model utilizing L-BAT.}
\label{tab:evaluation-results-lbat-performance-dnet}
\resizebox{1.0\columnwidth}{!}{%
\begin{tabular}{c|c|cccc|cccc|}
\cline{2-10}
                                                 &                                   & \multicolumn{4}{c|}{Original Model}                                                                                                                                                               & \multicolumn{4}{c|}{Original Model with L-BAT}                                                                                                                                                                      \\ \cline{3-10} 
\multirow{-2}{*}{}                               & \multirow{-2}{*}{Video}           & \multicolumn{1}{c|}{AUC}                          & \multicolumn{1}{c|}{TPR}                          & \multicolumn{1}{c|}{FPR}                                   & \makecell{Avg. TP \\ Confidence}           & \multicolumn{1}{c|}{AUC}                                   & \multicolumn{1}{c|}{TPR}                                   & \multicolumn{1}{c|}{FPR}                          & \makecell{Avg. TP \\ Confidence}                    \\ \hline
\multicolumn{1}{|c|}{}                           & \cellcolor[HTML]{C0C0C0}Daytime 1 & \multicolumn{1}{c|}{\cellcolor[HTML]{C0C0C0}0.68} & \multicolumn{1}{c|}{\cellcolor[HTML]{C0C0C0}0.56} & \multicolumn{1}{c|}{\cellcolor[HTML]{C0C0C0}\textbf{0.17}} & \cellcolor[HTML]{C0C0C0}0.39 & \multicolumn{1}{c|}{\cellcolor[HTML]{C0C0C0}\textbf{0.85}} & \multicolumn{1}{c|}{\cellcolor[HTML]{C0C0C0}\textbf{0.98}} & \multicolumn{1}{c|}{\cellcolor[HTML]{C0C0C0}0.29} & \cellcolor[HTML]{C0C0C0}\textbf{0.77} \\ \cline{2-10} 
\multicolumn{1}{|c|}{}                           & \cellcolor[HTML]{EFEFEF}Daytime 2 & \multicolumn{1}{c|}{\cellcolor[HTML]{EFEFEF}0.70} & \multicolumn{1}{c|}{\cellcolor[HTML]{EFEFEF}0.56} & \multicolumn{1}{c|}{\cellcolor[HTML]{EFEFEF}\textbf{0.14}} & \cellcolor[HTML]{EFEFEF}0.40 & \multicolumn{1}{c|}{\cellcolor[HTML]{EFEFEF}\textbf{0.86}} & \multicolumn{1}{c|}{\cellcolor[HTML]{EFEFEF}\textbf{0.97}} & \multicolumn{1}{c|}{\cellcolor[HTML]{EFEFEF}0.24} & \cellcolor[HTML]{EFEFEF}\textbf{0.75} \\ \cline{2-10} 
\multicolumn{1}{|c|}{}                           & \cellcolor[HTML]{C0C0C0}Afternoon & \multicolumn{1}{c|}{\cellcolor[HTML]{C0C0C0}0.66} & \multicolumn{1}{c|}{\cellcolor[HTML]{C0C0C0}0.55} & \multicolumn{1}{c|}{\cellcolor[HTML]{C0C0C0}\textbf{0.22}} & \cellcolor[HTML]{C0C0C0}0.39 & \multicolumn{1}{c|}{\cellcolor[HTML]{C0C0C0}\textbf{0.80}} & \multicolumn{1}{c|}{\cellcolor[HTML]{C0C0C0}\textbf{0.98}} & \multicolumn{1}{c|}{\cellcolor[HTML]{C0C0C0}0.45} & \cellcolor[HTML]{C0C0C0}\textbf{0.78} \\ \cline{2-10} 
\multicolumn{1}{|c|}{}                           & \cellcolor[HTML]{EFEFEF}Evening   & \multicolumn{1}{c|}{\cellcolor[HTML]{EFEFEF}0.70} & \multicolumn{1}{c|}{\cellcolor[HTML]{EFEFEF}0.63} & \multicolumn{1}{c|}{\cellcolor[HTML]{EFEFEF}\textbf{0.22}} & \cellcolor[HTML]{EFEFEF}0.39 & \multicolumn{1}{c|}{\cellcolor[HTML]{EFEFEF}\textbf{0.81}} & \multicolumn{1}{c|}{\cellcolor[HTML]{EFEFEF}\textbf{0.98}} & \multicolumn{1}{c|}{\cellcolor[HTML]{EFEFEF}0.44} & \cellcolor[HTML]{EFEFEF}\textbf{0.82} \\ \cline{2-10} 
\multicolumn{1}{|c|}{}                           & \cellcolor[HTML]{C0C0C0}Night 1   & \multicolumn{1}{c|}{\cellcolor[HTML]{C0C0C0}0.67} & \multicolumn{1}{c|}{\cellcolor[HTML]{C0C0C0}0.54} & \multicolumn{1}{c|}{\cellcolor[HTML]{C0C0C0}\textbf{0.18}} & \cellcolor[HTML]{C0C0C0}0.40 & \multicolumn{1}{c|}{\cellcolor[HTML]{C0C0C0}\textbf{0.83}} & \multicolumn{1}{c|}{\cellcolor[HTML]{C0C0C0}\textbf{0.98}} & \multicolumn{1}{c|}{\cellcolor[HTML]{C0C0C0}0.34} & \cellcolor[HTML]{C0C0C0}\textbf{0.76} \\ \cline{2-10} 
\multicolumn{1}{|c|}{\multirow{-6}{*}{\rotatebox[origin=c]{90}{Shibuya}}}  & \cellcolor[HTML]{EFEFEF}Night 2   & \multicolumn{1}{c|}{\cellcolor[HTML]{EFEFEF}0.70} & \multicolumn{1}{c|}{\cellcolor[HTML]{EFEFEF}0.62} & \multicolumn{1}{c|}{\cellcolor[HTML]{EFEFEF}\textbf{0.21}} & \cellcolor[HTML]{EFEFEF}0.39 & \multicolumn{1}{c|}{\cellcolor[HTML]{EFEFEF}\textbf{0.82}} & \multicolumn{1}{c|}{\cellcolor[HTML]{EFEFEF}\textbf{0.98}} & \multicolumn{1}{c|}{\cellcolor[HTML]{EFEFEF}0.39} & \cellcolor[HTML]{EFEFEF}\textbf{0.80} \\ \hline
\multicolumn{1}{|c|}{}                           & \cellcolor[HTML]{C0C0C0}Daytime 1 & \multicolumn{1}{c|}{\cellcolor[HTML]{C0C0C0}0.76} & \multicolumn{1}{c|}{\cellcolor[HTML]{C0C0C0}0.64} & \multicolumn{1}{c|}{\cellcolor[HTML]{C0C0C0}\textbf{0.10}} & \cellcolor[HTML]{C0C0C0}0.38 & \multicolumn{1}{c|}{\cellcolor[HTML]{C0C0C0}\textbf{0.88}} & \multicolumn{1}{c|}{\cellcolor[HTML]{C0C0C0}\textbf{0.93}} & \multicolumn{1}{c|}{\cellcolor[HTML]{C0C0C0}0.20} & \cellcolor[HTML]{C0C0C0}\textbf{0.80} \\ \cline{2-10} 
\multicolumn{1}{|c|}{}                           & \cellcolor[HTML]{EFEFEF}Daytime 2 & \multicolumn{1}{c|}{\cellcolor[HTML]{EFEFEF}0.82} & \multicolumn{1}{c|}{\cellcolor[HTML]{EFEFEF}0.75} & \multicolumn{1}{c|}{\cellcolor[HTML]{EFEFEF}\textbf{0.10}} & \cellcolor[HTML]{EFEFEF}0.42 & \multicolumn{1}{c|}{\cellcolor[HTML]{EFEFEF}\textbf{0.90}} & \multicolumn{1}{c|}{\cellcolor[HTML]{EFEFEF}\textbf{0.96}} & \multicolumn{1}{c|}{\cellcolor[HTML]{EFEFEF}0.22} & \cellcolor[HTML]{EFEFEF}\textbf{0.83} \\ \cline{2-10} 
\multicolumn{1}{|c|}{}                           & \cellcolor[HTML]{C0C0C0}Daytime 3 & \multicolumn{1}{c|}{\cellcolor[HTML]{C0C0C0}0.80} & \multicolumn{1}{c|}{\cellcolor[HTML]{C0C0C0}0.69} & \multicolumn{1}{c|}{\cellcolor[HTML]{C0C0C0}\textbf{0.09}} & \cellcolor[HTML]{C0C0C0}0.40 & \multicolumn{1}{c|}{\cellcolor[HTML]{C0C0C0}\textbf{0.90}} & \multicolumn{1}{c|}{\cellcolor[HTML]{C0C0C0}\textbf{0.96}} & \multicolumn{1}{c|}{\cellcolor[HTML]{C0C0C0}0.22} & \cellcolor[HTML]{C0C0C0}\textbf{0.82} \\ \cline{2-10} 
\multicolumn{1}{|c|}{}                           & \cellcolor[HTML]{EFEFEF}Night 1   & \multicolumn{1}{c|}{\cellcolor[HTML]{EFEFEF}0.75} & \multicolumn{1}{c|}{\cellcolor[HTML]{EFEFEF}0.61} & \multicolumn{1}{c|}{\cellcolor[HTML]{EFEFEF}\textbf{0.10}} & \cellcolor[HTML]{EFEFEF}0.36 & \multicolumn{1}{c|}{\cellcolor[HTML]{EFEFEF}\textbf{0.87}} & \multicolumn{1}{c|}{\cellcolor[HTML]{EFEFEF}\textbf{0.92}} & \multicolumn{1}{c|}{\cellcolor[HTML]{EFEFEF}0.19} & \cellcolor[HTML]{EFEFEF}\textbf{0.81} \\ \cline{2-10} 
\multicolumn{1}{|c|}{}                           & \cellcolor[HTML]{C0C0C0}Night 2   & \multicolumn{1}{c|}{\cellcolor[HTML]{C0C0C0}0.80} & \multicolumn{1}{c|}{\cellcolor[HTML]{C0C0C0}0.70} & \multicolumn{1}{c|}{\cellcolor[HTML]{C0C0C0}\textbf{0.10}} & \cellcolor[HTML]{C0C0C0}0.38 & \multicolumn{1}{c|}{\cellcolor[HTML]{C0C0C0}\textbf{0.91}} & \multicolumn{1}{c|}{\cellcolor[HTML]{C0C0C0}\textbf{0.96}} & \multicolumn{1}{c|}{\cellcolor[HTML]{C0C0C0}0.20} & \cellcolor[HTML]{C0C0C0}\textbf{0.83} \\ \cline{2-10} 
\multicolumn{1}{|c|}{\multirow{-6}{*}{\rotatebox[origin=c]{90}{Broadway}}} & \cellcolor[HTML]{EFEFEF}Night 3   & \multicolumn{1}{c|}{\cellcolor[HTML]{EFEFEF}0.87} & \multicolumn{1}{c|}{\cellcolor[HTML]{EFEFEF}0.83} & \multicolumn{1}{c|}{\cellcolor[HTML]{EFEFEF}\textbf{0.10}} & \cellcolor[HTML]{EFEFEF}0.43 & \multicolumn{1}{c|}{\cellcolor[HTML]{EFEFEF}\textbf{0.92}} & \multicolumn{1}{c|}{\cellcolor[HTML]{EFEFEF}\textbf{0.98}} & \multicolumn{1}{c|}{\cellcolor[HTML]{EFEFEF}0.27} & \cellcolor[HTML]{EFEFEF}\textbf{0.89} \\ \hline
\multicolumn{1}{|c|}{}                           & \cellcolor[HTML]{C0C0C0}Sunrise   & \multicolumn{1}{c|}{\cellcolor[HTML]{C0C0C0}0.84} & \multicolumn{1}{c|}{\cellcolor[HTML]{C0C0C0}0.75} & \multicolumn{1}{c|}{\cellcolor[HTML]{C0C0C0}\textbf{0.06}} & \cellcolor[HTML]{C0C0C0}0.48 & \multicolumn{1}{c|}{\cellcolor[HTML]{C0C0C0}\textbf{0.92}} & \multicolumn{1}{c|}{\cellcolor[HTML]{C0C0C0}\textbf{0.94}} & \multicolumn{1}{c|}{\cellcolor[HTML]{C0C0C0}0.15} & \cellcolor[HTML]{C0C0C0}\textbf{0.81} \\ \cline{2-10} 
\multicolumn{1}{|c|}{}                           & \cellcolor[HTML]{EFEFEF}Daytime   & \multicolumn{1}{c|}{\cellcolor[HTML]{EFEFEF}0.81} & \multicolumn{1}{c|}{\cellcolor[HTML]{EFEFEF}0.67} & \multicolumn{1}{c|}{\cellcolor[HTML]{EFEFEF}\textbf{0.04}} & \cellcolor[HTML]{EFEFEF}0.50 & \multicolumn{1}{c|}{\cellcolor[HTML]{EFEFEF}\textbf{0.91}} & \multicolumn{1}{c|}{\cellcolor[HTML]{EFEFEF}\textbf{0.91}} & \multicolumn{1}{c|}{\cellcolor[HTML]{EFEFEF}0.11} & \cellcolor[HTML]{EFEFEF}\textbf{0.79} \\ \cline{2-10} 
\multicolumn{1}{|c|}{}                           & \cellcolor[HTML]{C0C0C0}Sunset    & \multicolumn{1}{c|}{\cellcolor[HTML]{C0C0C0}0.87} & \multicolumn{1}{c|}{\cellcolor[HTML]{C0C0C0}0.82} & \multicolumn{1}{c|}{\cellcolor[HTML]{C0C0C0}\textbf{0.08}} & \cellcolor[HTML]{C0C0C0}0.53 & \multicolumn{1}{c|}{\cellcolor[HTML]{C0C0C0}\textbf{0.93}} & \multicolumn{1}{c|}{\cellcolor[HTML]{C0C0C0}\textbf{0.98}} & \multicolumn{1}{c|}{\cellcolor[HTML]{C0C0C0}0.20} & \cellcolor[HTML]{C0C0C0}\textbf{0.86} \\ \cline{2-10} 
\multicolumn{1}{|c|}{}                           & \cellcolor[HTML]{EFEFEF}Night 1   & \multicolumn{1}{c|}{\cellcolor[HTML]{EFEFEF}0.94} & \multicolumn{1}{c|}{\cellcolor[HTML]{EFEFEF}0.93} & \multicolumn{1}{c|}{\cellcolor[HTML]{EFEFEF}\textbf{0.07}} & \cellcolor[HTML]{EFEFEF}0.55 & \multicolumn{1}{c|}{\cellcolor[HTML]{EFEFEF}\textbf{0.96}} & \multicolumn{1}{c|}{\cellcolor[HTML]{EFEFEF}\textbf{0.99}} & \multicolumn{1}{c|}{\cellcolor[HTML]{EFEFEF}0.17} & \cellcolor[HTML]{EFEFEF}\textbf{0.91} \\ \cline{2-10} 
\multicolumn{1}{|c|}{\multirow{-5}{*}{\rotatebox[origin=c]{90}{Castro}}}   & \cellcolor[HTML]{C0C0C0}Night 2   & \multicolumn{1}{c|}{\cellcolor[HTML]{C0C0C0}0.93} & \multicolumn{1}{c|}{\cellcolor[HTML]{C0C0C0}0.93} & \multicolumn{1}{c|}{\cellcolor[HTML]{C0C0C0}\textbf{0.07}} & \cellcolor[HTML]{C0C0C0}0.55 & \multicolumn{1}{c|}{\cellcolor[HTML]{C0C0C0}\textbf{0.95}} & \multicolumn{1}{c|}{\cellcolor[HTML]{C0C0C0}\textbf{0.99}} & \multicolumn{1}{c|}{\cellcolor[HTML]{C0C0C0}0.17} & \cellcolor[HTML]{C0C0C0}\textbf{0.91} \\ \hline
\multicolumn{1}{l|}{}                            & Average                           & \multicolumn{1}{c|}{0.78}                         & \multicolumn{1}{c|}{0.69}                         & \multicolumn{1}{c|}{\textbf{0.12}}                                  & 0.43                         & \multicolumn{1}{c|}{\textbf{0.88}}                                  & \multicolumn{1}{c|}{\textbf{0.96}}                                  & \multicolumn{1}{c|}{0.25}                         & \textbf{0.82}                                  \\ \cline{2-10} 
\end{tabular}%
}
\end{table}

\end{document}